\documentclass{article}

\PassOptionsToPackage{numbers, compress}{natbib}
 \usepackage[preprint]{neurips_2026}

\usepackage[utf8]{inputenc} 
\usepackage[T1]{fontenc}    
\usepackage{hyperref}       
\usepackage{url}            
\usepackage{booktabs}       
\usepackage{amsfonts}       
\usepackage{nicefrac}       
\usepackage{microtype}      
\usepackage{xcolor}         
\usepackage{amsmath}         

\usepackage{graphicx}
\usepackage{subcaption}
\usepackage{wrapfig}
\usepackage{booktabs}
\usepackage{multirow}
\usepackage{float}
\usepackage{wrapfig}

\usepackage{mwe}

\title{Cross-scale Aligned Supervision for Training GANs}

\author{Sangeek Hyun \quad MinKyu Lee \quad Jae-Pil Heo\thanks{Corresponding Author.} \\
\\
Sungkyunkwan University
}

\begin{document}

\maketitle

\begin{abstract}
Modern GANs often introduce adversarial supervision on intermediate generator outputs and interpret the resulting multi-stage synthesis as coarse-to-fine hierarchical generation. 
In this work, we challenge this interpretation. 
We argue that standard scale-wise adversarial supervision does not construct a proper coarse-to-fine hierarchy: each intermediate image is independently pushed toward the real distribution at its own resolution, but this scale-wise realism does not ensure that outputs across stages represent the identical generated sample. 
Moreover, the scale-specific image produced at each stage is not used as an explicit refinement target for the subsequent stage.
Therefore, its adversarial loss can improve a scale-specific output without constraining later stages to preserve the same sample trajectory, allowing them to move toward a different sample rather than refine the previous output.
We refer to this problem as a cross-scale trajectory misalignment problem.
To resolve it, we propose CAT, a Cross-scale Aligned Transformer for multi-scale adversarial generation. 
CAT keeps the discriminator scale-wise, so each intermediate output is evaluated at its own resolution, while adding a simple generator-side consistency regularization that aligns intermediate outputs with the final output.
On class-conditional ImageNet-256, CAT-H/2 achieves an FID-50K of 1.56 with one-step inference after only 60 training epochs, outperforming strong one-step GAN and diffusion/flow baselines. 
Our analyses further show that CAT reduces cross-scale discrepancy, decreases inter-stage rewriting, and improves alignment with the final refinement direction.
\end{abstract}

\section{Introduction}

\begin{figure*}[h]
  \centering
  \begin{minipage}{0.60\textwidth}
    \centering
    \includegraphics[width=0.49\linewidth]{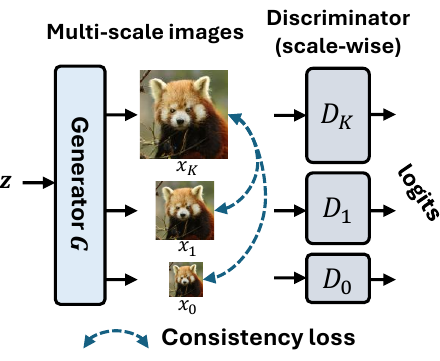}
    \includegraphics[width=0.49\linewidth]{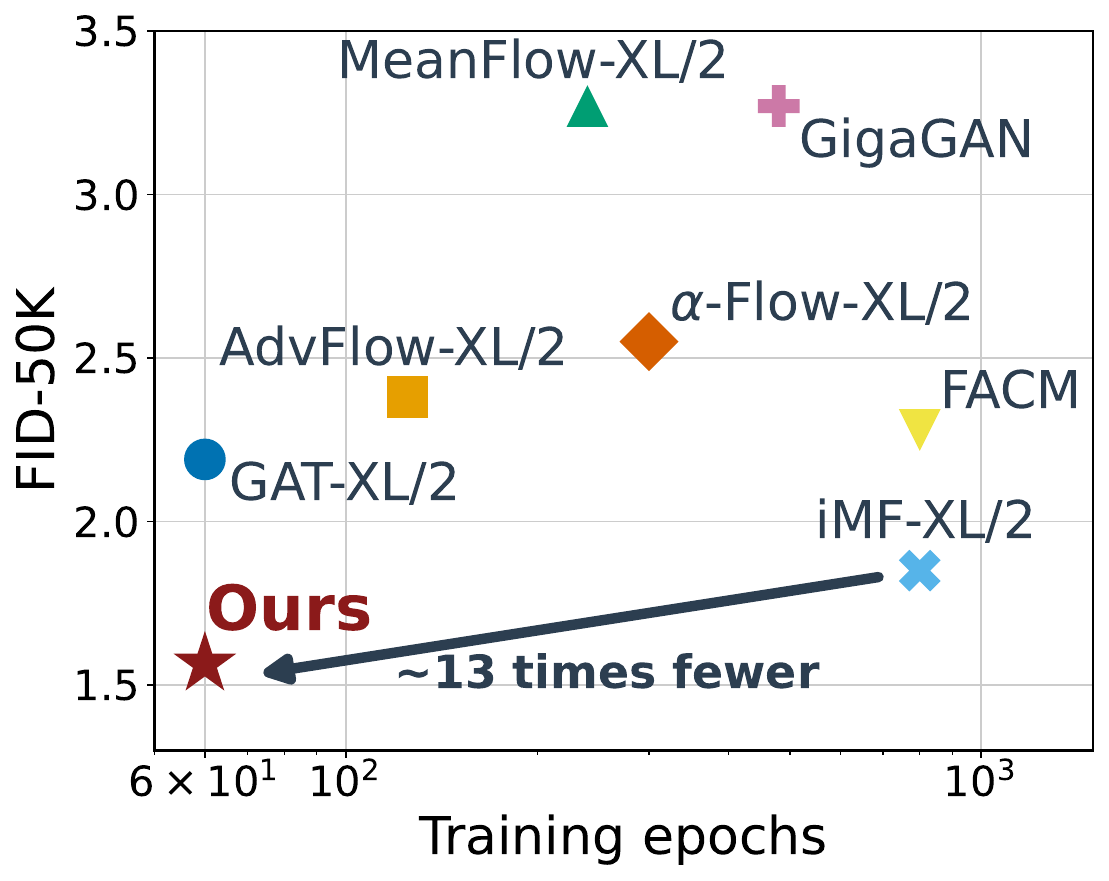}
  \end{minipage}
  \hfill
  \begin{minipage}{0.38\textwidth}
        \caption{\textbf{Method overview and comparison with 1-step baselines.}
        (Left) Our method combines generator-side consistency regularization with a scale-wise discriminator.
        (Right) Our method achieves strong FID with substantially fewer training epochs in one-step generation in ImageNet-256.
        }
    \label{fig: intro method overview and comparison}
  \end{minipage}
  \vspace{-3mm}
\end{figure*}

\begin{figure}[t]
    \centering
    \begin{subfigure}[t]{0.49\linewidth}
        \centering
        \includegraphics[width=\linewidth]{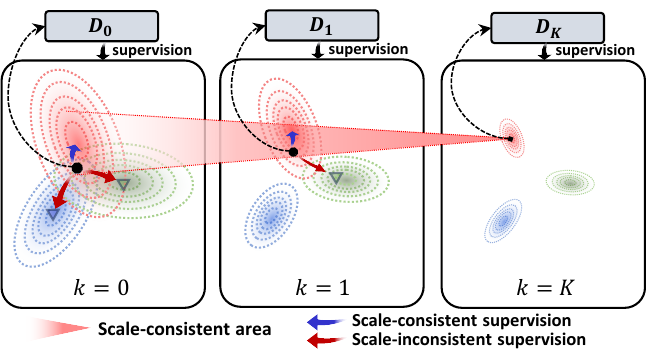}
        \vspace{-4.5mm}
        \caption{Training dynamics of scale-wise supervision}
        \label{fig: problem scale-wise supervision}
    \end{subfigure}
    \hfill
    \begin{subfigure}[t]{0.49\linewidth}
        \centering
            \includegraphics[width=\linewidth]{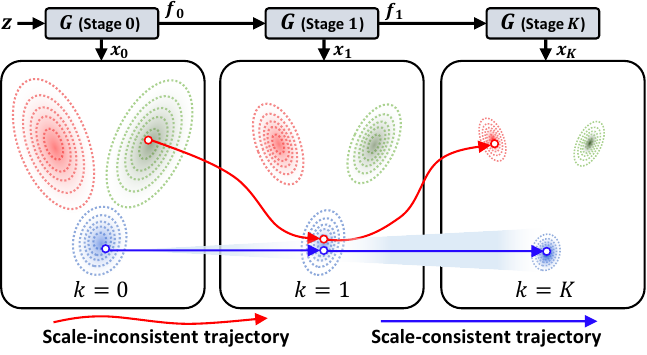}
        \vspace{-4mm}
        \caption{Possible generator-side trajectories}
        \label{fig: problem generator-side inconsisetncy}
    \end{subfigure}
    \vspace{-0mm}
    \caption{
    \textbf{Failure modes of standard scale-wise adversarial supervision.}
    (a) Since each scale-specific image is independently supervised by the discriminator at its own resolution, adversarial gradients can push different stages toward different realistic modes. 
    This enforces scale-wise realism, but not sample-wise cross-scale alignment. 
    (b) At each stage, $x_k$ is produced for adversarial supervision, while the subsequent stage continues from the generator feature $f_k$ rather than taking $x_k$ itself as input.
    Thus, under inconsistent supervision, later stages can follow a different sample trajectory rather than explicitly refine the previous output~($x_k$).
    In summary, adversarial training with scale-wise supervision does not by itself construct a proper coarse-to-fine hierarchy.
    }
    \vspace{-5mm}
    \label{fig: intro problem}
\end{figure}

Recent generative models have achieved remarkable progress in image synthesis.
A common principle behind many of these advances is to decompose generation into intermediate stages, so that the model solves a sequence of simpler prediction problems rather than synthesizing a complete image at once.
Diffusion models~\cite{ddpm,LDM,flowmatching,dit} follow this principle through iterative denoising, while autoregressive and masked prediction models~\cite{vqgan,maskgit,var,mar} factorize image generation into a sequence of prediction steps.
In these paradigms, intermediate states are not merely auxiliary predictions; they actively participate in the generation process and progressively guide the model toward the final sample.

Generative Adversarial Networks (GANs) have also pursued hierarchical generation through multi-stage synthesis.
Since GANs generate samples in a single forward pass, prior work has introduced adversarial supervision on intermediate generator outputs~\cite{msg-gan,progan,anycostgan,gigagan,gat}.
In modern GAN architectures~\cite{gigagan,aurora}, this idea is commonly instantiated as scale-wise adversarial supervision, where each generator-stage output is converted into a scale-specific image and the discriminator evaluates each resolution independently.
This design is usually interpreted as coarse-to-fine generation, where early stages form global structure and later stages refine details.

In this work, we challenge this interpretation.
We argue that standard scale-wise adversarial supervision does not construct a proper coarse-to-fine hierarchy: it independently optimizes each intermediate image as a scale-specific supervised output, without constraining how outputs across stages relate to one another.
As a result, intermediate images can become realistic at their own resolutions, but need not form progressively refined states of the same generated sample.

This failure originates from the supervision objective itself, as illustrated in Fig.~\ref{fig: problem scale-wise supervision}.
In the basic scale-wise formulation, each intermediate image is compared with real images at its corresponding resolution.
Such supervision provides direct scale-wise realism feedback, but it only matches per-scale distributions.
Since each scale is judged independently, the adversarial gradient at one stage can push its output toward a realistic mode that differs from the mode selected at another stage.
Therefore, outputs from different stages can become realistic at their own resolutions while failing to represent the same sample.
This breaks sample-wise cross-scale alignment, which is necessary for a proper coarse-to-fine hierarchy.

This issue is further reinforced by how intermediate outputs are used in multi-stage generators, as illustrated in Fig.~\ref{fig: problem generator-side inconsisetncy}.
At each stage, the scale-specific image $x_k$ is optimized for adversarial supervision, but it is not enforced as the image-level refinement target of the next stage.
Subsequent synthesis proceeds through the generator feature $f_k$, so later outputs can deviate from $x_k$ when scale-wise objectives provide inconsistent signals.
Thus, later stages may follow a different sample trajectory rather than refine the previous output.
Together, the supervision objective and the generator-side usage explain why standard scale-wise adversarial supervision can produce realistic intermediate images without constructing a coherent coarse-to-fine generation hierarchy.

Motivated by this, we propose CAT, a Cross-scale Aligned Transformer for multi-scale adversarial generation.
CAT keeps the discriminator scale-wise, preserving direct adversarial feedback for each generated image.
At the same time, it introduces a simple generator-side consistency regularization that aligns intermediate outputs with the final output.
This design applies scale-wise adversarial feedback to coordinated intermediate targets, allowing intermediate supervision to support final-stage synthesis rather than optimizing disconnected side predictions.

On ImageNet-256, CAT-H/2 achieves an FID-50K of 1.56 with one-step inference after only 60 training epochs, outperforming strong one-step GAN and diffusion/flow baselines with up to \(\sim 13\times\) fewer training epochs.
These results suggest that, when hierarchical adversarial supervision is properly organized, transformer-based GANs can serve as highly competitive one-step generative models.
\section{Preliminary}

\label{sec:3 backgrounds}
\paragraph{Generative Adversarial Networks.}
Generative Adversarial Networks~(GANs)~\citep{GAN} formulate image generation as an adversarial game between a generator $G(z,c)$ and a discriminator $D(x,c)$.
Here, $x \in \mathbb{R}^{H \times W \times 3}$ denotes an image sample that can come either from the real data distribution $p_{\mathrm{data}}(x \mid c)$ or from the generated distribution $p_G(x \mid c)$ induced by $G(z,c)$ with $z \sim p_z$, where $z$ and $c$ are random noise and condition, respectively.
The discriminator distinguishes real and generated samples, while the generator learns to fool it.

\paragraph{Multi-stage adversarial generation in GANs.}
Rather than supervising only the final generator output, several GAN frameworks expose intermediate images at multiple generator stages and apply adversarial feedback to them~\cite{msg-gan,progan,gigagan,gat}.
This design is often motivated as hierarchical or coarse-to-fine generation, where earlier stages provide coarse predictions and later stages refine them.
Such intermediate supervision can be realized through multi-scale images~\cite{msg-gan,gigagan} or multi-level noise perturbations~\cite{gat}; in this work, we focus on the multi-scale image formulation.

We denote by \(x_k\) the image used for adversarial supervision at stage or scale \(k\), where \(k=0,\ldots,K\) and \(x_K\) is the final output.
Each \(x_k\) is evaluated against real images represented at the corresponding scale, providing adversarial feedback throughout the generator.
At a high level, the generator maintains stage features \(f_k\) that carry the synthesis process from one stage to the next.
Thus, \(x_k\) denotes the image supervised at stage \(k\), while \(f_k\) denotes the generator hidden features from which subsequent synthesis proceeds.

A common approach for this multi-stage generation is \emph{scale-wise adversarial supervision}~\cite{anycostgan,gigagan}, where each intermediate image is evaluated independently at its corresponding resolution.
Let \(d_k\) denote the discriminator prediction for \(x_k\).
In scale-wise supervision, \(d_k\) is computed only from the corresponding image \(x_k\), without cross-scale information exchange inside the discriminator.
This provides direct scale-wise realism feedback to each generator-stage output.

Using these scale-specific predictions, the multi-scale adversarial objective is written as
\[
\mathcal{L}_{\mathrm{adv}}
=
\frac{1}{K+1}
\sum_{k=0}^{K}
\mathcal{L}_{\mathrm{GAN}}(G, D, k),
\]
where \(\mathcal{L}_{\mathrm{GAN}}(G,D,k)\) denotes the adversarial loss computed from the scale-\(k\) discriminator.

\section{Proposed Method}
\subsection{Cross-scale trajectory misalignment in scale-wise supervision}
\label{sec:trajectory_misalignment}

\paragraph{Problem.}
A proper coarse-to-fine hierarchy requires intermediate outputs to remain on the same sample trajectory.
That is, each intermediate image should not only look realistic at its own resolution, but also correspond to the final image that later stages will produce.
Standard scale-wise supervision optimizes each intermediate image independently against the real distribution at its corresponding resolution.
At each resolution, the real distribution contains many plausible samples, and the scale-wise objective only requires an intermediate output to match this distribution.
Therefore, realism at one scale does not impose a sample-wise correspondence with outputs at other scales.
As a result, different stages can receive valid adversarial feedback while converging toward different realistic samples, breaking the intended coarse-to-fine hierarchy.
We refer to this failure as \emph{cross-scale trajectory misalignment}.

\begin{figure*}[t] 
  \centering
  \begin{minipage}{0.62\textwidth}
    \centering
    \vspace{-2mm}
    \includegraphics[width=\linewidth]{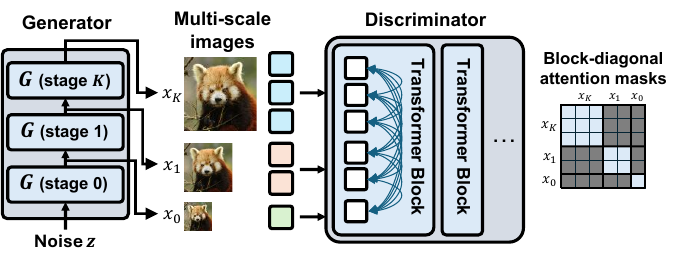}
  \end{minipage}
  \hfill
  \begin{minipage}{0.37\textwidth}
    \caption{
    \textbf{Analysis setup.}
    The generator produces scale-specific images \(\{x_k\}_{k=0}^{K}\).
    The discriminator concatenates tokens from all scales, but a block-diagonal attention mask prevents cross-scale information exchange, so each scale prediction is computed only from \(x_k\).
    }
    \label{fig: analysis setup}
  \end{minipage}
  \vspace{-3mm}
\end{figure*}

\paragraph{Analysis setup.}
We analyze cross-scale trajectory misalignment using a GAT-style transformer generator~\cite{gat}.
Since the transformer generator operates on a fixed latent grid, its stage-wise outputs are produced at the same latent resolution.
We denote the output of generator stage \(k\) by \(h_k\), and construct the scale-specific image for adversarial supervision by resizing it:
\[
x_k = r_k(h_k),
\]
where \(r_k(\cdot)\) denotes the resizing operation for scale \(k\).

The discriminator embeds each scale-specific image \(x_k\) into patch tokens and concatenates tokens from all scales along the sequence dimension.
This concatenation is used only for implementation efficiency; it does not allow cross-scale information exchange.
To enforce scale-wise discrimination, we apply a block-diagonal attention mask across scales.
Thus, tokens from scale \(k\), including its scale-specific prediction token~(\([{\rm cls}]\)), can attend only to tokens from the same scale.
Consequently, the scale-\(k\) prediction \(d_k\) is computed from \(x_k\) alone, without using information from other scale-specific images \(\{x_j\}_{j\neq k}\).
This implements scale-wise adversarial supervision while keeping all scales inside a single shared transformer discriminator.
The entire framework is illustrated in Fig.~\ref{fig: analysis setup}.

Unless otherwise specified, all analyses use the ImageNet-256 latent-space setting with SD-VAE latents~\cite{LDM}; for brevity, we refer to latents as images.
We use \textit{Base}-scale generator and discriminator, each with 12 layers and 768 hidden channels, and train for 20 epochs, corresponding to 50K iterations.

\paragraph{Metrics.}
We measure whether intermediate outputs are coherently accumulated toward the final output.
Since outputs at different scales have different resolutions, we compare them after resizing to the highest resolution.
Let \(x_K\) denote the final-stage output, and
we define
\begin{equation}
\begin{aligned}
\delta_k &=
\frac{
\left\| x_K - r_K(x_k) \right\|_2
}{
\left\| x_K \right\|_2
},
\\
R_k &=
\frac{
\left\| r_K(x_{k+1}) - r_K(x_k) \right\|_2
}{
\left\| x_K \right\|_2
},
\\
A_k &=
\cos\left(
r_K(x_{k+1}) - r_K(x_k),\,
x_K - r_K(x_k)
\right).
\end{aligned}
\end{equation}
Each metric captures a different requirement of progressive generation.
The discrepancy \(\delta_k\) measures whether the intermediate output remains close to the final sample after resolution matching; large \(\delta_k\) indicates that the stage output is not well aligned with the final image it is supposed to support.
The rewrite magnitude \(R_k\) measures how much the image changes from stage \(k\) to stage \(k+1\); large \(R_k\) indicates that the next stage substantially rewrites the previous output rather than refining it incrementally.
The direction alignment \(A_k\) measures whether the stage-wise update points toward the remaining difference to the final image; high \(A_k\) means that the update moves in the direction of the final output, while low \(A_k\) indicates that the update is poorly aligned with the intended refinement trajectory.

Together, these metrics diagnose whether intermediate outputs are accumulated into the final sample through a coherent coarse-to-fine process.

\begin{figure}[t]
    \centering
    \begin{subfigure}[t]{0.32\linewidth}
        \centering
        \includegraphics[width=\linewidth]{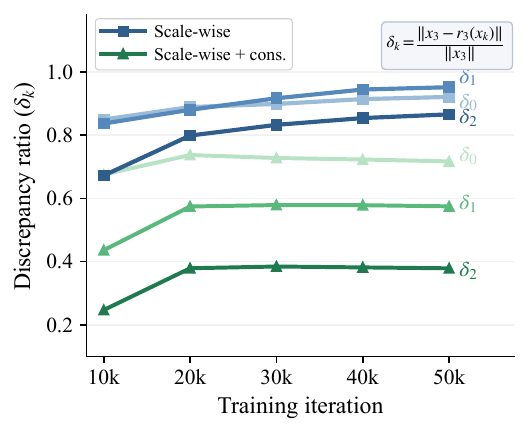}
        \vspace{-6mm}
        \caption{Distance to highest-scale}
        \label{fig: discrepancy and rewrite}
    \end{subfigure}
    \begin{subfigure}[t]{0.32\linewidth}
        \centering
        \includegraphics[width=\linewidth]{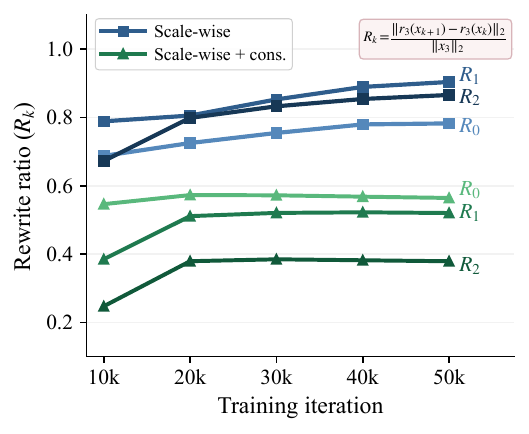}
        \vspace{-6mm}
        \caption{Inter-stage rewrite magnitude}
        \label{fig: inter-stage rewrite ratio}
    \end{subfigure}
    \begin{subfigure}[t]{0.32\linewidth}
        \centering
        \includegraphics[width=\linewidth]{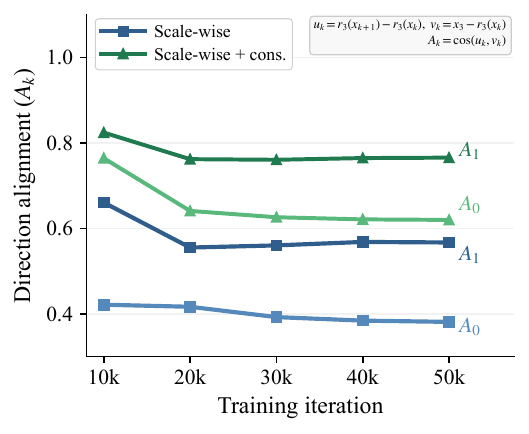}
        \vspace{-6mm}
        \caption{Rewrite direction alignment}
        \label{fig: rewrite direction alignment}
    \end{subfigure}
    \caption{
        \textbf{Cross-scale inconsistency analysis.}
        We analyze whether intermediate outputs are coherently accumulated toward the final output under standard scale-wise supervision.
        (a) The distance to the highest-scale image~($\delta_k$) remains large, showing that intermediate outputs stay far from the final image.
        (b) The inter-stage rewrite magnitude~($R_k$) is also large, indicating that later stages substantially rewrite the inherited image rather than applying small refinements.
        (c) The rewrite direction alignment~($A_k$) remains low, showing that stage-wise updates are not well aligned with the remaining direction toward the final image.
        Adding the proposed consistency regularization mitigates these failures by reducing $\delta_k$ and $R_k$ while increasing $A_k$.
    }
    \vspace{-0.25cm}
    \label{fig:cross_scale_inconsistency}
\end{figure}

\paragraph{Observation.}
As shown in Fig.~\ref{fig:cross_scale_inconsistency}, scale-wise supervision exhibits substantial cross-scale trajectory misalignment.
The discrepancy \(\delta_k\) remains large throughout training, often exceeding \(0.8\), meaning that the distance from an intermediate output to the final output is comparable to the magnitude of the final image itself.
Thus, the mismatch is not a small residual difference, but a large deviation from the final sample trajectory.
The rewrite magnitude \(R_k\) is also consistently large, again often above \(0.8\), showing that later stages do not merely add missing details but substantially revise the outputs produced by earlier stages.
Moreover, \(A_k\) remains low, indicating that these large stage-wise changes are only weakly aligned with the remaining direction toward the final image.

Notably, both \(\delta_k\) and \(R_k\) tend to increase over the course of training, rather than decrease.
Also, they do not diminish as the stage becomes finer; that is, moving to higher-resolution stages (\(k\)) does not reduce the discrepancy to the final output or the amount of rewriting.
If scale-wise supervision induced a proper coarse-to-fine hierarchy, we would expect later stages to progressively reduce the remaining difference and apply more localized refinements.
Instead, the observed trend suggests that training under standard scale-wise supervision amplifies cross-scale inconsistency, causing later stages to repeatedly revise earlier outputs rather than coherently refine them.

\subsection{Cross-scale aligned supervision}

The analysis above suggests that one missing component in scale-wise supervision is cross-scale trajectory alignment, rather than additional per-scale realism feedback alone.
Each intermediate image already receives direct adversarial feedback at its own resolution, yet these outputs are not constrained to remain on the same sample trajectory as the final image.
We therefore keep the discriminator scale-wise and add an explicit generator-side constraint that aligns intermediate outputs with the final output.
This preserves direct scale-wise realism feedback while encouraging intermediate stages to support the same final sample.

\paragraph{Generator-side consistency regularization.}
We implement this generator-side alignment as a consistency loss on the stage-wise outputs.
The goal is not to add another realism objective, but to ensure that the intermediate outputs receiving scale-wise adversarial feedback remain on the cross-scale consistent trajectory.
To this end, we use the final-stage output as a common anchor for all intermediate stages.
By aligning intermediate outputs to this anchor, the consistency loss directly targets the failure observed above: it reduces excessive discrepancy to the final output and discourages later stages from rewriting earlier outputs toward a different sample.

Let \(h_k\) denote the direct output of the \(k\)-th generator stage before the resizing operation used to provide it to the discriminator.
We align each intermediate stage with the final stage by
\begin{equation}
\mathcal{L}_{\mathrm{cons}}
=
\frac{1}{K}
\sum_{k=0}^{K-1}
w_k
\left\|
h_k - h_K
\right\|_2^2 ,
\end{equation}
where \(h_K\) is the final-stage output and \(w_k\) is a scale weight.
We use weaker weights for lower-resolution stages because coarse outputs are inherently ambiguous: many high-resolution samples can share similar low-resolution structure.
This prevents the consistency loss from imposing an overly rigid point-to-point constraint on early stages, while still encouraging them to remain on the same sample trajectory as the final output.

\paragraph{Cross-scale Aligned Transformer.}
Based on this regularization, we propose CAT~(Cross-scale Aligned Transformer), which combines a scale-wise discriminator with generator-side consistency regularization.
Specifically, each stage output \(h_k\) is resized to form the scale-specific discriminator input \(x_k=r_k(h_k)\), and the discriminator prediction is computed only from the corresponding scale.
Thus, the discriminator preserves direct scale-wise adversarial feedback, while the consistency loss aligns the intermediate outputs receiving this feedback.

The generator objective is
\begin{equation}
\mathcal{L}_{G}
=
\mathcal{L}_{\mathrm{adv}}
+
\lambda_{\mathrm{cons}}
\mathcal{L}_{\mathrm{cons}},
\end{equation}
where \(\mathcal{L}_{\mathrm{adv}}\) is computed from the scale-wise discriminator predictions.
The discriminator objective remains unchanged.

\section{Experiments}
\label{sec.4: experiments}

\paragraph{Experimental settings.}
We evaluate class-conditional image generation on ImageNet-256~\cite{imagenet} at \(256 \times 256\) resolution.
Following prior latent-space one-step generators, we train all models in the latent space of SD-VAE~\cite{LDM}.
Our implementation is largely based on GAT~\cite{gat}: we adopt its generator and most configurations, including the objective functions.
For CAT, we use the scale-wise discriminator with generator-side consistency regularization.
Following prior work~\cite{meanflow,improvedmeanflow}, we report FID-50K~\cite{fid} using statistics computed from the full ImageNet training set.

\paragraph{Implementation details.}

\begin{wraptable}{r}{0.3\linewidth}
\vspace{-4mm}
\centering
\scriptsize
\caption{Model configs.}
\label{tab:model_config}
\setlength{\tabcolsep}{2pt}
\renewcommand{\arraystretch}{0.95}
\begin{tabular}{@{}lcccc@{}}
\toprule
Model & $L$ & $C$ & Params & GFLOPs \\
\midrule
G-B/2 & 12 & 768  & 133M & 23.0 \\
G-M/2 & 24 & 768  & 261M & 46.0 \\
G-H/2 & 32 & 1280 & 960M & 166.8 \\
\cmidrule(lr){1-5}
D-B/2 & 12 & 768  & 96M  & 33.9 \\
\bottomrule
\end{tabular}
\vspace{-4mm}
\end{wraptable}

We use \(\lambda_{\mathrm{cons}}=0.1\) for all experiments.
The scale weights $w_i$ are decreased toward lower resolutions: for \(K=3\), we set \(w_0=1/3\), \(w_1=1/2\), and \(w_2=1\).

We find that stable generator scaling can be achieved without increasing the discriminator capacity or reducing the generator learning rate, unlike the recipe from GAT~\cite{gat}.
Thus, unless otherwise specified, we use a \textit{Base} discriminator for all generator scales~(\textit{Base}, \textit{Medium}, \textit{Huge}) and use the same learning rate for the generator and discriminator, as summarized in Table~\ref{tab:model_config}.
We use a batch size of 512, where 50K iterations correspond to 20 epochs, and a learning rate of \(2 \times 10^{-4}\).
For multi-scale adversarial supervision, we use token resolutions of \(2^2, 4^2, 8^2,\) and \(16^2\).
Compared with the original GAT discriminator operating on \(16^2\) tokens, this increases the discriminator input token count from \(256\) to \(340\), i.e., by about \(33\%\).
This introduces only a modest overhead, especially since we keep the discriminator at the \textit{Base} scale for all generator sizes instead of scaling it together with the generator.
For further details, please refer to Appendix~\ref{sec: appendix implementation}.

\begin{table*}[t]
\centering
\caption{\textbf{Class-conditional generation on ImageNet-256$\times$256~(FID-50K)}.
(Left) 1 Number of Function Evaluation~(NFE) generative models.
(Right) Other generative models including autoregressive models and multi-step diffusion/flow models.
Diffusion/flow entries are reported under CFG, when applicable. 
Reported GFLOPs indicates the inference cost of the generator.
}
\begin{minipage}[t]{0.52\textwidth}
    \centering
    \setlength{\tabcolsep}{5pt}
    \footnotesize
    \begin{tabular}{lcccc}
        \toprule
        {Method} & {Params} & GFLOPs & Epoch & {FID} \\
        \midrule
        \multicolumn{3}{l}{\textit{\textbf{1-NFE diffusion/flow from scratch}}} \\
        ~~iCT-XL/2~\cite{ict}                             & 675M & 119 & - & 34.24 \\
        ~~Shortcut-XL/2~\cite{shortcutmodel}                         & 675M & 119 & 250 & 10.60 \\
        ~~{MeanFlow}-XL/2~\cite{meanflow}                       & 676M & 119 & 240 & 3.43 \\
        ~~$\alpha$-Flow-XL/2~\cite{alphaflow}                    & 676M & 119 & 300 & 2.58 \\
        ~~FACM~\cite{facm}                                  & 675M & 119 & 800 & 2.27 \\
        ~~iMF-XL/2~\cite{improvedmeanflow}                              & 610M & 175 & 800 & 1.72 \\
        \midrule
        \multicolumn{3}{l}{\textit{\textbf{1-NFE GANs from scratch}}} \\
        ~~BigGAN~\cite{biggan}                                & 112M & 59 & - & 6.95 \\
        ~~GigaGAN~\cite{gigagan}                               & 569M & - & 480 & 3.45 \\
        ~~AdvFlow-XL/2~\cite{advflow}                          & 673M & - & 125 & 2.38 \\
        ~~StyleGAN-XL~\cite{sgxl}                            & 166M & 1574 & - & 2.30 \\
        ~~GAT-XL/2~\cite{gat}                              & 602M & 119 & 60 & 2.18 \\
        \midrule
        ~~CAT-M/2~(Ours)                                  & 261M & 46 & \textbf{40} & \textbf{1.93} \\
        ~~CAT-H/2~(Ours)                                  & 960M & 167 & \textbf{60} & \textbf{1.56} \\
        \bottomrule
    \end{tabular}
\end{minipage}
\hfill
\hspace{0.007\textwidth}
\begin{minipage}[t]{0.46\textwidth}
    \vspace{-89pt}
    \centering
    \setlength{\tabcolsep}{4pt}
    \footnotesize
    \begingroup
    \color{gray}
    \renewcommand{\arraystretch}{0.97} 
    \begin{tabular}{lccc}
        \toprule
        {Method} & {Params} & {NFE} & FID \\
        \midrule
        \multicolumn{4}{l}{\textbf{\textit{Multi-step autoregressive/masking}}} \\
        ~~MaskGIT~\cite{maskgit}           & 227M & 8 & 6.18 \\
        ~~STARFlow~\cite{starflow}          & 1.4B & 1024$\times$2 & 2.40 \\
        ~~VAR-$d30$~\cite{var}         & 2B & 10$\times$2 & 1.92 \\
        ~~MAR-H~\cite{mar}             & 943M & 256$\times$2 & 1.55 \\
        ~~RAR-XXL~\cite{RAR}           & 1.5B & 256$\times$2 & 1.48 \\
        ~~xAR-H~\cite{xAR}             & 1.1B & 50$\times$2 & 1.24 \\
        \midrule
        \multicolumn{4}{l}{\textbf{\textit{Multi-step diffusion/flow}}} \\
        ~~LDM-4-G~\cite{LDM}           & 400M & 250$\times$2 & 3.60 \\
        ~~SimDiff~\cite{simplediff}           & 2B & 512$\times$2 & 2.77 \\
        ~~DiT-XL/2~\cite{dit}          & 675M & 250$\times$2 & 2.27 \\
        ~~SiT-XL/2~\cite{sit}          & 675M & 250$\times$2 & 2.06 \\
        ~~SiT-XL/2+REPA~\cite{repa}     & 675M & 250$\times$2 & 1.42 \\
        ~~LightningDiT-XL/2~\cite{lightningdit} & 675M & 250$\times$2 & 1.35 \\
        ~~DDT-XL/2~\cite{ddt} & 675M & 250$\times$2 & 1.26 \\
        ~~RAE+DiT$^{\text{DH}}$-XL~\cite{rae} & 839M & 250$\times$2 & \textbf{1.13} \\
        \bottomrule
    \end{tabular}
    \endgroup
\end{minipage}
\label{tab:all_results}
\vspace{-4mm}
\end{table*}

\paragraph{Comparison with prior work.}
As shown in Table~\ref{tab:all_results}, we compare the proposed method~(CAT) with prior work.
Among 1-Number of Function Evaluation~(NFE) models trained from scratch, CAT-H/2 achieves a new state-of-the-art FID of 1.56.
Notably, it substantially improves over recent 1-NFE diffusion/flow models, including iMF-XL/2~\cite{improvedmeanflow}, reducing FID from 1.72 to 1.56 while requiring only 60 training epochs, significantly fewer than 800.
CAT-H/2 also establishes a new state of the art among GAN-based models, outperforming strong recent baselines such as GAT-XL/2~\cite{gat}.
We also highlight that, although CAT-H/2 uses a larger number of parameters, this does not directly translate into higher practical cost~(Tab.~\ref{tab:compute_main}).
In terms of both training and inference GFLOPs, CAT-H/2 is cheaper than iMF-XL/2 while achieving better FID.

\begin{figure}[t]
\centering
\small
\begin{minipage}[c]{0.38\linewidth}
    \centering
    \includegraphics[width=\linewidth]{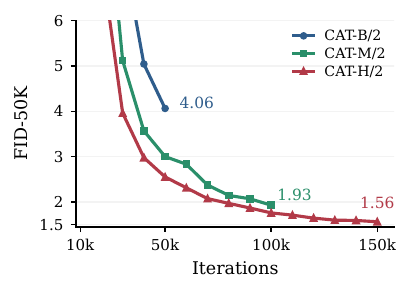}
    \vspace{-0.7cm}
    \captionof{figure}{
    FID-50K training curve.
    }
    \label{fig: fid curve}
\end{minipage}
\hspace{-10mm}
\begin{minipage}[c]{0.65\linewidth}
    \centering
    \footnotesize

    \begin{minipage}[c]{0.62\linewidth}
        \centering
        \setlength{\tabcolsep}{2.5pt}
        \renewcommand{\arraystretch}{0.95}
        \begin{tabular}{@{}lccc@{}}
        \toprule
        \multirow{2}{*}{Model}
        & \multirow{2}{*}{$\mathcal{L}_{\mathrm{cons}}$}
        & \multicolumn{2}{c}{FID-50K} \\
        & & 20 epo & 40 epo \\
        \midrule
        G-B/2 &            & 5.43 & -- \\
        G-B/2 & \checkmark & 4.06 & -- \\
        G-M/2 &            & 3.27 & 2.34 \\
        G-M/2 & \checkmark & \textbf{3.00} & \textbf{1.93} \\
        \bottomrule
        \end{tabular}
    \end{minipage}
    \hspace{-7mm}
    \begin{minipage}[c]{0.38\linewidth}
        \captionsetup{type=table}
        \vspace{3mm}
        \caption{
        Ablation study on $\mathcal{L}_{\mathrm{cons}}$. Consistency regularization becomes more effective for longer training in larger model.
        }
        \label{tab:ablation_cons}
    \end{minipage}

    \vspace{-1mm}

    \begin{minipage}[c]{0.62\linewidth}
        \centering
        \setlength{\tabcolsep}{5pt}
        \renewcommand{\arraystretch}{0.95}
        \begin{tabular}{@{}lccc@{}}
        \toprule
        $\lambda_{\mathrm{cons}}$ & 0.0 & 0.1 & 1.0 \\
        \midrule
        FID-50K & 5.43 & \textbf{4.06} & 4.45 \\
        \bottomrule
        \end{tabular}
    \end{minipage}
    \hspace{-7mm}
    \begin{minipage}[c]{0.38\linewidth}
        \captionsetup{type=table}
        \vspace{3mm}
        \caption{
        Analysis on the strength of $\lambda_{\mathrm{cons}}$~(G-B/2, 20 epoch).
        }
        \label{tab:ablation_cons_coeff}
    \end{minipage}

\end{minipage}

\vspace{-2mm}
\end{figure}

\paragraph{Training dynamics and consistency ablations.}
Fig.~\ref{fig: fid curve} shows the FID-50K training curves of CAT with different generator sizes.
CAT consistently benefits from scaling the generator, and larger models continue to improve with longer training.
In particular, CAT-H/2 steadily improves up to 150K iterations, reaching an FID of 1.56, while CAT-M/2 reaches 1.93 at 100K iterations.
This suggests that the proposed method remains stable and scalable under longer training.

Table~\ref{tab:ablation_cons} further verifies the effect of the proposed consistency regularization.
For G-B/2, adding $\mathcal{L}_{\mathrm{cons}}$ improves FID from 5.43 to 4.06 at 20 epochs.
For the larger G-M/2 model, the gain becomes more pronounced with longer training: $\mathcal{L}_{\mathrm{cons}}$ improves FID from 3.27 to 3.00 at 20 epochs, and from 2.34 to 1.93 at 40 epochs.
These results indicate that consistency regularization is particularly important when scaling the generator and extending training, where scale-wise supervision can otherwise accumulate larger cross-scale discrepancies.
Finally, Table~\ref{tab:ablation_cons_coeff} studies the strength of the consistency loss.
A moderate weight of $\lambda_{\mathrm{cons}}=0.1$ gives the best result, showing that explicit cross-scale alignment is beneficial, while overly strong consistency can over-constrain the generator.

\begin{table*}[t]
\centering
\footnotesize

\begin{minipage}[t]{0.58\linewidth}
\centering
\caption{
Compute comparison~(GFLOPs). Training and inference compute is measured per-iteration and sample.
}
\label{tab:compute_main}
\setlength{\tabcolsep}{6pt}
\begin{tabular}{@{}lccccc@{}}
\toprule
\multirow{2}{*}{Method} 
& Train & Infer. 
& \multirow{2}{*}{Epochs} 
& Total & Rel. \\
& GFLOPs & GFLOPs 
& 
& $(\times 10^3)$ & Total \\
\midrule
iMF-XL/2 & $1{,}306.3$ & $174.6$ & $800$ & $1{,}045.0$ & $16.7\times$ \\
GAT-XL/2 & $2{,}297.2$ & $118.6$ & $60$  & $137.8$   & $2.2\times$ \\
CAT-H/2  & $1{,}040.2$ & $166.7$ & $60$  & $62.4$    & $1.0\times$ \\
\bottomrule
\end{tabular}
\end{minipage}
\hfill
\begin{minipage}[t]{0.4\linewidth}
\centering
\caption{
Comparison with GAT across various model sizes~(20 epochs).
}
\label{tab:gat_comparison}
\setlength{\tabcolsep}{3pt}
\begin{tabular}{@{}lccc@{}}
\toprule
Method & $(G{+}D)$ Params & FID$\downarrow$ & IS$\uparrow$ \\
\midrule
GAT-B/2  & $116{+}104$M & $9.534$ & $124.69$ \\
GAT-XL/2 & $602{+}467$M & $4.021$ & $204.26$ \\
CAT-B/2  & $133{+}96$M  & $4.063$ & $174.54$ \\
CAT-H/2  & $960{+}96$M  & $2.552$ & $242.45$ \\
\bottomrule
\end{tabular}
\end{minipage}

\vspace{-2mm}
\end{table*}

\paragraph{Training and inference efficiency.}
\label{sec: training and inference efficiency}
Table~\ref{tab:compute_main} shows that CAT-H/2 is computationally efficient among strong one-step generators.
Compared with the one-step diffusion/flow baseline iMF-XL/2~\cite{improvedmeanflow}, CAT-H/2 achieves better FID with lower training and inference cost, while requiring over $16\times$ fewer training GFLOPs.
CAT-H/2 is also much cheaper to train than GAT-XL/2~\cite{gat}, reducing total training compute by about $2.2\times$.
For details about the compute comparison, please refer to Appendix~\ref{sec: appendix gflops}.

Table~\ref{tab:gat_comparison} further highlights that the gain does not simply come from increasing adversarial model capacity.
GAT jointly processes multi-scale evidence inside the discriminator, whereas CAT keeps discriminator feedback scale-wise and imposes alignment through generator-side consistency.
We observe that this design leads to a substantially stronger trade-off between capacity and performance.
CAT-B/2 already achieves an FID comparable to GAT-XL/2 ($4.063$ vs. $4.021$), despite operating at a \textit{Base}-scale configuration.
Moreover, CAT-H/2 significantly outperforms GAT-XL/2 ($2.552$ vs. $4.021$ FID), even though the two models have comparable total $(G{+}D)$ parameters.
These results suggest that the key advantage of CAT comes from organizing hierarchical adversarial supervision to provide clean scale-wise feedback while maintaining cross-scale alignment, rather than from simply scaling both the generator and discriminator.

\begin{figure}[t] 
  \centering
\includegraphics[width=0.98\linewidth]{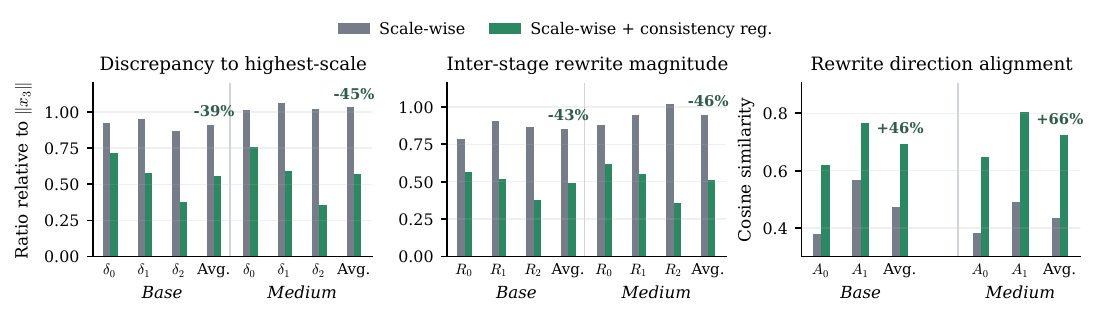}
    \vspace{-2mm}
    \caption{
    \textbf{Effect of generator-side consistency regularization.}
    Under scale-wise supervision, adding $\mathcal{L}_{\mathrm{cons}}$ improves cross-scale alignment across different model sizes, reducing $\delta_k$ and $R_k$ while increasing $A_k$.
    We use the 20 epoch checkpoint for \textit{Base} and the 40 epoch checkpoint for \textit{Medium}.
    }
    \label{fig: consistency effect}
\end{figure}

\paragraph{Effect of consistency regularization.}
We also verify whether the proposed consistency loss improves the cross-scale alignment of scale-wise supervision.
As shown in Fig.~\ref{fig: consistency effect}, adding $\mathcal{L}_{\mathrm{cons}}$ consistently reduces both the discrepancy to the highest-scale output $\delta_k$ and the inter-stage rewrite magnitude $R_k$ across \textit{Base} and \textit{Medium} models.
In terms of average values, $\mathcal{L}_{\mathrm{cons}}$ reduces $\delta_k$ by 39\% and 45\%, and reduces $R_k$ by 43\% and 46\%, for \textit{Base} and \textit{Medium} models, respectively.
It also improves the rewrite direction alignment $A_k$ by 46\% and 66\%, indicating that stage-wise updates become more aligned with the final refinement trajectory.
These results show that the proposed regularization mitigates the inconsistency of scale-wise supervision and encourages intermediate outputs to contribute more coherently to the final image.

\paragraph{Effect of discriminator scaling.}

\begin{figure}[t]
  \centering
  \begin{minipage}[t]{0.59\linewidth}
    \vspace{0pt}
    \centering
    \includegraphics[width=0.46\linewidth]{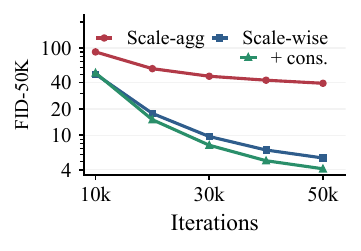}
    \includegraphics[width=0.50\linewidth]{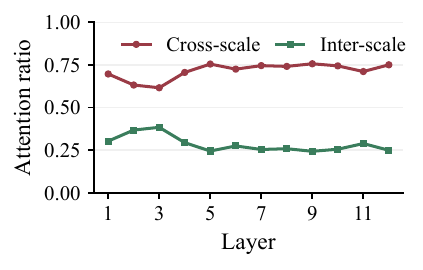}
    \vspace{-3mm}
    \caption{
    Analysis on scale-aggregated discriminator.
    }
    \label{fig: scale-agg fid and CA}
  \end{minipage}
  \hfill
  \begin{minipage}[t]{0.40\linewidth}
    \vspace{10pt}
    \centering
    \small
    \setlength{\tabcolsep}{5pt}
    \renewcommand{\arraystretch}{1.3}
    \begin{tabular}{lc}
    \toprule
    Model & FID-50K \\
    \midrule
    G-M/2 / D-B/2 & 5.11 \\
    G-M/2 / D-M/2 & \textbf{4.28} \\
    \bottomrule
    \end{tabular}
    \vspace{1mm}
    \captionof{table}{Discriminator scaling~(12 epo).}
    \label{tab:d_scaling}
  \end{minipage}
  \vspace{-2mm}
\end{figure}
We further examine the effect of discriminator scaling under a fixed G-M/2 generator.
To isolate the role of discriminator capacity, we change only the discriminator size while keeping the generator and training setup unchanged.
As shown in Tab.~\ref{tab:d_scaling}, increasing the discriminator from D-B/2 to D-M/2 improves FID-50K from 5.11 to 4.28 after 12 epochs.
This shows that CAT can still benefit from a stronger discriminator, although our main experiments intentionally use a \textit{Base} discriminator to maintain training efficiency.
These results suggest that further performance gains may be obtained by studying the generator-discriminator capacity balance more systematically, which we leave to future work.

\paragraph{Discussion on scale-aggregated discrimination.}
As an additional diagnostic, we examine a natural alternative to our design: allowing the discriminator to observe all scale-specific images jointly, following prior work~\cite{msg-gan}.
This analysis addresses whether cross-scale alignment can be obtained simply by giving the discriminator access to the full image pyramid.
We implement it simply by removing the generator-side consistency regularization and discriminator attention mask, while keeping the remaining settings unchanged, with \textit{Base}-sized models.

As shown in Fig.~\ref{fig: scale-agg fid and CA}, this scale-aggregated variant performs severely worse than the scale-wise counterpart.
Also, they show strong cross-scale interaction, where tokens from one resolution attend substantially to tokens from other resolutions.
This suggests that the discriminator can rely on cross-scale evidence or cross-scale compatibility, rather than judging each scale only by its own image-level realism, as noted in prior work~\cite{gigagan,anycostgan}.
Therefore, simply giving all scales to the discriminator can entangle the adversarial feedback across scales, instead of preserving direct scale-wise supervision.
This provides additional support for keeping the discriminator scale-wise and imposing cross-scale alignment on the generator side.

\begin{figure}[t]
\centering
\small

\begin{subfigure}[t]{0.49\linewidth}
    \centering
    \includegraphics[width=\linewidth]{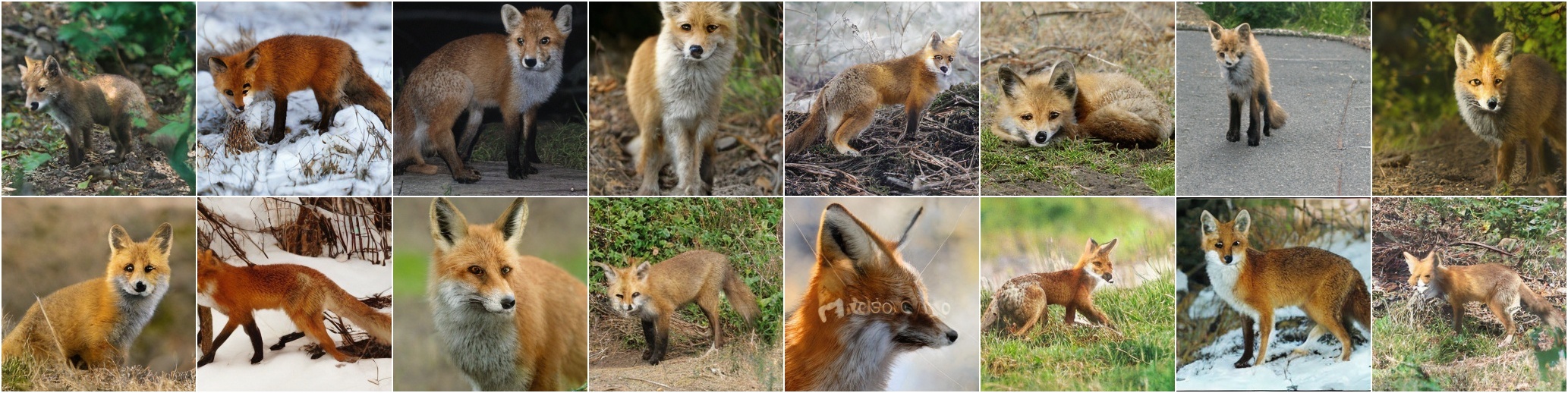}
    \vspace{-5mm}
    \caption{class 277: red fox}
    \label{fig:grid_1}
\end{subfigure}
\hfill
\begin{subfigure}[t]{0.49\linewidth}
    \centering
    \includegraphics[width=\linewidth]{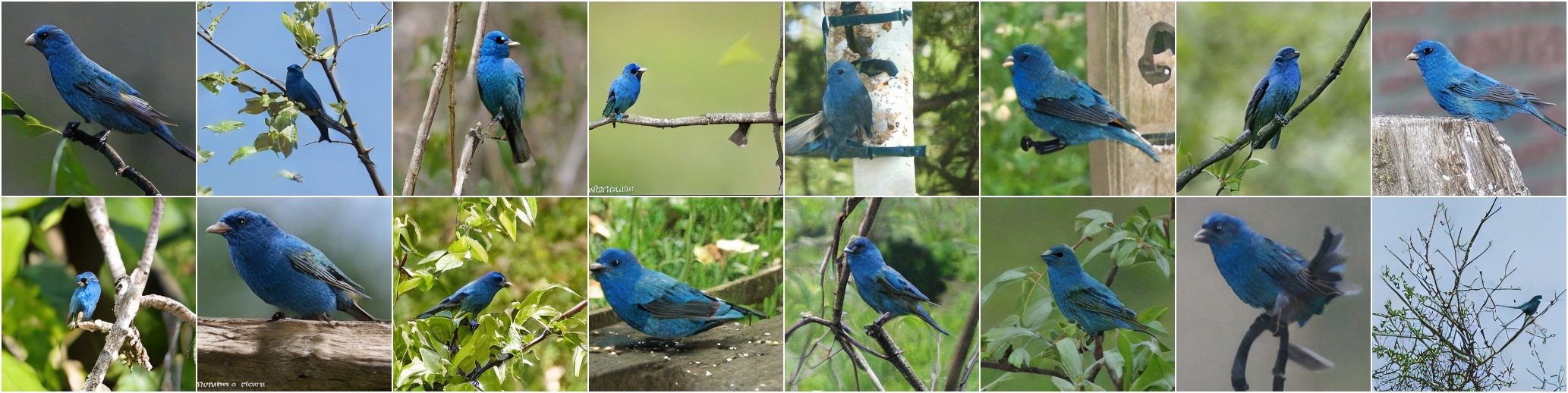}
    \vspace{-5mm}
    \caption{class 14: indigo bunting}
    \label{fig:grid_2}
\end{subfigure}

\vspace{1mm}

\begin{subfigure}[t]{0.49\linewidth}
    \centering
    \includegraphics[width=\linewidth]{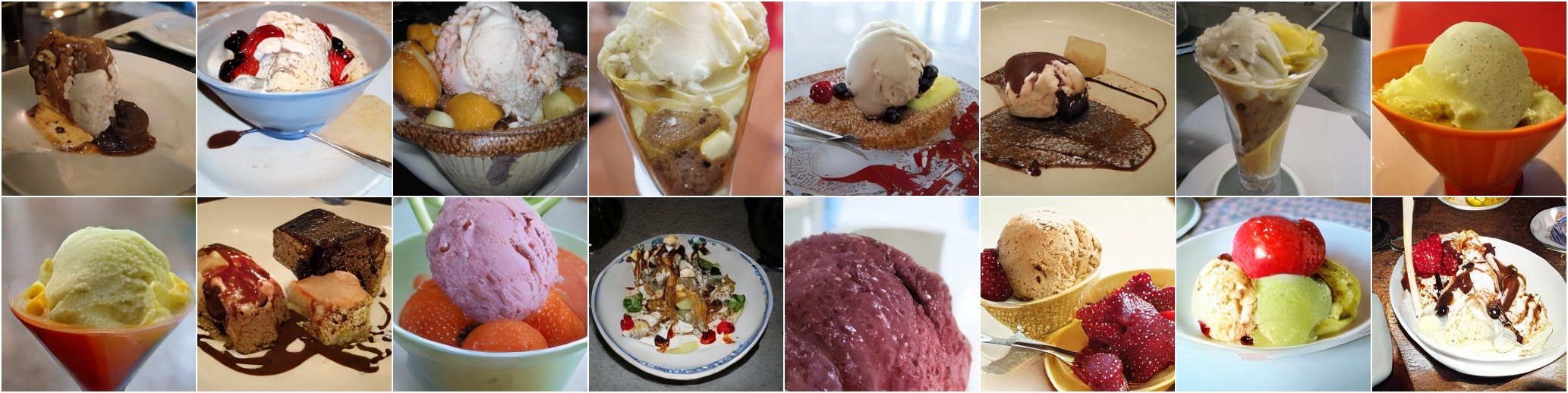}
    \vspace{-5mm}
    \caption{class 928: ice cream}
    \label{fig:grid_3}
\end{subfigure}
\hfill
\begin{subfigure}[t]{0.49\linewidth}
    \centering
    \includegraphics[width=\linewidth]{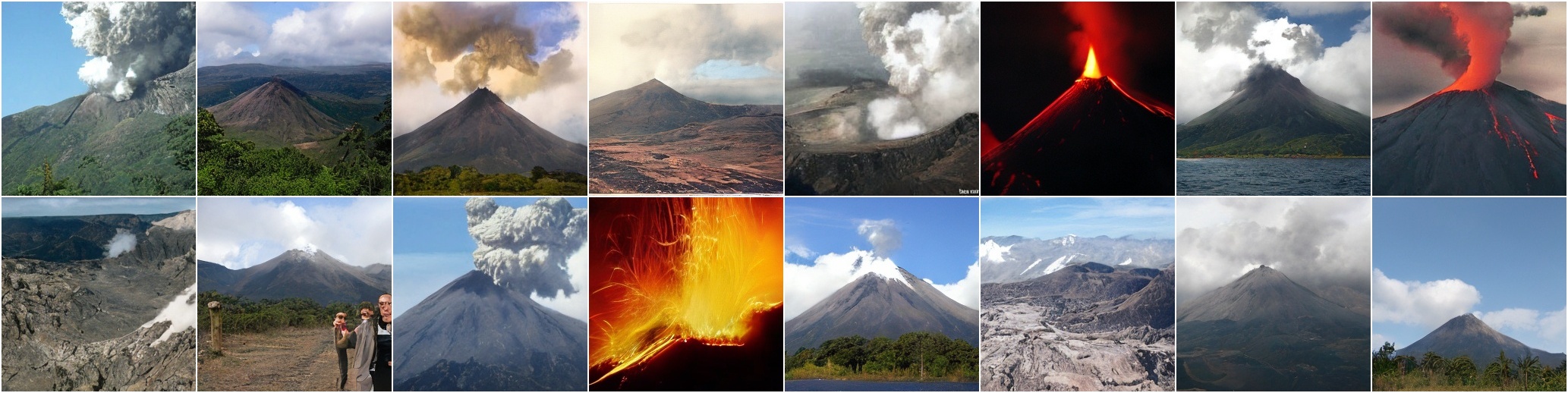}
    \vspace{-5mm}
    \caption{class 980: volcano}
    \label{fig:grid_4}
\end{subfigure}

\vspace{1mm}

\begin{subfigure}[t]{0.49\linewidth}
    \centering
    \includegraphics[width=\linewidth]{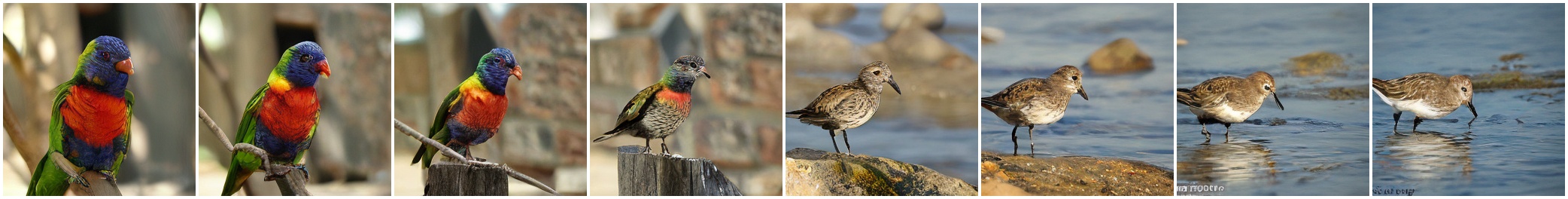}
    \vspace{-5mm}
    \caption{class 90 to 140: lorikeet to red-backed sandpiper}
    \label{fig:grid_5}
\end{subfigure}
\hfill
\begin{subfigure}[t]{0.49\linewidth}
    \centering
    \includegraphics[width=\linewidth]{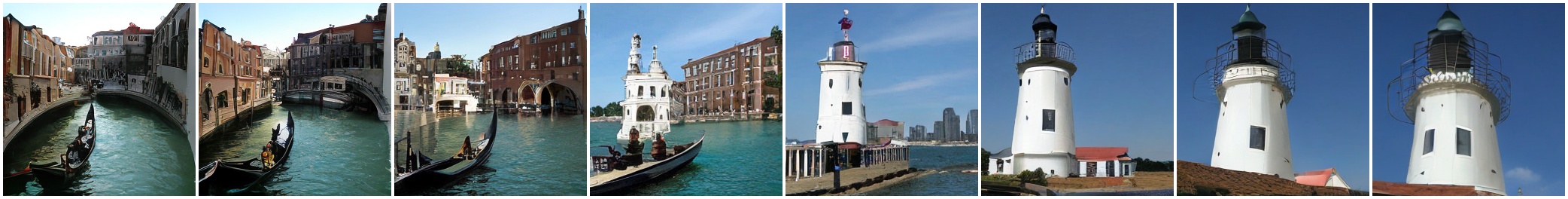}
    \vspace{-5mm}
    \caption{class 576 to 437: gondola to beacon}
    \label{fig:grid_6}
\end{subfigure}

\caption{
Uncurated generated examples and latent interpolation on ImageNet-256.
}
\label{fig:qualitative_grid}
\vspace{-3mm}
\end{figure}
\paragraph{Qualitative results.}
Fig.~\ref{fig:qualitative_grid} shows uncurated samples generated by CAT-H/2 on ImageNet-256.
The samples demonstrate that CAT produces diverse and high-fidelity images across different object categories with a single generator forward pass.
We provide additional uncurated samples and qualitative comparison with iMF-XL/2 in Appendix~\ref{sec: appendix qualitative comparison}.

\section{Related Work}

\paragraph{One-step image generation.}
A major goal of recent generative modeling is to reduce sampling cost while preserving image quality.
GANs~\cite{GAN,biggan,sg2,sgxl,gigagan,gat} naturally provide one-step generation, but their scalability has often lagged behind diffusion and autoregressive models.
Recent diffusion and flow-based methods~\cite{consistencymodel,ict,shortcutmodel,meanflow,improvedmeanflow} also pursue few-step or one-step generation by learning direct or averaged transport trajectories.
While these methods improve inference efficiency, they often require long training schedules or specialized objectives.
In contrast, our work revisits one-step generation from the adversarial learning perspective and shows that properly organized hierarchical supervision can make transformer-based GANs efficient and competitive.

\paragraph{Multi-scale GANs.}
Multi-scale supervision has been widely used in GANs, from progressive growing~\cite{progan} to intermediate-output supervision~\cite{msg-gan,anycostgan,gigagan,gat}.
Recent scalable GANs commonly apply adversarial losses to intermediate generator outputs at multiple resolutions.
Although this design is often interpreted as hierarchical or coarse-to-fine generation, we show that scale-wise realism alone does not ensure that intermediate outputs follow a coherent refinement trajectory.
Our method preserves direct scale-wise adversarial feedback, while adding generator-side consistency regularization to align outputs across scales.

\section{Limitations and broader impact}
CAT still relies on a manually specified scale hierarchy, such as the number of stages and scale resolutions, and more adaptive scale selection remains an important direction. 
Due to computational constraints, we also provide only a limited study of discriminator scaling, leaving a systematic analysis of generator-discriminator capacity balance to future work. 
Like other generative models, CAT may inherit dataset biases and could be misused to create misleading content, requiring careful evaluation and safeguards.

\section{Conclusion}
We studied whether standard scale-wise adversarial supervision constructs a proper coarse-to-fine hierarchy in multi-stage GANs.
Our analysis shows that, although scale-wise supervision provides direct realism feedback at each resolution, intermediate outputs can remain misaligned with the final image, exhibiting large discrepancy, large inter-stage rewriting, and weak refinement-direction alignment.
To address this cross-scale trajectory misalignment, we proposed CAT, which preserves scale-wise discriminator feedback while introducing generator-side consistency regularization to align intermediate outputs.
On ImageNet-256, CAT-H/2 achieves an FID-50K of 1.56 with single-step inference and 60 training epochs, setting a new state of the art among one-step GAN and diffusion/flow models.
These results suggest that generator-side cross-scale alignment is an effective principle for scaling transformer-based GANs.

{
    \bibliographystyle{abbrv}
    \bibliography{reference}

@inproceedings{repa,
  title={Representation Alignment for Generation: Training Diffusion Transformers Is Easier Than You Think},
  author={Yu, Sihyun and Kwak, Sangkyung and Jang, Huiwon and Jeong, Jongheon and Huang, Jonathan and Shin, Jinwoo and Xie, Saining},
  booktitle={The Thirteenth International Conference on Learning Representations}
}

@article{ddpm,
  title={Denoising diffusion probabilistic models},
  author={Ho, Jonathan and Jain, Ajay and Abbeel, Pieter},
  journal={Advances in neural information processing systems},
  volume={33},
  pages={6840--6851},
  year={2020}
}

@article{flowmatching,
  title={Flow matching for generative modeling},
  author={Lipman, Yaron and Chen, Ricky TQ and Ben-Hamu, Heli and Nickel, Maximilian and Le, Matt},
  journal={arXiv preprint arXiv:2210.02747},
  year={2022}
}

@inproceedings{dit,
  title={Scalable diffusion models with transformers},
  author={Peebles, William and Xie, Saining},
  booktitle={Proceedings of the IEEE/CVF international conference on computer vision},
  pages={4195--4205},
  year={2023}
}

@inproceedings{LDM,
  title={High-resolution image synthesis with latent diffusion models},
  author={Rombach, Robin and Blattmann, Andreas and Lorenz, Dominik and Esser, Patrick and Ommer, Bj{\"o}rn},
  booktitle={Proceedings of the IEEE/CVF conference on computer vision and pattern recognition},
  pages={10684--10695},
  year={2022}
}

@inproceedings{vqgan,
  title={Taming transformers for high-resolution image synthesis},
  author={Esser, Patrick and Rombach, Robin and Ommer, Bjorn},
  booktitle={Proceedings of the IEEE/CVF conference on computer vision and pattern recognition},
  pages={12873--12883},
  year={2021}
}

@inproceedings{maskgit,
  title={Maskgit: Masked generative image transformer},
  author={Chang, Huiwen and Zhang, Han and Jiang, Lu and Liu, Ce and Freeman, William T},
  booktitle={Proceedings of the IEEE/CVF conference on computer vision and pattern recognition},
  pages={11315--11325},
  year={2022}
}

@article{var,
  title={Visual autoregressive modeling: Scalable image generation via next-scale prediction},
  author={Tian, Keyu and Jiang, Yi and Yuan, Zehuan and Peng, Bingyue and Wang, Liwei},
  journal={Advances in neural information processing systems},
  volume={37},
  pages={84839--84865},
  year={2024}
}

@article{mar,
  title={Autoregressive image generation without vector quantization},
  author={Li, Tianhong and Tian, Yonglong and Li, He and Deng, Mingyang and He, Kaiming},
  journal={Advances in Neural Information Processing Systems},
  volume={37},
  pages={56424--56445},
  year={2024}
}

@inproceedings{msg-gan,
  title={Msg-gan: Multi-scale gradients for generative adversarial networks},
  author={Karnewar, Animesh and Wang, Oliver},
  booktitle={Proceedings of the IEEE/CVF conference on computer vision and pattern recognition},
  pages={7799--7808},
  year={2020}
}

@article{progan,
  title={Progressive growing of gans for improved quality, stability, and variation},
  author={Karras, Tero and Aila, Timo and Laine, Samuli and Lehtinen, Jaakko},
  journal={arXiv preprint arXiv:1710.10196},
  year={2017}
}

@inproceedings{anycostgan,
  title={Anycost gans for interactive image synthesis and editing},
  author={Lin, Ji and Zhang, Richard and Ganz, Frieder and Han, Song and Zhu, Jun-Yan},
  booktitle={Proceedings of the IEEE/CVF conference on computer vision and pattern recognition},
  pages={14986--14996},
  year={2021}
}

@inproceedings{gigagan,
  title={Scaling up gans for text-to-image synthesis},
  author={Kang, Minguk and Zhu, Jun-Yan and Zhang, Richard and Park, Jaesik and Shechtman, Eli and Paris, Sylvain and Park, Taesung},
  booktitle={Proceedings of the IEEE/CVF conference on computer vision and pattern recognition},
  pages={10124--10134},
  year={2023}
}

@article{gat,
  title={Scalable GANs with Transformers},
  author={Hyun, Sangeek and Lee, MinKyu and Heo, Jae-Pil},
  journal={arXiv preprint arXiv:2509.24935},
  year={2025}
}

@inproceedings{aurora,
  title={Exploring sparse MoE in GANs for text-conditioned image synthesis},
  author={Zhu, Jiapeng and Yang, Ceyuan and Zheng, Kecheng and Xu, Yinghao and Shi, Zifan and Zhang, Yifei and Chen, Qifeng and Shen, Yujun},
  booktitle={Proceedings of the Computer Vision and Pattern Recognition Conference},
  pages={18411--18423},
  year={2025}
}

@article{GAN,
  title={Generative adversarial nets},
  author={Goodfellow, Ian J and Pouget-Abadie, Jean and Mirza, Mehdi and Xu, Bing and Warde-Farley, David and Ozair, Sherjil and Courville, Aaron and Bengio, Yoshua},
  journal={Advances in neural information processing systems},
  volume={27},
  year={2014}
}

@inproceedings{
ict,
title={Improved Techniques for Training Consistency Models},
author={Yang Song and Prafulla Dhariwal},
booktitle={The Twelfth International Conference on Learning Representations},
year={2024},
url={https://openreview.net/forum?id=WNzy9bRDvG}
}

@inproceedings{shortcutmodel,
  title={One Step Diffusion via Shortcut Models},
  author={Frans, Kevin and Hafner, Danijar and Levine, Sergey and Abbeel, Pieter},
  booktitle={The Thirteenth International Conference on Learning Representations}
}

@inproceedings{meanflow,
  title={Mean Flows for One-step Generative Modeling},
  author={Geng, Zhengyang and Deng, Mingyang and Bai, Xingjian and Kolter, J Zico and He, Kaiming},
  booktitle={The Thirty-ninth Annual Conference on Neural Information Processing Systems}
}

@inproceedings{
alphaflow,
title={AlphaFlow: Understanding and Improving MeanFlow Models},
author={Huijie Zhang and Aliaksandr Siarohin and Willi Menapace and Michael Vasilkovsky and Sergey Tulyakov and Qing Qu and Ivan Skorokhodov},
booktitle={The Fourteenth International Conference on Learning Representations},
year={2026},
url={https://openreview.net/forum?id=adacb4JTIv}
}

@inproceedings{
facm,
title={{FACM}: Flow-Anchored Consistency Models},
author={Yansong Peng and Kai Zhu and Yu Liu and Pingyu Wu and Hebei Li and Xiaoyan Sun and Feng Wu},
booktitle={The Fourteenth International Conference on Learning Representations},
year={2026},
url={https://openreview.net/forum?id=k9BpW1c4in}
}

@article{improvedmeanflow,
  title={Improved mean flows: On the challenges of fastforward generative models},
  author={Geng, Zhengyang and Lu, Yiyang and Wu, Zongze and Shechtman, Eli and Kolter, J Zico and He, Kaiming},
  journal={arXiv preprint arXiv:2512.02012},
  year={2025}
}

@inproceedings{sgxl,
  title={Stylegan-xl: Scaling stylegan to large diverse datasets},
  author={Sauer, Axel and Schwarz, Katja and Geiger, Andreas},
  booktitle={ACM SIGGRAPH 2022 conference proceedings},
  pages={1--10},
  year={2022}
}

@article{biggan,
  title={Large scale adversarial representation learning},
  author={Donahue, Jeff and Simonyan, Karen},
  journal={Advances in neural information processing systems},
  volume={32},
  year={2019}
}

@article{advflow,
  title={Adversarial flow models},
  author={Lin, Shanchuan and Yang, Ceyuan and Lin, Zhijie and Chen, Hao and Fan, Haoqi},
  journal={arXiv preprint arXiv:2511.22475},
  year={2025}
}

@inproceedings{
starflow,
title={{STARF}low: Scaling Latent Normalizing Flows for High-resolution Image Synthesis},
author={Jiatao Gu and Tianrong Chen and David Berthelot and Huangjie Zheng and Yuyang Wang and Ruixiang ZHANG and Laurent Dinh and Miguel {\'A}ngel Bautista and Joshua M. Susskind and Shuangfei Zhai},
booktitle={The Thirty-ninth Annual Conference on Neural Information Processing Systems},
year={2025},
url={https://openreview.net/forum?id=3YguS2rxdk}
}

@inproceedings{RAR,
  title={Randomized autoregressive visual generation},
  author={Yu, Qihang and He, Ju and Deng, Xueqing and Shen, Xiaohui and Chen, Liang-Chieh},
  booktitle={Proceedings of the IEEE/CVF International Conference on Computer Vision},
  pages={18431--18441},
  year={2025}
}

@inproceedings{xAR,
  title={Beyond next-token: Next-x prediction for autoregressive visual generation},
  author={Ren, Sucheng and Yu, Qihang and He, Ju and Shen, Xiaohui and Yuille, Alan and Chen, Liang-Chieh},
  booktitle={Proceedings of the IEEE/CVF International Conference on Computer Vision},
  pages={15781--15791},
  year={2025}
}

@inproceedings{simplediff,
  title={simple diffusion: End-to-end diffusion for high resolution images},
  author={Hoogeboom, Emiel and Heek, Jonathan and Salimans, Tim},
  booktitle={International Conference on Machine Learning},
  pages={13213--13232},
  year={2023},
  organization={PMLR}
}

@inproceedings{sit,
  title={Sit: Exploring flow and diffusion-based generative models with scalable interpolant transformers},
  author={Ma, Nanye and Goldstein, Mark and Albergo, Michael S and Boffi, Nicholas M and Vanden-Eijnden, Eric and Xie, Saining},
  booktitle={European Conference on Computer Vision},
  pages={23--40},
  year={2024},
  organization={Springer}
}

@inproceedings{lightningdit,
  title={Reconstruction vs. generation: Taming optimization dilemma in latent diffusion models},
  author={Yao, Jingfeng and Yang, Bin and Wang, Xinggang},
  booktitle={Proceedings of the Computer Vision and Pattern Recognition Conference},
  pages={15703--15712},
  year={2025}
}

@article{ddt,
  title={Ddt: Decoupled diffusion transformer},
  author={Wang, Shuai and Tian, Zhi and Huang, Weilin and Wang, Limin},
  journal={arXiv preprint arXiv:2504.05741},
  year={2025}
}

@inproceedings{
rae,
title={Diffusion Transformers with Representation Autoencoders},
author={Boyang Zheng and Nanye Ma and Shengbang Tong and Saining Xie},
booktitle={The Fourteenth International Conference on Learning Representations},
year={2026},
url={https://openreview.net/forum?id=0u1LigJaab}
}

@inproceedings{sg2,
  title={Analyzing and improving the image quality of stylegan},
  author={Karras, Tero and Laine, Samuli and Aittala, Miika and Hellsten, Janne and Lehtinen, Jaakko and Aila, Timo},
  booktitle={Proceedings of the IEEE/CVF conference on computer vision and pattern recognition},
  pages={8110--8119},
  year={2020}
}

@inproceedings{consistencymodel,
  title={Consistency models},
  author={Song, Yang and Dhariwal, Prafulla and Chen, Mark and Sutskever, Ilya},
  booktitle={Proceedings of the 40th International Conference on Machine Learning},
  pages={32211--32252},
  year={2023}
}

@inproceedings{rpgan,
  title={The relativistic discriminator: a key element missing from standard GAN},
  author={Jolicoeur-Martineau, Alexia},
  booktitle={International Conference on Learning Representations}
}

@inproceedings{seaweedapt,
  title={Diffusion Adversarial Post-Training for One-Step Video Generation},
  author={Lin, Shanchuan and Xia, Xin and Ren, Yuxi and Yang, Ceyuan and Xiao, Xuefeng and Jiang, Lu},
  booktitle={International Conference on Machine Learning},
  pages={37959--37974},
  year={2025},
  organization={PMLR}
}

@INPROCEEDINGS{imagenet,
  author={Deng, Jia and Dong, Wei and Socher, Richard and Li, Li-Jia and Kai Li and Li Fei-Fei},
  booktitle={2009 IEEE Conference on Computer Vision and Pattern Recognition}, 
  title={ImageNet: A large-scale hierarchical image database}, 
  year={2009},
  volume={},
  number={},
  pages={248-255},
  doi={10.1109/CVPR.2009.5206848}}

@article{fid,
  title={Gans trained by a two time-scale update rule converge to a local nash equilibrium},
  author={Heusel, Martin and Ramsauer, Hubert and Unterthiner, Thomas and Nessler, Bernhard and Hochreiter, Sepp},
  journal={Advances in neural information processing systems},
  volume={30},
  year={2017}
}

@article{rope,
  title={Roformer: Enhanced transformer with rotary position embedding},
  author={Su, Jianlin and Ahmed, Murtadha and Lu, Yu and Pan, Shengfeng and Bo, Wen and Liu, Yunfeng},
  journal={Neurocomputing},
  volume={568},
  pages={127063},
  year={2024},
  publisher={Elsevier}
}

@article{swigluffn,
  title={Glu variants improve transformer},
  author={Shazeer, Noam},
  journal={arXiv preprint arXiv:2002.05202},
  year={2020}
}

@article{dinov2,
  title={DINOv2: Learning Robust Visual Features without Supervision},
  author={Oquab, Maxime and Darcet, Timoth{\'e}e and Moutakanni, Th{\'e}o and Vo, Huy and Szafraniec, Marc and Khalidov, Vasil and Fernandez, Pierre and Haziza, Daniel and Massa, Francisco and El-Nouby, Alaaeldin and others},
  journal={Transactions on Machine Learning Research Journal},
  year={2024}
}

@article{pmf,
  title={One-step Latent-free Image Generation with Pixel Mean Flows},
  author={Lu, Yiyang and Lu, Susie and Sun, Qiao and Zhao, Hanhong and Jiang, Zhicheng and Wang, Xianbang and Li, Tianhong and Geng, Zhengyang and He, Kaiming},
  journal={arXiv preprint arXiv:2601.22158},
  year={2026}
}

@inproceedings{convnextv2,
  title={Convnext v2: Co-designing and scaling convnets with masked autoencoders},
  author={Woo, Sanghyun and Debnath, Shoubhik and Hu, Ronghang and Chen, Xinlei and Liu, Zhuang and Kweon, In So and Xie, Saining},
  booktitle={Proceedings of the IEEE/CVF conference on computer vision and pattern recognition},
  pages={16133--16142},
  year={2023}
}
}


\appendix
\newpage

\section{Appendix}

\subsection{Implementation details}
\label{sec: appendix implementation}

\begin{table*}[h]
\centering
\footnotesize
\caption{
Model configurations and training hyperparameters.
}
\label{tab:config}
\begin{minipage}[t]{0.60\linewidth}
\centering
\setlength{\tabcolsep}{6pt}
\begin{tabular}{@{}lccc@{}}
\toprule
configs & G-B/2 & G-M/2 & G-H/2 \\
\midrule
depth & $12$ & $24$ & $32$ \\
hidden dim & $768$ & $768$ & $1280$ \\
attention heads & $12$ & $12$ & $16$ \\
head dim & $64$ & $64$ & $80$ \\
patch size & $2{\times}2$ & $2{\times}2$ & $2{\times}2$ \\
MLP ratio & $4.0$ & $4.0$ & $4.0$ \\
\# outputs & $4$ & $4$ & $4$ \\
D input resolutions & \multicolumn{3}{c}{$(32,16,8,4)$} \\
output layers & ($3,6,9,12$) & ($6,12,18,24$) & ($8,16,24,32$) \\
\# $G$ spatial tokens & \multicolumn{3}{c}{$256$} \\
\# $D$ spatial tokens & \multicolumn{3}{c}{$256{+}64{+}16{+}4$} \\
\# $D$ cls tokens & \multicolumn{3}{c}{$4$} \\
G params & $133$M & $261$M & $960$M \\
D params & \multicolumn{3}{c}{$96$M} \\
total params & $229$M & $357$M & $1056$M \\
D depth / dim / heads & \multicolumn{3}{c}{$12$ / $768$ / $12$} \\
\bottomrule
\end{tabular}
\end{minipage}
\hfill
\begin{minipage}[t]{0.38\linewidth}
\centering
\setlength{\tabcolsep}{6pt}
\renewcommand{\arraystretch}{1.03}
\begin{tabular}{@{}lc@{}}
\toprule
hyperparameter & value \\
\midrule
image resolution & $256 \times 256$ \\
VAE & SD-VAE~\cite{LDM} \\
REPA encoder & DINOv2-ViT-B~\cite{dinov2} \\
$\lambda_{\mathrm{REPA}}$ & $1.0$ \\
consistency weight & $0.1$ \\
$R_1$ weight / interval & $1.0$ / $1$ \\
$R_2$ weight / interval & $1.0$ / $1$ \\
approx. GP $\epsilon$ & $0.01$ \\
\midrule
batch size & $512$ \\
optimizer & AdamW \\
Adam $(\beta_1,\beta_2)$ & $(0.0,0.99)$ \\
weight decay & $0.0$ \\
learning rate & $2 \times 10^{-4}$ \\
EMA decay & $0.999$ \\
mixed precision & bfloat16 \\
\bottomrule
\end{tabular}
\end{minipage}
\vspace{-2mm}
\end{table*}

\paragraph{Backbone and training details.}
We follow the implementation style of GAT~\cite{gat} and train all models in the SD-VAE latent space, where ImageNet-256 images are represented as $32 \times 32 \times 4$ latents. 
Both the generator and discriminator are implemented as ViT-style transformer networks. 
The generator starts from fixed 2D sinusoidal positional tokens and injects class and latent conditioning through a two-layer mapping network. 
Each generator block uses RMSNorm, multi-head self-attention with RoPE~\cite{rope} and qk-normalization, and a SwiGLU feed-forward network~\cite{swigluffn}. 
We use AdaLN-style modulation in the generator, where the style vector predicts scale, shift, and residual gate parameters for both the attention and MLP branches. 
The generator produces four accumulated outputs from uniformly spaced transformer blocks through output skip connections. 

Unless otherwise specified, the discriminator is fixed to the Base configuration. 
We use the same transformer components, including RoPE, qk-normalization, and SwiGLU feed-forward layers. 
For the projected discriminator objective, we use a frozen DINOv2-ViT-B encoder. 
All models are trained with AdamW, batch size 512, bfloat16 mixed precision, gradient clipping, and generator EMA for evaluation. 
We set the consistency weight to $\lambda_{\mathrm{cons}}=0.1$, apply $R_1$ and $R_2$ regularization at every iteration, and use an approximated gradient-penalty perturbation scale of $10^{-2}$.

The detailed model configurations and shared training hyperparameters are summarized in Table~\ref{tab:config}.

\paragraph{Construction of multi-scale discriminator inputs.}
A key implementation detail is that the generator outputs are produced at the same latent resolution.
Although we denote the hierarchical supervision by scale index $k$ in the main text, the transformer generator maintains a fixed token grid across depth and therefore does not naturally produce different spatial resolutions.
Let $h_k \in \mathbb{R}^{32 \times 32 \times 4}$ denote the $k$-th accumulated output produced by the generator.
We then construct the scale-specific discriminator input by resizing $h_k$ to the resolution associated with scale $k$:
\[
x_k = r_k(h_k),
\]
where $r_k(\cdot)$ denotes the resizing operator for the $k$-th scale.
The real latent is resized in the same way to form the matching real input at each scale.
Thus, the generator always synthesizes same-resolution latent outputs, while the multi-scale hierarchy used for adversarial supervision is constructed only when forming the discriminator inputs.
This lets us control cross-scale interaction in the discriminator without changing the fixed-resolution transformer synthesis path.

\paragraph{Overall pipeline.}
\begin{figure}
    \centering
    \includegraphics[width=1.0\linewidth]{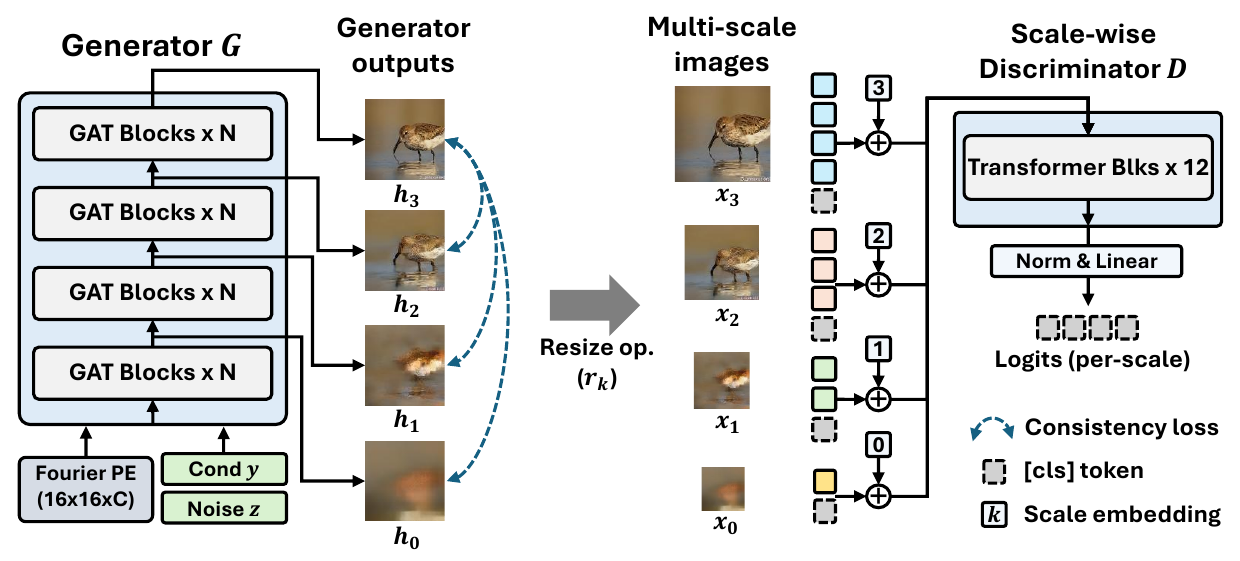}
    \caption{
    Overall pipeline of CAT.
    The generator produces accumulated intermediate outputs $\{h_k\}_{k=0}^{3}$ from uniformly spaced transformer blocks.
    Since the transformer generator operates on a fixed token grid, all $h_k$ are synthesized at the same resolution.
    We construct multi-scale discriminator inputs by resizing each output with $r_k$, yielding $x_k = r_k(h_k)$.
    The discriminator receives the resulting multi-scale images with scale embeddings and per-scale [cls] tokens, while a block-diagonal attention mask enforces scale-wise discrimination without cross-scale token exchange.
    A generator-side consistency loss aligns the intermediate outputs, while the discriminator provides direct scale-wise adversarial feedback.
    }
    \label{fig: appendix architectural details}
\end{figure}
Fig.~\ref{fig: appendix architectural details} illustrates the overall pipeline of CAT.
The generator $G$ is divided into multiple stages, and each stage produces an accumulated output $h_k$ through an output skip connection.
Because the generator is transformer-based and maintains a fixed token grid, these intermediate outputs are generated at the identical resolution rather than at progressively different resolutions.
To construct hierarchical adversarial supervision, we resize each $h_k$ with a scale-specific resizing operator $r_k$ and obtain the discriminator input
\[
x_k = r_k(h_k).
\]
The real latent is resized in the same way for the corresponding scale.

The discriminator $D$ then receives the resulting multi-scale images $\{x_k\}$.
For each scale, we append a separate [cls] token and add a learnable scale embedding.
Although tokens from all scales are processed in a single discriminator for implementation efficiency, cross-scale attention is blocked with a block-diagonal attention mask, yielding a scale-wise discriminator in which each scale is discriminated using only same-scale tokens.
This provides clean scale-wise adversarial feedback to the generator.
At the same time, CAT applies a generator-side consistency loss to the intermediate outputs $\{h_k\}$, encouraging them to remain aligned with the final output.
As a result, CAT combines scale-wise discrimination with explicit cross-scale alignment in the generator.

\subsection{Details of GFLOPs computation}
\label{sec: appendix gflops}
We provide details on how the training and inference compute in Table~\ref{tab:compute_main} is estimated.
All numbers are analytical GFLOPs estimates.
Inference compute is measured per generated sample.
For GAN-based models, inference corresponds to a single generator forward pass, while for iMF it corresponds to one evaluation-network path.
Training compute is measured per sample per training iteration and includes both forward and backward computation.
For training estimates, we follow the standard analytical convention that each trainable forward-backward computation costs approximately $3F$, while branches blocked by stop-gradient are counted only by their forward cost.

\paragraph{Training computation cost.}
\begin{table*}[t]
\centering
\small
\caption{
Summary of GFLOPs estimation.
Training compute is reported per sample per iteration, and inference compute is reported per generated sample.
All $F$ terms denote one forward-pass GFLOPs per sample.
For training estimates, we follow the standard analytical convention that each trainable forward-backward computation costs approximately $3F$, while branches blocked by stop-gradient are counted only by their forward cost.
}
\label{tab:gflops_estimation_details}
\setlength{\tabcolsep}{4pt}
\begin{tabular}{@{}lclcc@{}}
\toprule
Method & Forward GFLOPs & Training-step approximation & Train GFLOPs & Infer. GFLOPs \\
       & /sample        &                             & /sample and iter & /sample \\
\midrule
iMF-XL/2
& $F_v{=}F_u{=}174,\ F_{uv}{=}203$
& $3F_v + F_u + 3F_{uv}$
& $1{,}306.3$
& $174.6$ \\
GAT-XL/2
& $F_G{=}119,\ F_D{=}122$
& $4F_G + 15F_D$
& $2{,}297.2$
& $118.6$ \\
CAT-H/2
& $F_G{=}167,\ F_D{=}36$
& $4F_G + 10.5F_D$
& $1{,}040.2$
& $166.7$ \\
\bottomrule
\end{tabular}
\vspace{-2mm}
\end{table*}

For GAN-based models, one training iteration consists of a discriminator step and a generator step.
In the discriminator step, the generator first synthesizes fake samples without gradient update.
The discriminator is then trained with a relativistic adversarial loss~\cite{rpgan} computed from real and fake logits, together with the approximated gradient-penalty terms.
For CAT, we compute the approximated gradient penalty~\cite{seaweedapt} using only a quarter of the batch, which reduces the discriminator-side training cost compared with a full-batch gradient-penalty computation.
In the generator step, the generator synthesizes fake samples with gradients enabled, and the discriminator is used to provide the real/fake logits for the relativistic generator loss while its parameters are kept frozen.
Thus, the discriminator still needs to propagate gradients to the generated samples, but not to its own parameters.

Let $F_G$ and $F_D$ denote one generator and discriminator forward pass, respectively.
Under the above accounting rule, a CAT training iteration is approximated as
\[
4F_G + 10.5F_D.
\]
The generator term accounts for the no-gradient fake generation in the discriminator step and the trainable generator forward-backward computation in the generator step.
The discriminator term accounts for adversarial real/fake discrimination, discriminator-side gradient penalties, and the frozen discriminator evaluation used for generator training.
In the final reported numbers, we also include the small additional costs from discriminator-side auxiliary feature prediction and the frozen DINOv2 forward pass used for representation alignment.

The discriminator forward cost $F_D$ of CAT-H/2 is much smaller than that of GAT-XL/2 because CAT keeps the discriminator at the \textit{Base} scale while scaling the generator.
This design is enabled by our scale-wise discriminator and generator-side consistency regularization: the discriminator can provide clean scale-wise adversarial feedback without requiring large cross-scale capacity, while consistency regularization maintains alignment across generated scales.
As a result, CAT obtains strong scalability and generation quality with a relatively lightweight discriminator.

For GAT-XL/2, which scales both the generator and discriminator and uses a full-batch gradient-penalty setting, the corresponding estimate becomes
\[
4F_G + 15F_D.
\]
This leads to a substantially larger per-iteration training cost, despite its lower generator-only inference cost.

For iMF-XL/2, we use the code from official implementation\footnote{https://github.com/Lyy-iiis/imeanflow} that contains velocity-guidance and meanflow training paths.
During training, iMF first evaluates a velocity network for classifier-free guidance.
This guidance branch consists of one doubled conditional/unconditional velocity evaluation and one additional conditional velocity evaluation, giving $3F_v$ in total, where $F_v$ denotes one velocity-network forward pass.
Because the guided velocity is blocked by stop-gradient before being used in the target, these guidance evaluations are counted as forward-only.

The average velocity branch then applies a JVP through the $u$ network.
Let $F_u$ denote the forward cost of the $u$ network used in this tangent path, and let $F_{uv}$ denote the trainable joint path that returns the primal mean-flow output $u$ together with the auxiliary velocity output $v$.
The JVP tangent term is also stopped in the target construction, so this tangent path is counted as a forward-only $F_u$ term.
Reverse-mode gradients are counted only for the trainable primal $u/v$ path, which is approximated as $3F_{uv}$.
Thus, the iMF-XL/2 training step is estimated as
\[
3F_v + F_u + 3F_{uv}.
\]
The resulting call models and compute estimates are summarized in Table~\ref{tab:gflops_estimation_details}.

\subsection{Preliminary Pixel-space Experiment}
\label{app:pixel_results}

We further conduct a preliminary pixel-space experiment to examine whether the proposed supervision strategy is applicable beyond latent-space training.
For this experiment, we use the same model configuration as our main setting, namely G-B/2 and D-B/2, while adapting the patch size for pixel-space training.
We use a resolution hierarchy of $[16, 32, 64, 256]$ and keep the same token grid resolutions as in our latent-space model, i.e., $[2, 4, 8, 16]$.
Accordingly, the patch sizes are set to $[8, 8, 8, 16]$ for the four resolutions, respectively.
Following pMF~\cite{pmf}, we additionally leverage a pretrained encoder, implemented as a ConvNeXtV2~\cite{convnextv2}-based vision-aided loss.

Table~\ref{tab:pixel_results} reports the results.
Our method obtains an FID-50K of 3.54 after 40 epochs, which is comparable to the pMF baseline trained for 160 epochs.
Although this experiment is less extensively tuned than our main latent-space setting, the result suggests that our method is compatible with pixel-space training and large-patch image-space generation.
We leave more extensive tuning and a fully controlled pixel-space comparison to future work.

\begin{table}[t]
\centering
\caption{
\textbf{Preliminary pixel-space results.}
We evaluate our method in pixel space using the same G-B/2 and D-B/2 configuration as the main setting, with an adapted patch size and a resolution hierarchy of $[16, 32, 64, 256]$.
Following pMF, we additionally use a ConvNeXtV2-based vision-aided loss.
}
\label{tab:pixel_results}
\begin{tabular}{lcc}
\toprule
Method & Training epochs & FID-50K \\
\midrule
pMF-B/2~\cite{pmf} & 160 & 3.53 \\
CAT-B/2 & 40 & 3.54 \\
\bottomrule
\end{tabular}
\end{table}

\newpage
\subsection{Qualitative comparison with iMF.}
\label{sec: appendix qualitative comparison}
We further compare the proposed CAT with iMF~\cite{improvedmeanflow} using uncurated ImageNet-256 samples as belows.
For sampling parameter, we use truncation $\psi=0.85$ for CAT-H/2, while we follow the settings of iMF's the official implementation (interval \([0.42, 0.62]\) and \(\omega=8.0\)).
\begin{figure}[h]
\centering
\small

\begin{subfigure}[t]{0.49\linewidth}
    \centering
    \includegraphics[width=\linewidth]{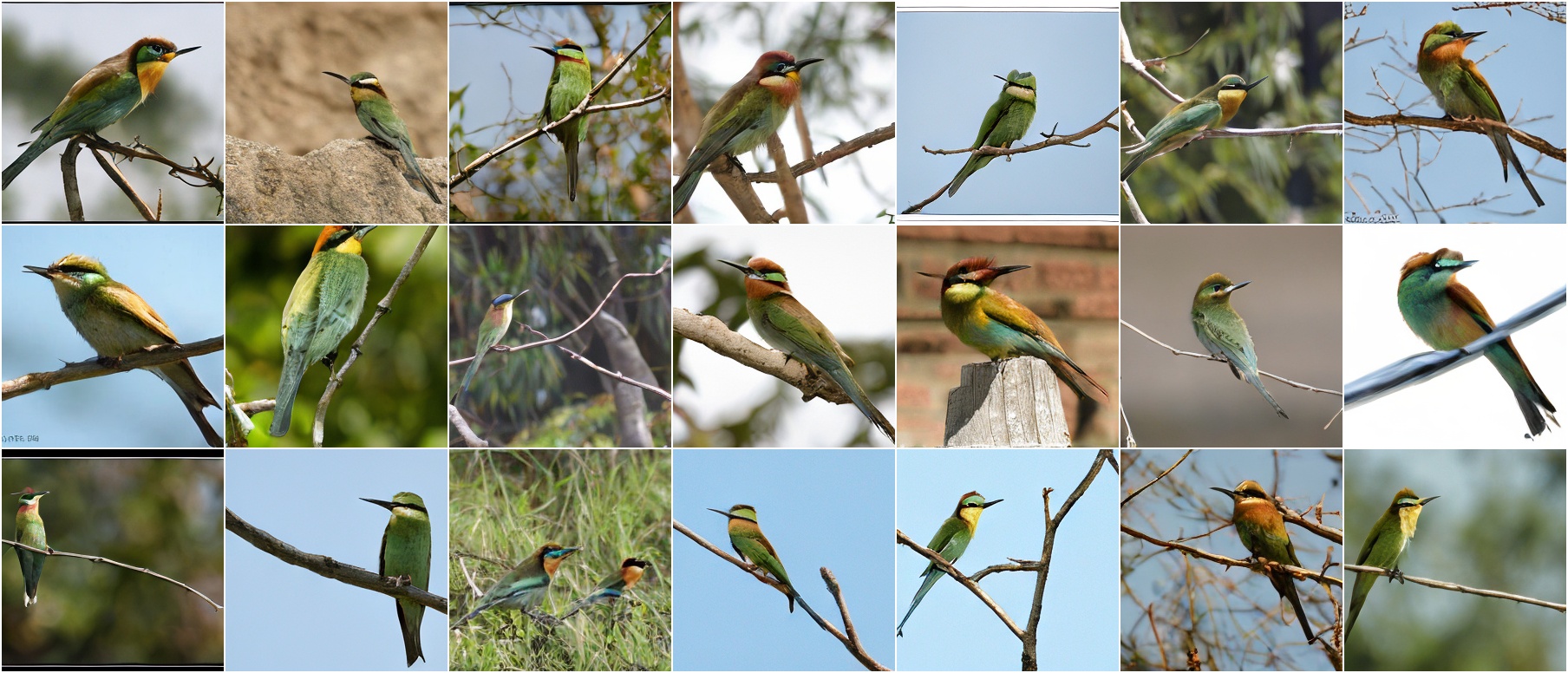}
    \vspace{-5mm}
    \caption{CAT-H/2 (Ours): class 277 (bee eater)}
\end{subfigure}
\hfill
\begin{subfigure}[t]{0.49\linewidth}
    \centering
    \includegraphics[width=\linewidth]{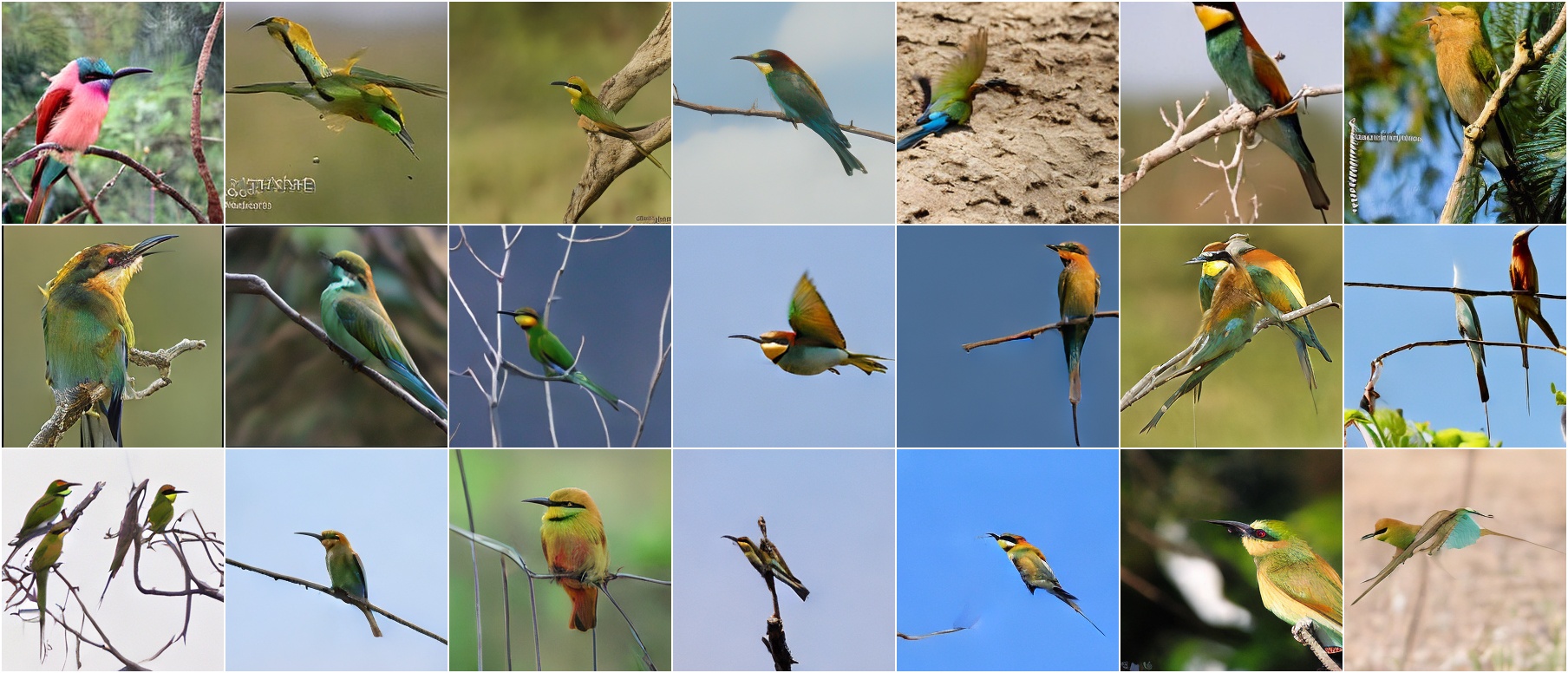}
    \vspace{-5mm}
    \caption{iMF-XL/2: class 277 (bee eater)}
\end{subfigure}


\begin{subfigure}[t]{0.49\linewidth}
    \centering
    \includegraphics[width=\linewidth]{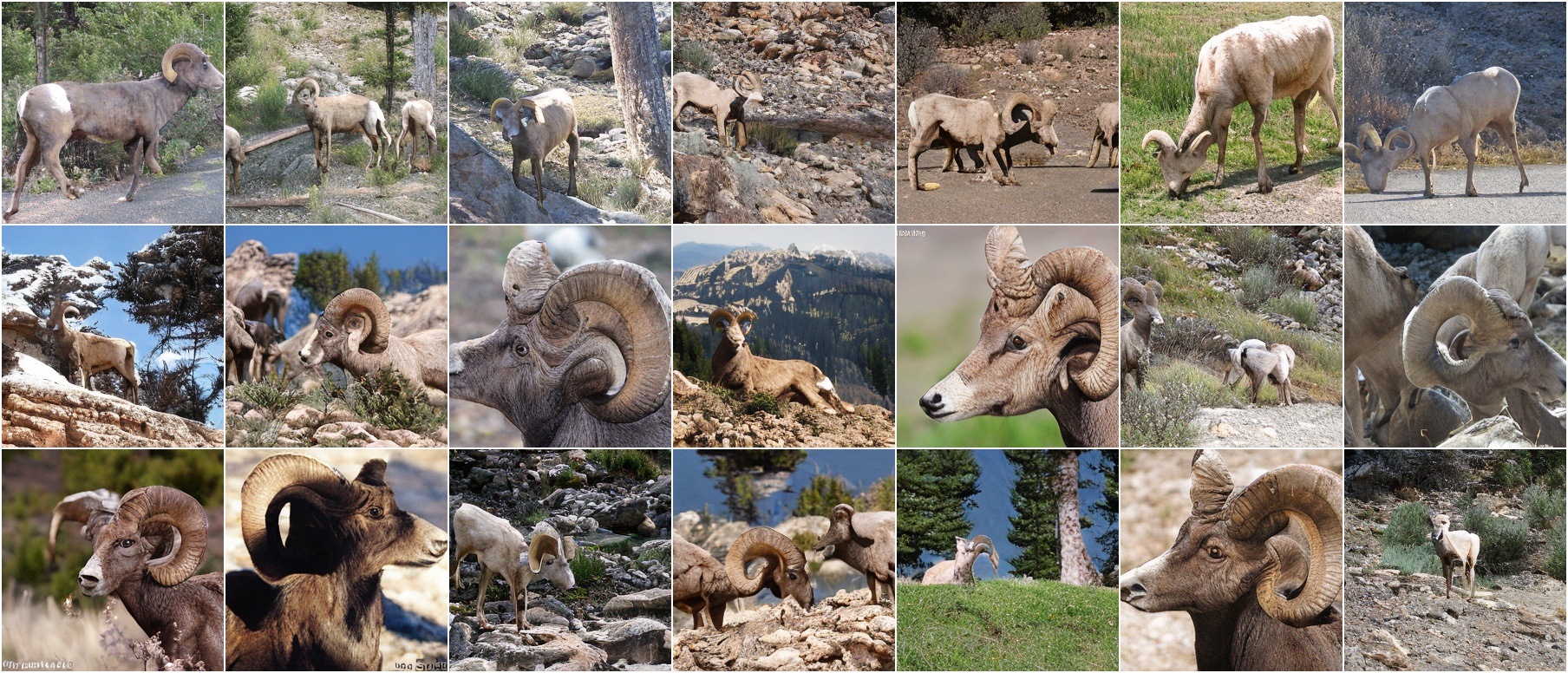}
    \vspace{-5mm}
    \caption{CAT-H/2 (Ours): class 349 (bighorn)}
\end{subfigure}
\hfill
\begin{subfigure}[t]{0.49\linewidth}
    \centering
    \includegraphics[width=\linewidth]{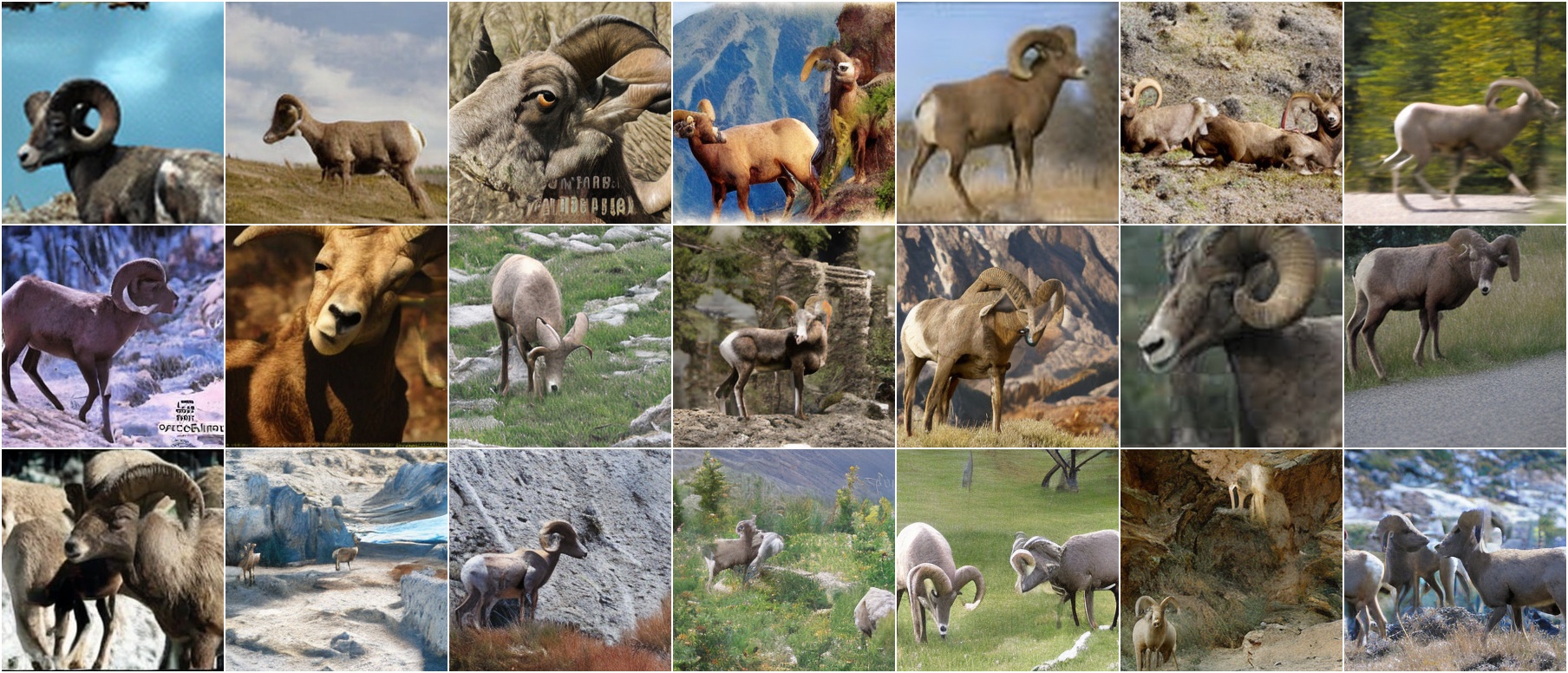}
    \vspace{-5mm}
    \caption{iMF-XL/2: class 349 (bighorn)}
\end{subfigure}


\begin{subfigure}[t]{0.49\linewidth}
    \centering
    \includegraphics[width=\linewidth]{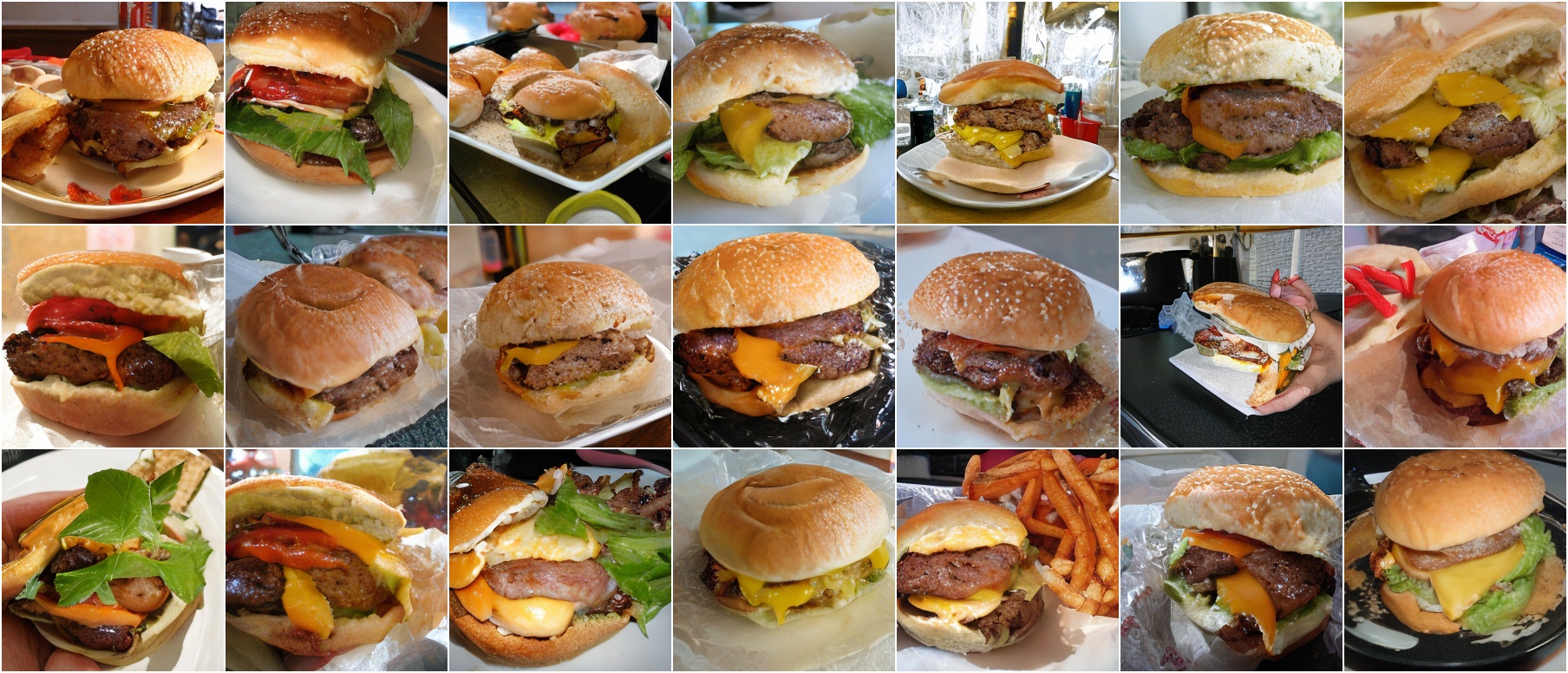}
    \vspace{-5mm}
    \caption{CAT-H/2 (Ours): class 933 (cheeseburger)}
\end{subfigure}
\hfill
\begin{subfigure}[t]{0.49\linewidth}
    \centering
    \includegraphics[width=\linewidth]{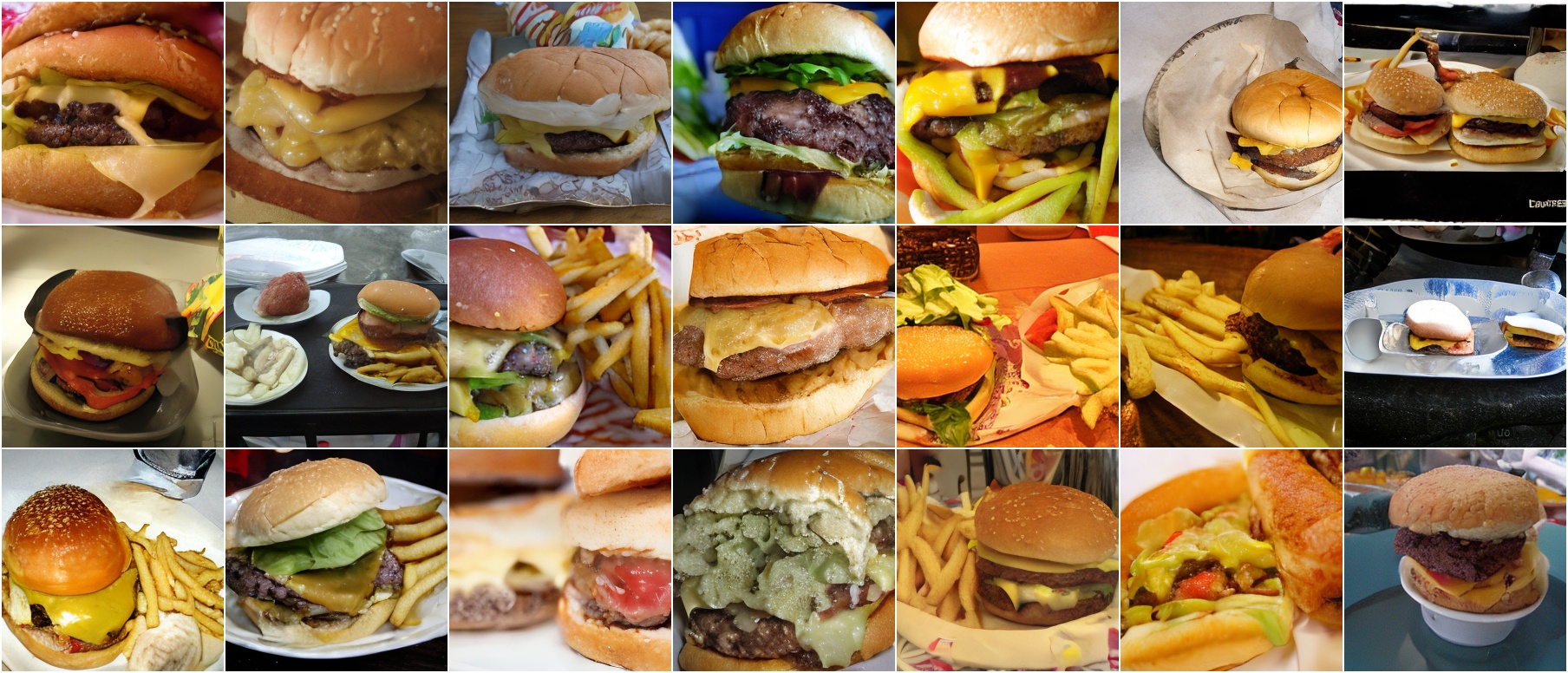}
    \vspace{-5mm}
    \caption{iMF-XL/2: class 933 (cheeseburger)}
\end{subfigure}


\begin{subfigure}[t]{0.49\linewidth}
    \centering
    \includegraphics[width=\linewidth]{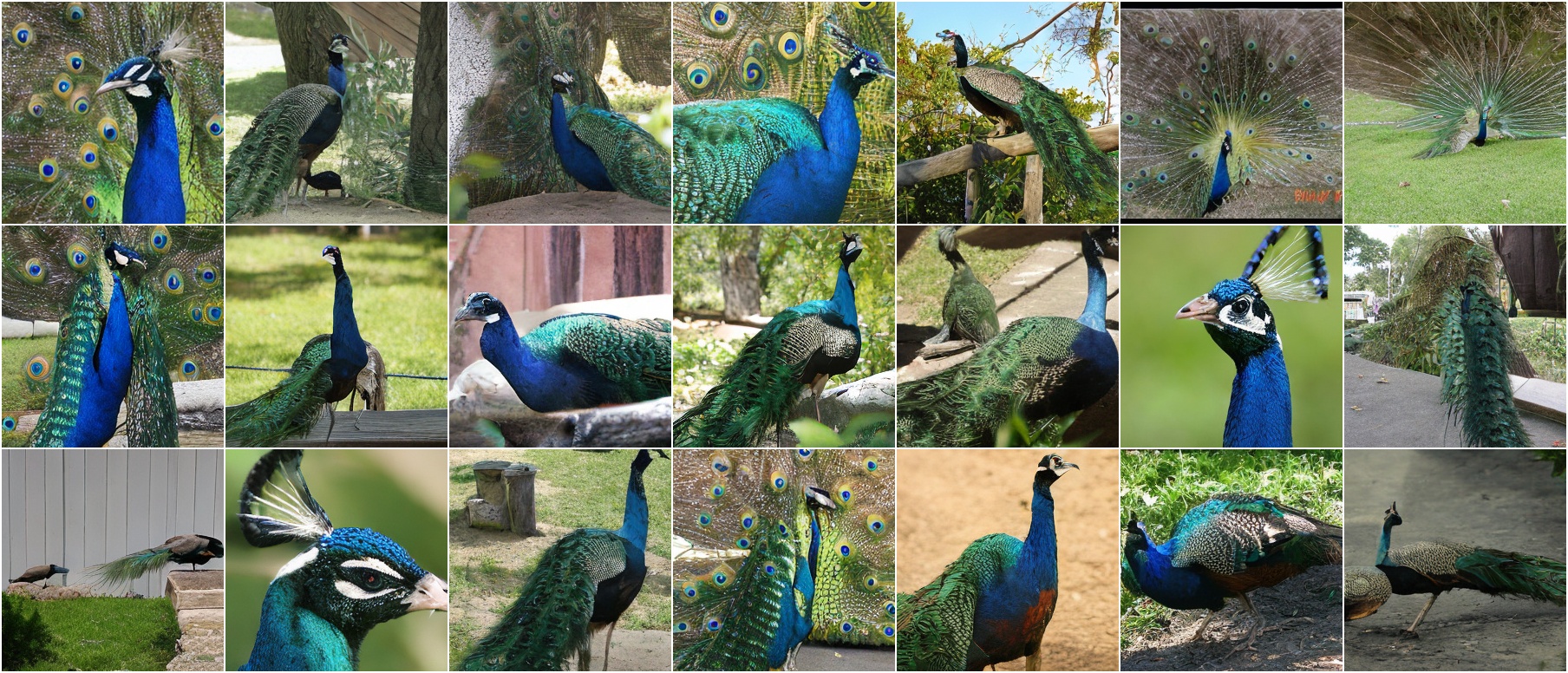}
    \vspace{-5mm}
    \caption{CAT-H/2 (Ours): class 84 (peacock)}
\end{subfigure}
\hfill
\begin{subfigure}[t]{0.49\linewidth}
    \centering
    \includegraphics[width=\linewidth]{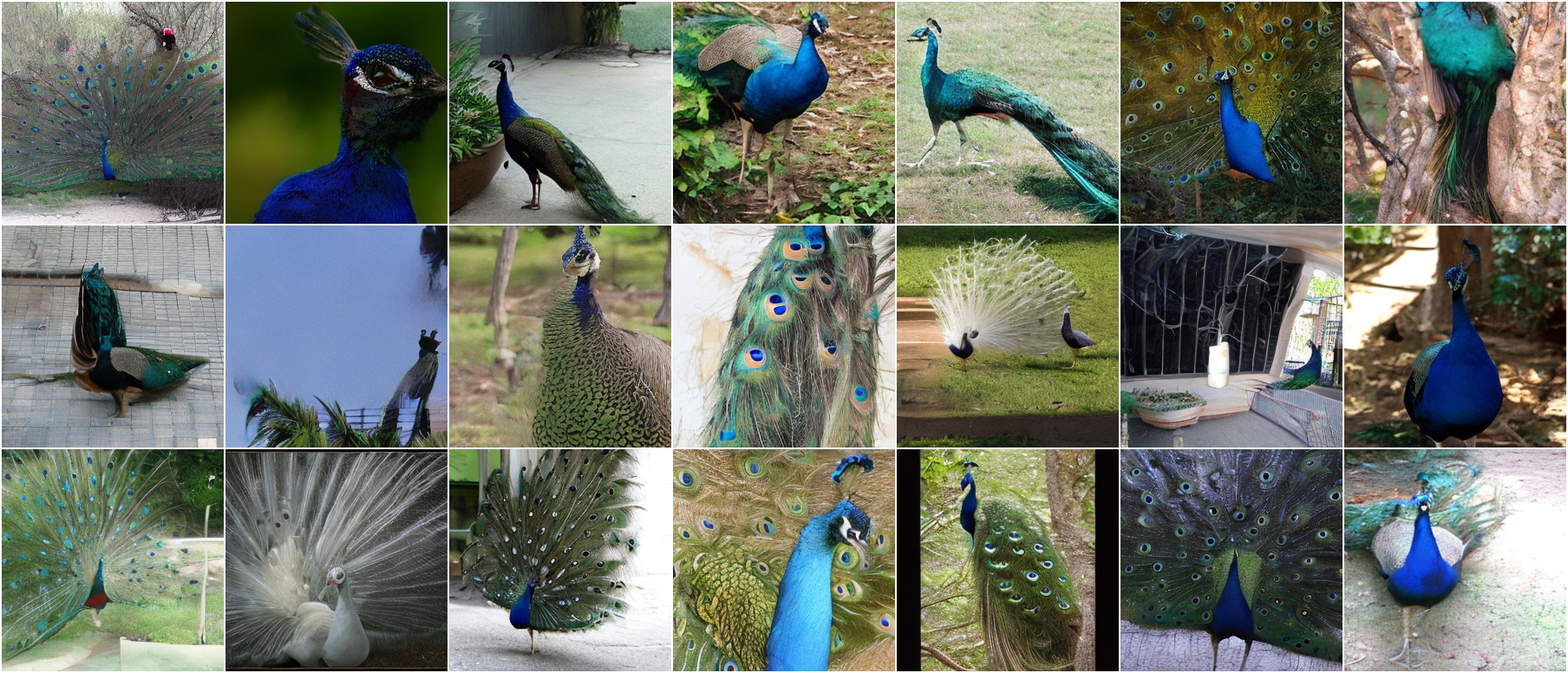}
    \vspace{-5mm}
    \caption{iMF-XL/2: class 84 (peacock)}
\end{subfigure}

\begin{subfigure}[t]{0.49\linewidth}
    \centering
    \includegraphics[width=\linewidth]{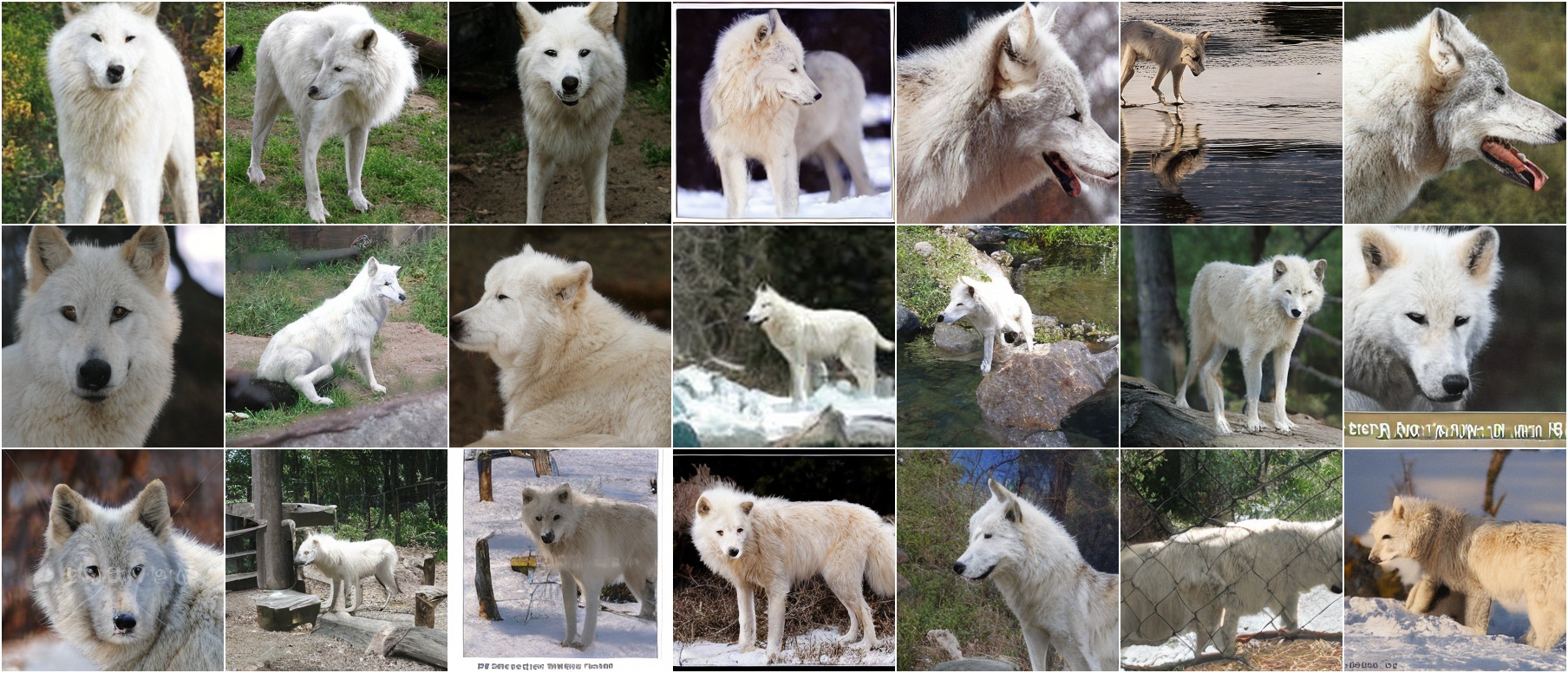}
    \vspace{-5mm}
    \caption{CAT-H/2 (Ours): class 270 (white wolf)}
\end{subfigure}
\hfill
\begin{subfigure}[t]{0.49\linewidth}
    \centering
    \includegraphics[width=\linewidth]{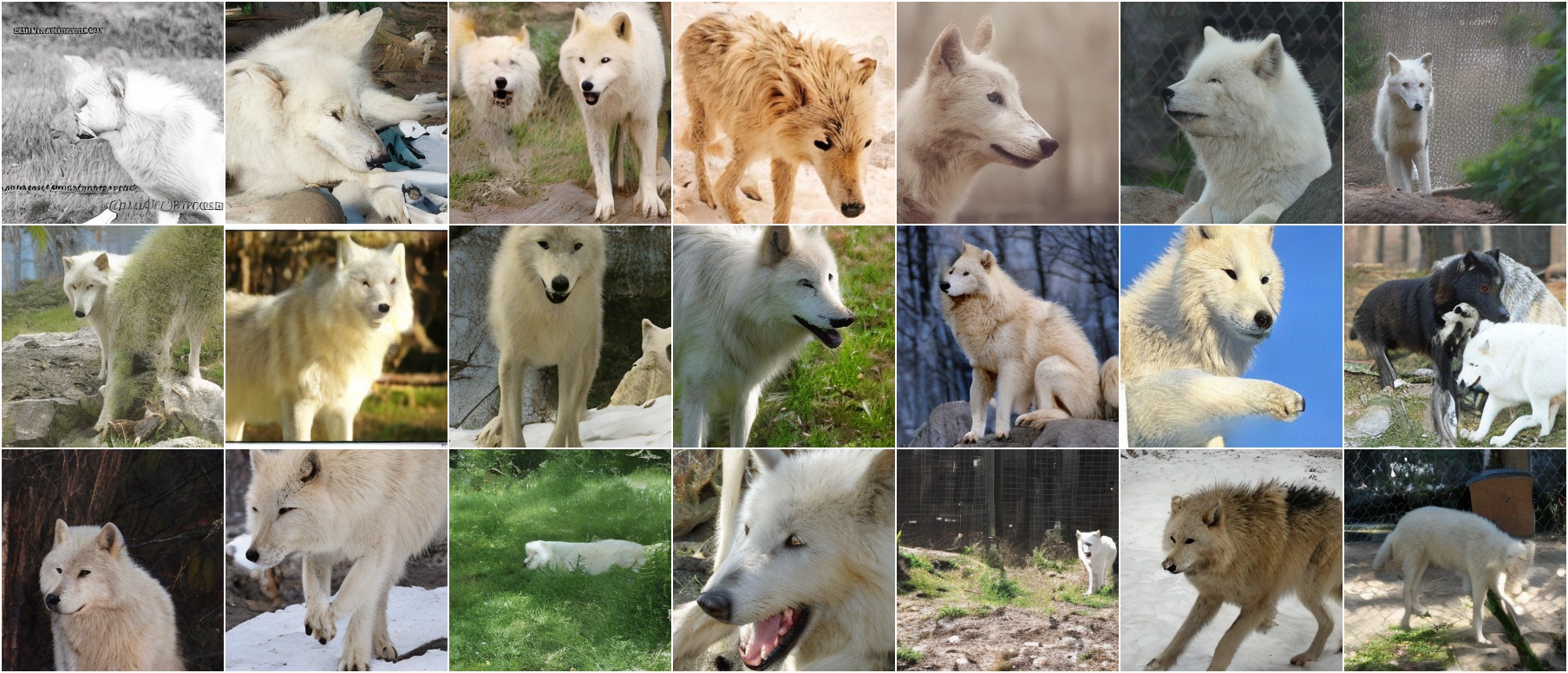}
    \vspace{-5mm}
    \caption{iMF-XL/2: class 270 (white wolf)}
\end{subfigure}

    \caption{
    \textbf{Uncurated sample comparison with iMF.}
    We compare randomly generated ImageNet-256 samples from our model (left) and iMF-XL/2 (right), without manual curation.
    Our model is trained for 60 epochs and uses 166.7 GFLOPs for one-step inference, whereas iMF-XL/2 is trained for 800 epochs and uses 174.6 GFLOPs.
    For sampling, our model uses truncation \(\psi=0.85\), and iMF-XL/2 follows the official configuration with interval \([0.42, 0.62]\) and \(\omega=8.0\).
    }
\label{fig:appendix qualitative_grid 1}
\vspace{-3mm}
\end{figure}

\begin{figure}[H]
\centering
\small

\begin{subfigure}[t]{0.49\linewidth}
    \centering
    \includegraphics[width=\linewidth]{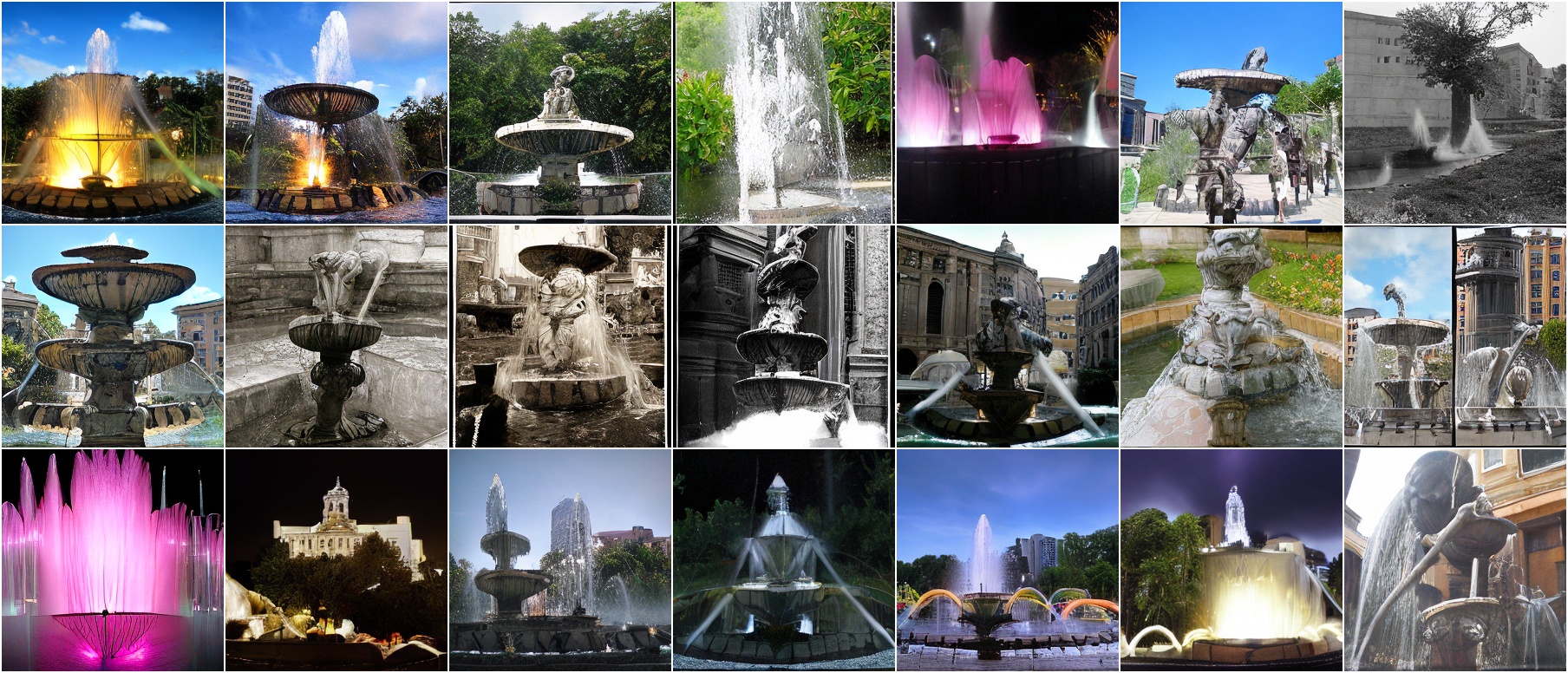}
    \vspace{-5mm}
    \caption{CAT-H/2 (Ours): class 562 (fountain)}
\end{subfigure}
\hfill
\begin{subfigure}[t]{0.49\linewidth}
    \centering
    \includegraphics[width=\linewidth]{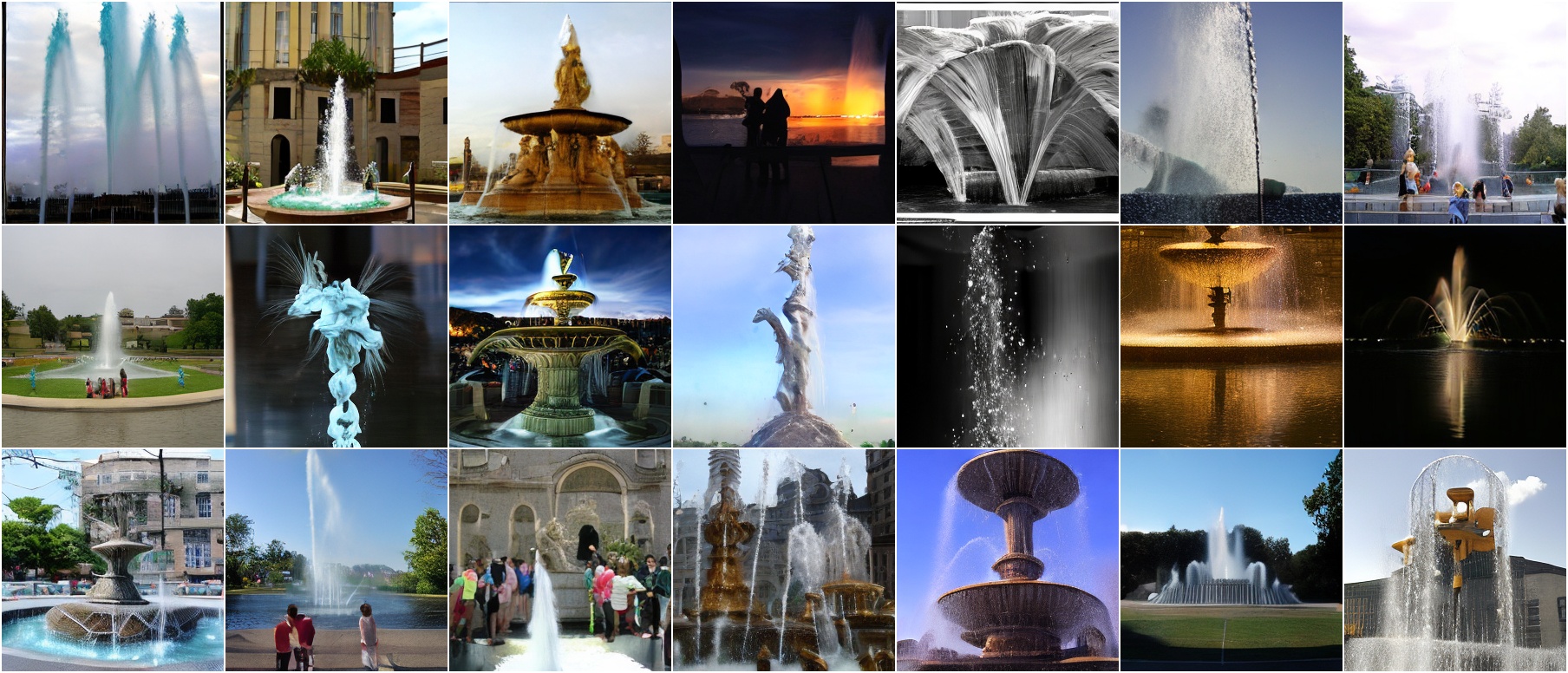}
    \vspace{-5mm}
    \caption{iMF-XL/2: class 562 (fountain)}
\end{subfigure}


\begin{subfigure}[t]{0.49\linewidth}
    \centering
    \includegraphics[width=\linewidth]{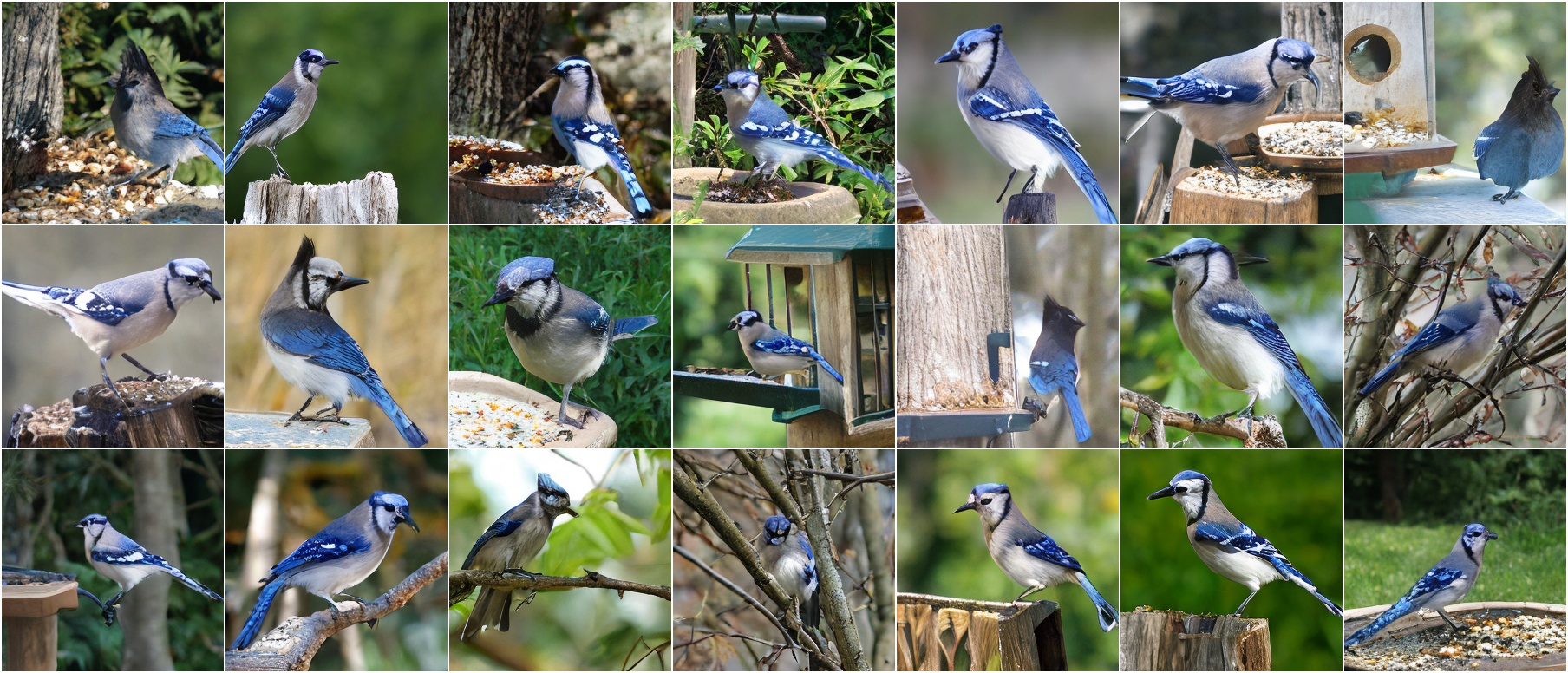}
    \vspace{-5mm}
    \caption{CAT-H/2 (Ours): class 17 (jay)}
\end{subfigure}
\hfill
\begin{subfigure}[t]{0.49\linewidth}
    \centering
    \includegraphics[width=\linewidth]{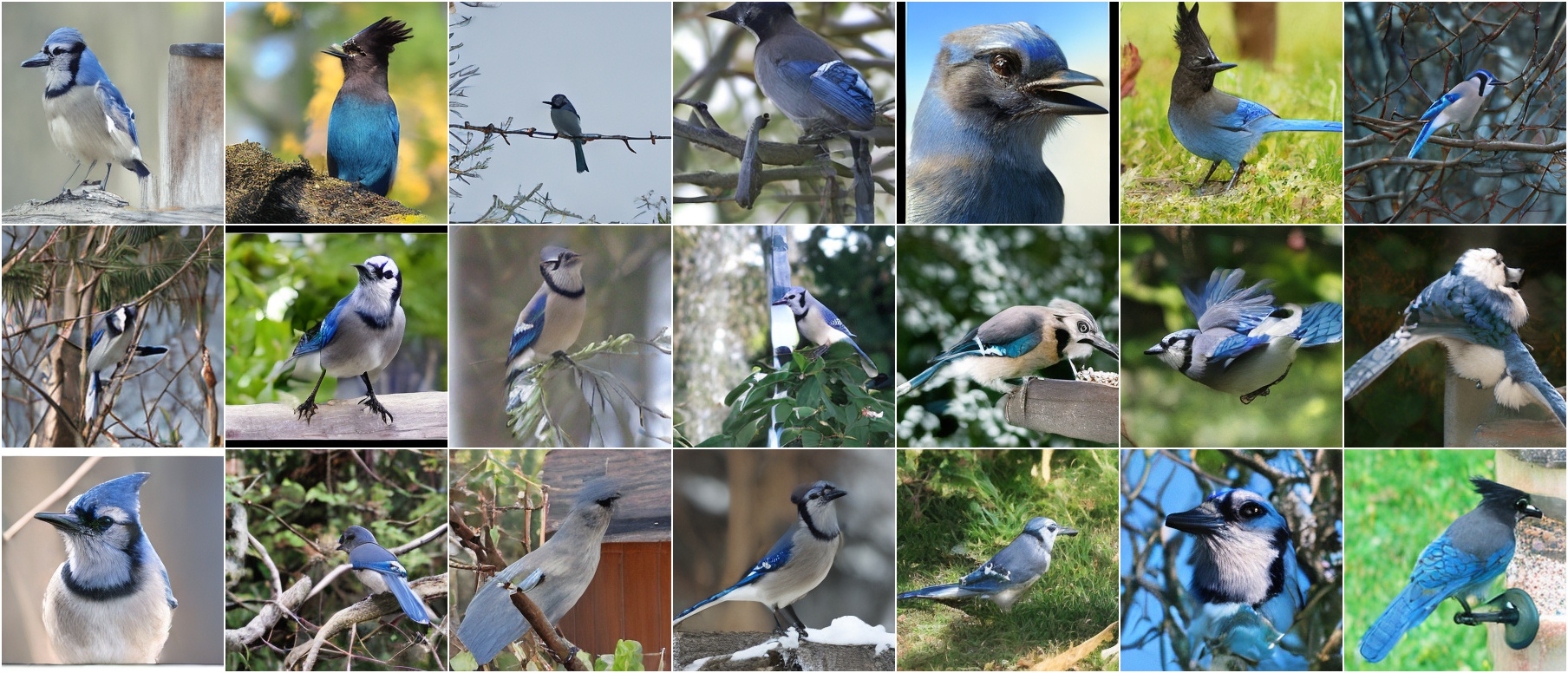}
    \vspace{-5mm}
    \caption{iMF-XL/2: class 17 (jay)}
\end{subfigure}


\begin{subfigure}[t]{0.49\linewidth}
    \centering
    \includegraphics[width=\linewidth]{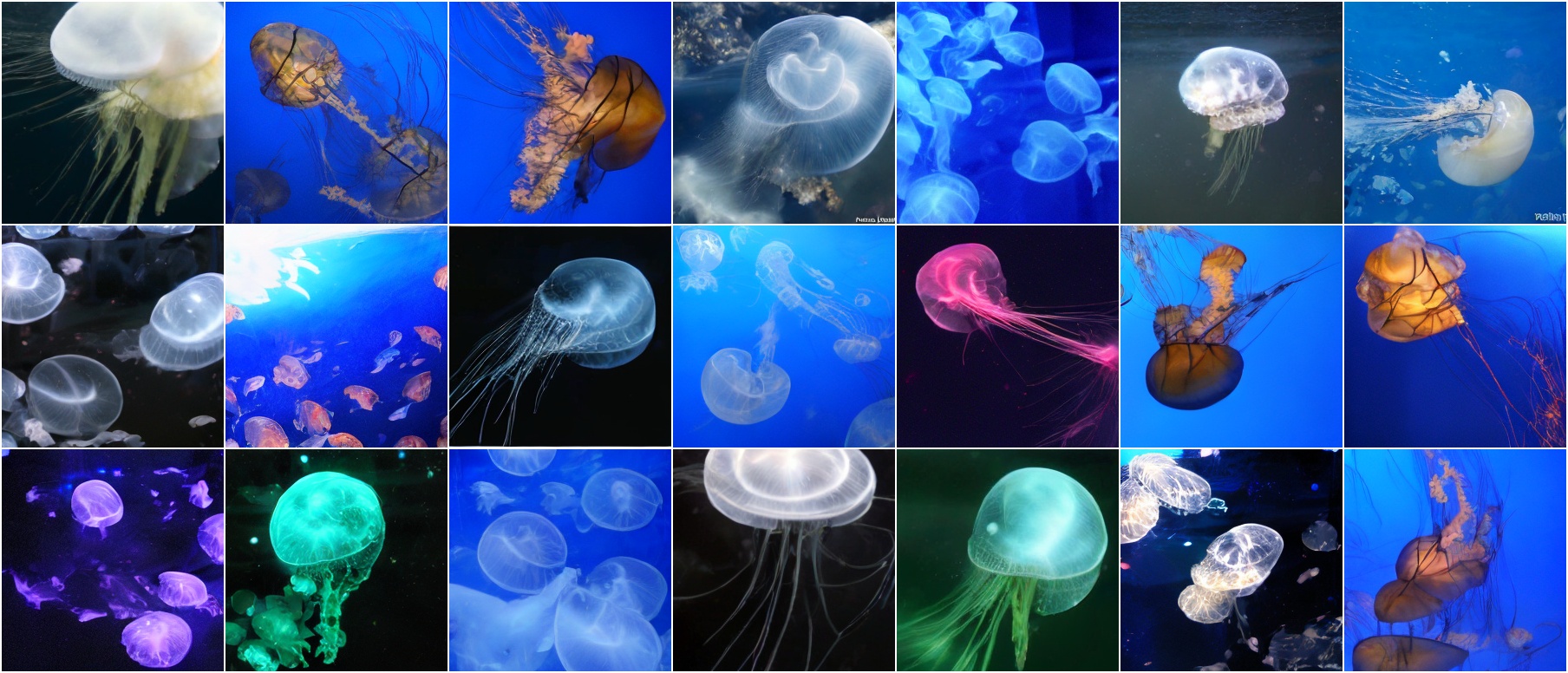}
    \vspace{-5mm}
    \caption{CAT-H/2 (Ours): class 107 (jellyfish)}
\end{subfigure}
\hfill
\begin{subfigure}[t]{0.49\linewidth}
    \centering
    \includegraphics[width=\linewidth]{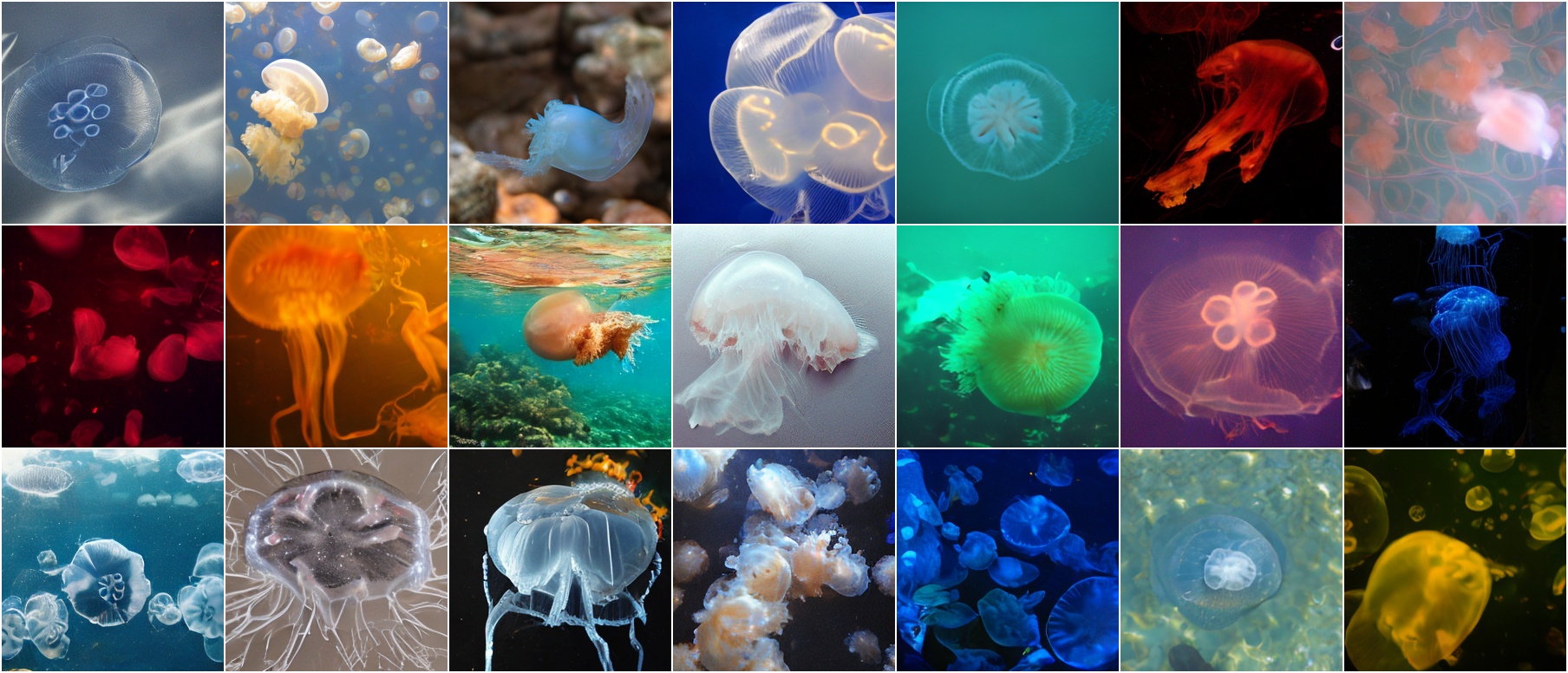}
    \vspace{-5mm}
    \caption{iMF-XL/2: class 107 (jellyfish)}
\end{subfigure}


\begin{subfigure}[t]{0.49\linewidth}
    \centering
    \includegraphics[width=\linewidth]{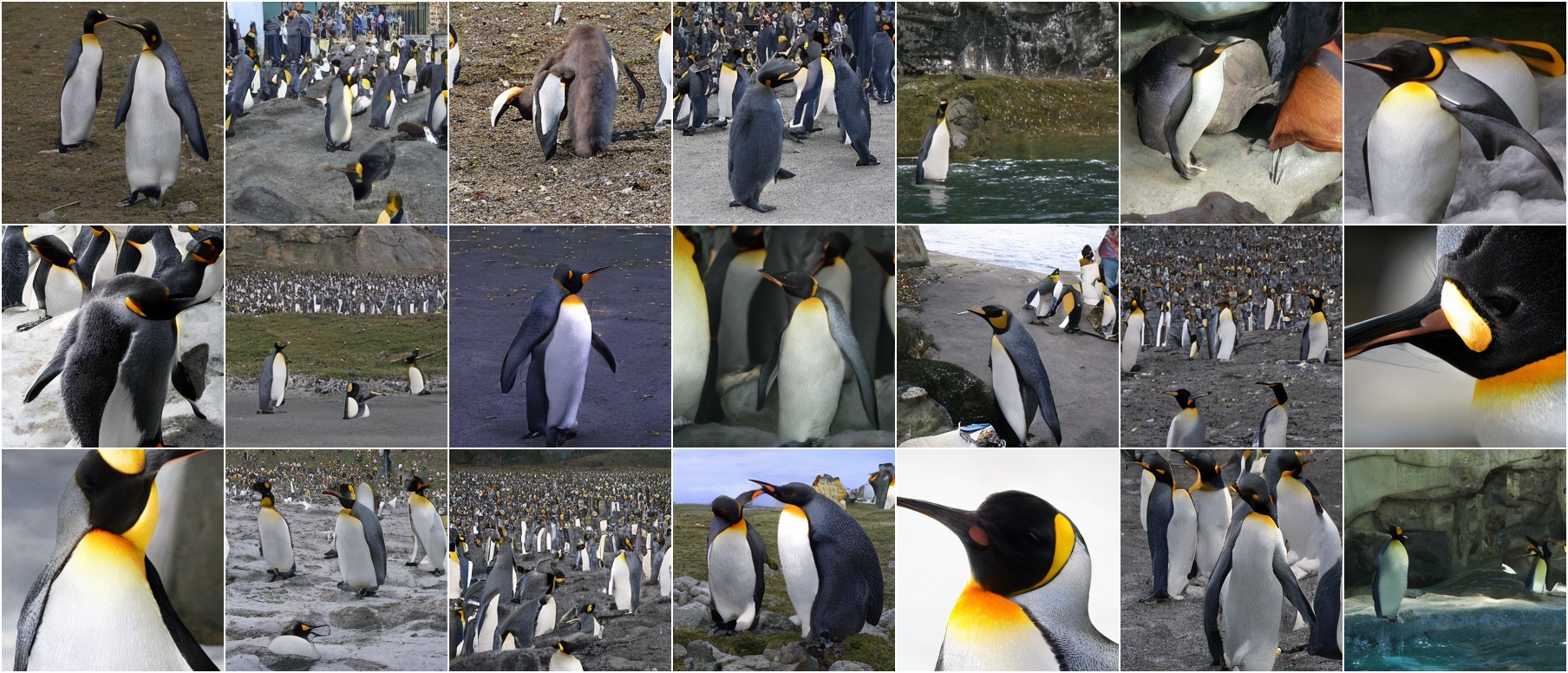}
    \vspace{-5mm}
    \caption{CAT-H/2 (Ours): class 145 (king penguin)}
\end{subfigure}
\hfill
\begin{subfigure}[t]{0.49\linewidth}
    \centering
    \includegraphics[width=\linewidth]{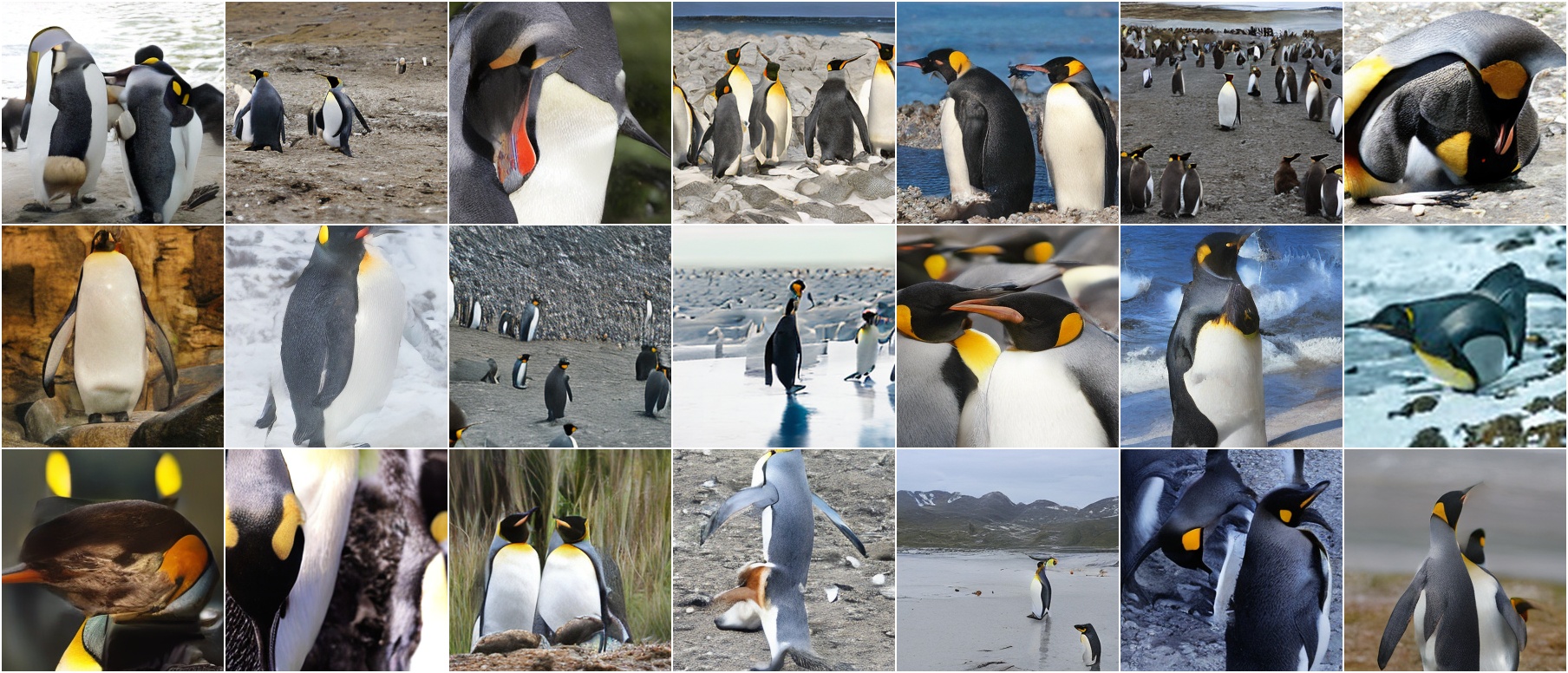}
    \vspace{-5mm}
    \caption{iMF-XL/2: class 145 (king penguin)}
\end{subfigure}


\begin{subfigure}[t]{0.49\linewidth}
    \centering
    \includegraphics[width=\linewidth]{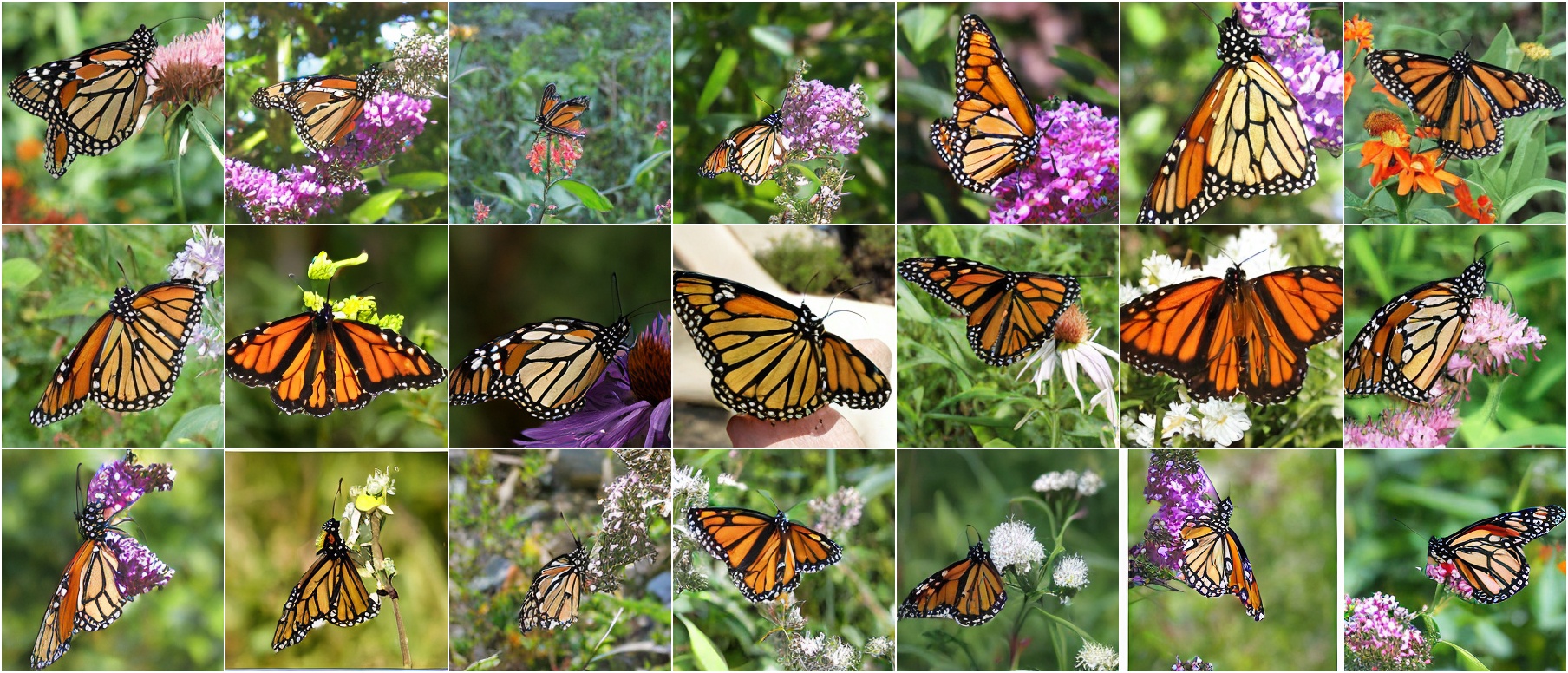}
    \vspace{-5mm}
    \caption{CAT-H/2 (Ours): class 323 (monarch)}
\end{subfigure}
\hfill
\begin{subfigure}[t]{0.49\linewidth}
    \centering
    \includegraphics[width=\linewidth]{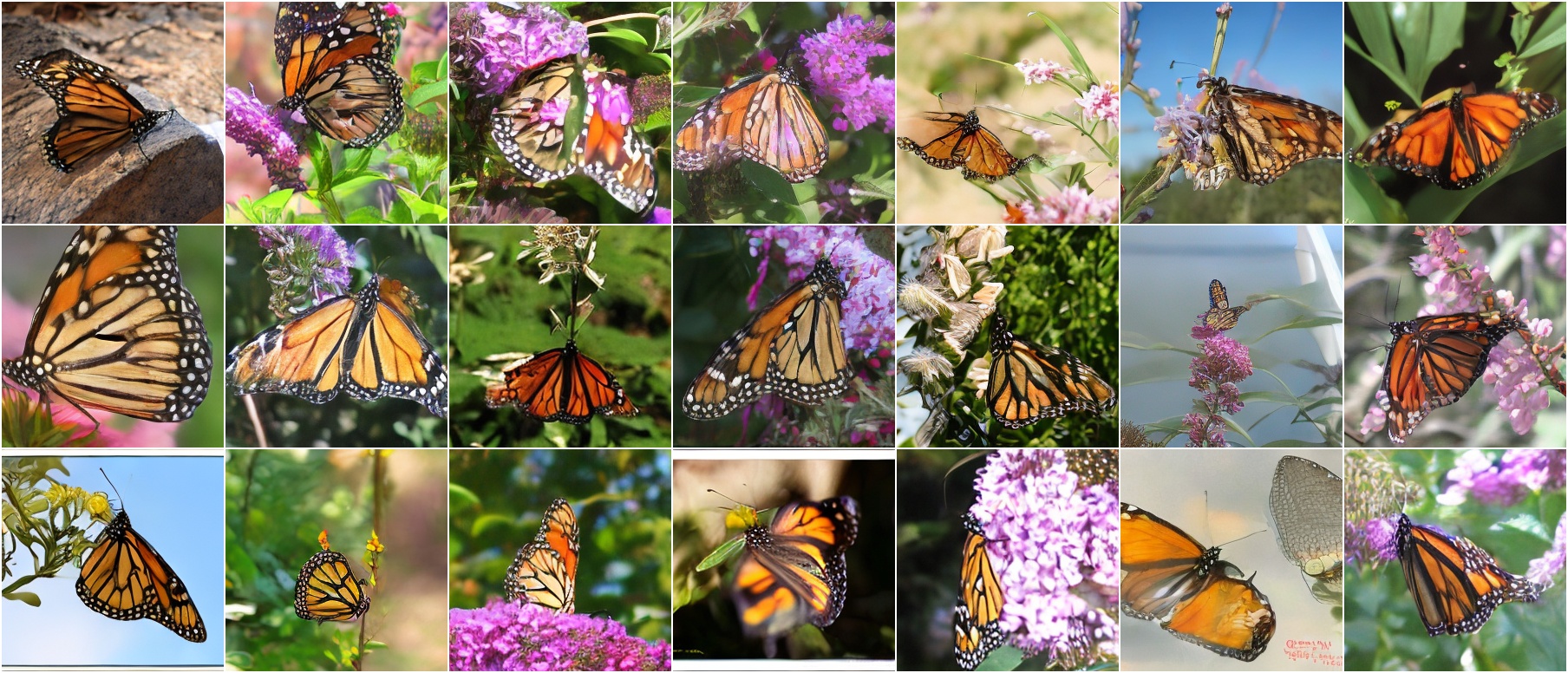}
    \vspace{-5mm}
    \caption{iMF-XL/2: class 323 (monarch)}
\end{subfigure}


\begin{subfigure}[t]{0.49\linewidth}
    \centering
    \includegraphics[width=\linewidth]{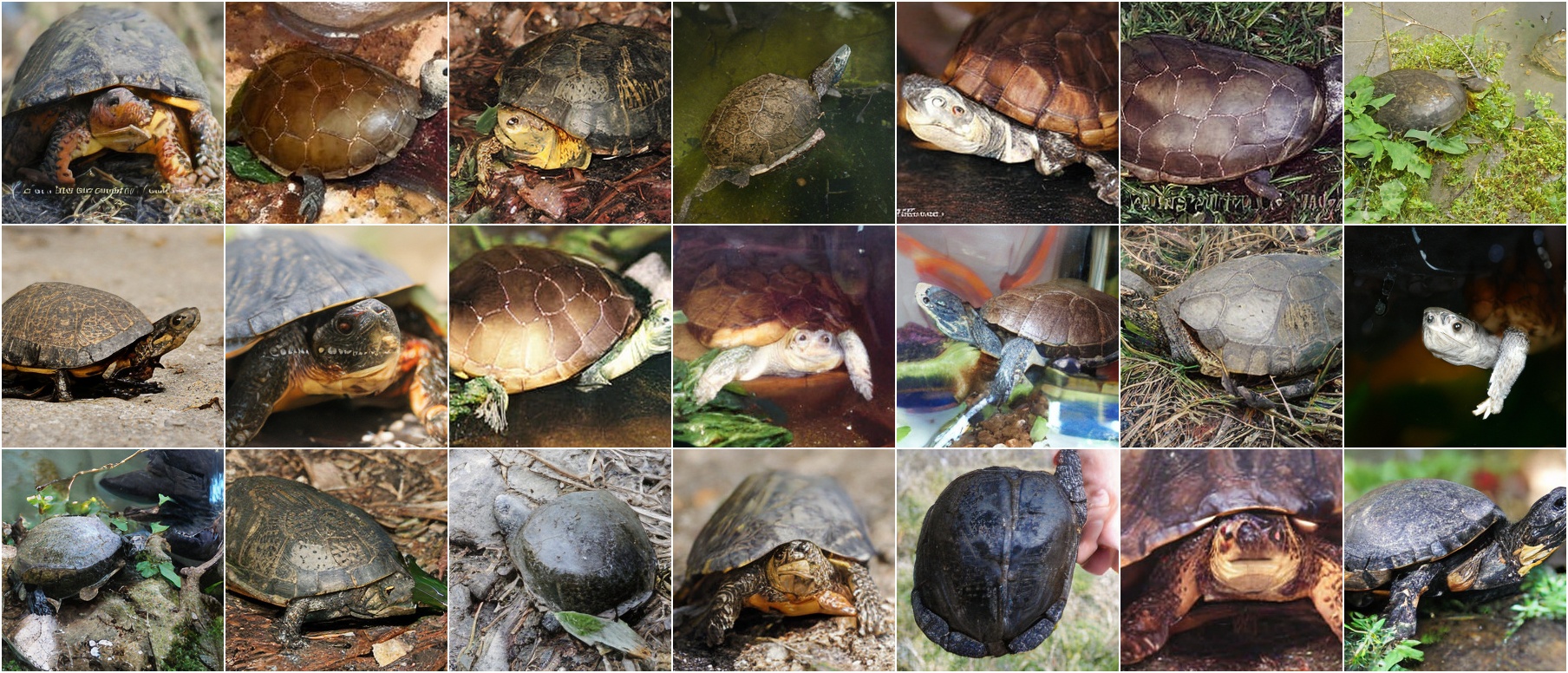}
    \vspace{-5mm}
    \caption{CAT-H/2 (Ours): class 35 (mud turtle)}
\end{subfigure}
\hfill
\begin{subfigure}[t]{0.49\linewidth}
    \centering
    \includegraphics[width=\linewidth]{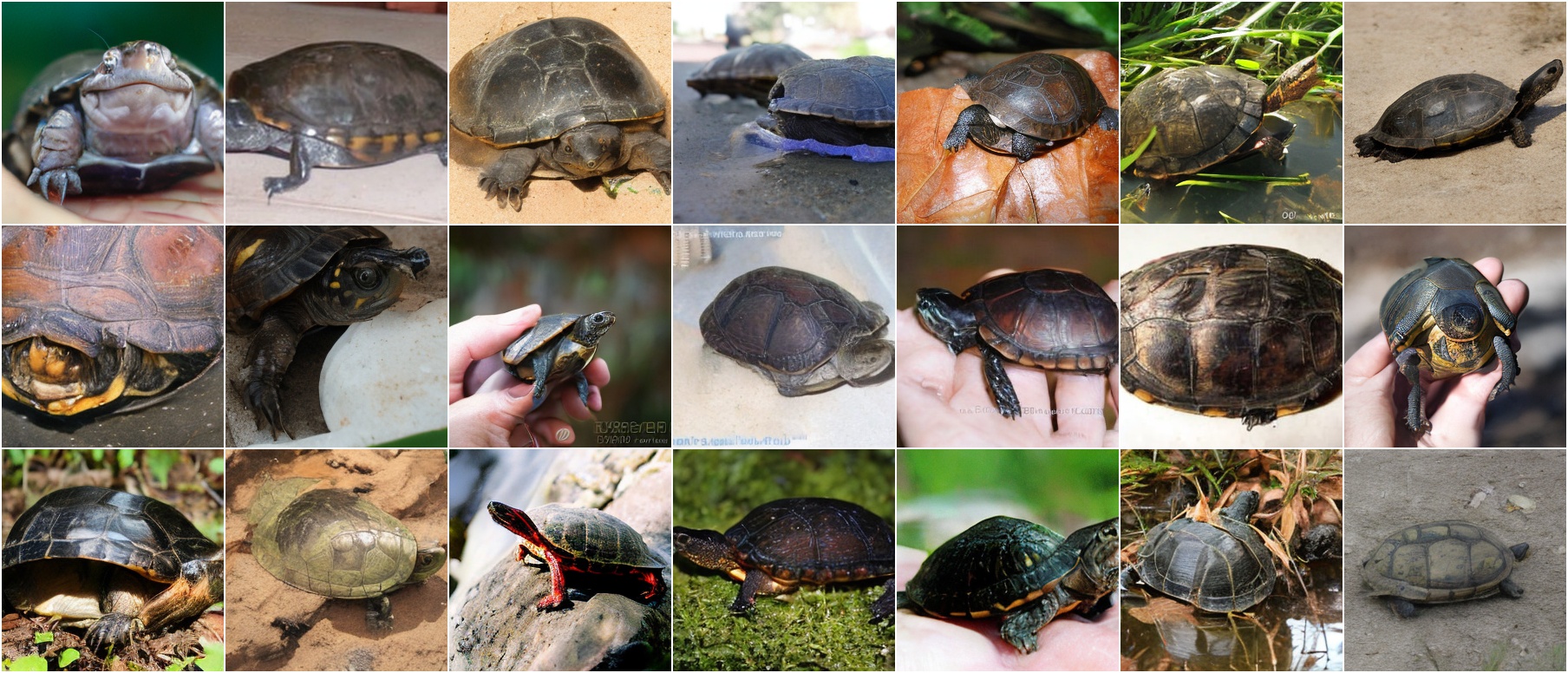}
    \vspace{-5mm}
    \caption{iMF-XL/2: class 35 (mud turtle)}
\end{subfigure}

    \caption{
    \textbf{Uncurated sample comparison with iMF.}
    We compare randomly generated ImageNet-256 samples from our model (left) and iMF-XL/2 (right), without manual curation.
    CAT-H/2 requires 60 epochs training and 166.7 inference GFLOPs, while iMF-XL/2 needs 800 epochs and 174.6 GFLOPs. 
    For sampling, our model uses truncation \(\psi=0.85\), and iMF-XL/2 follows the official configuration with interval \([0.42, 0.62]\) and \(\omega=8.0\).
    }
\label{fig:appendix qualitative_grid 2}
\vspace{-3mm}
\end{figure}

\begin{figure}[H]
\centering
\small

\begin{subfigure}[t]{0.49\linewidth}
    \centering
    \includegraphics[width=\linewidth]{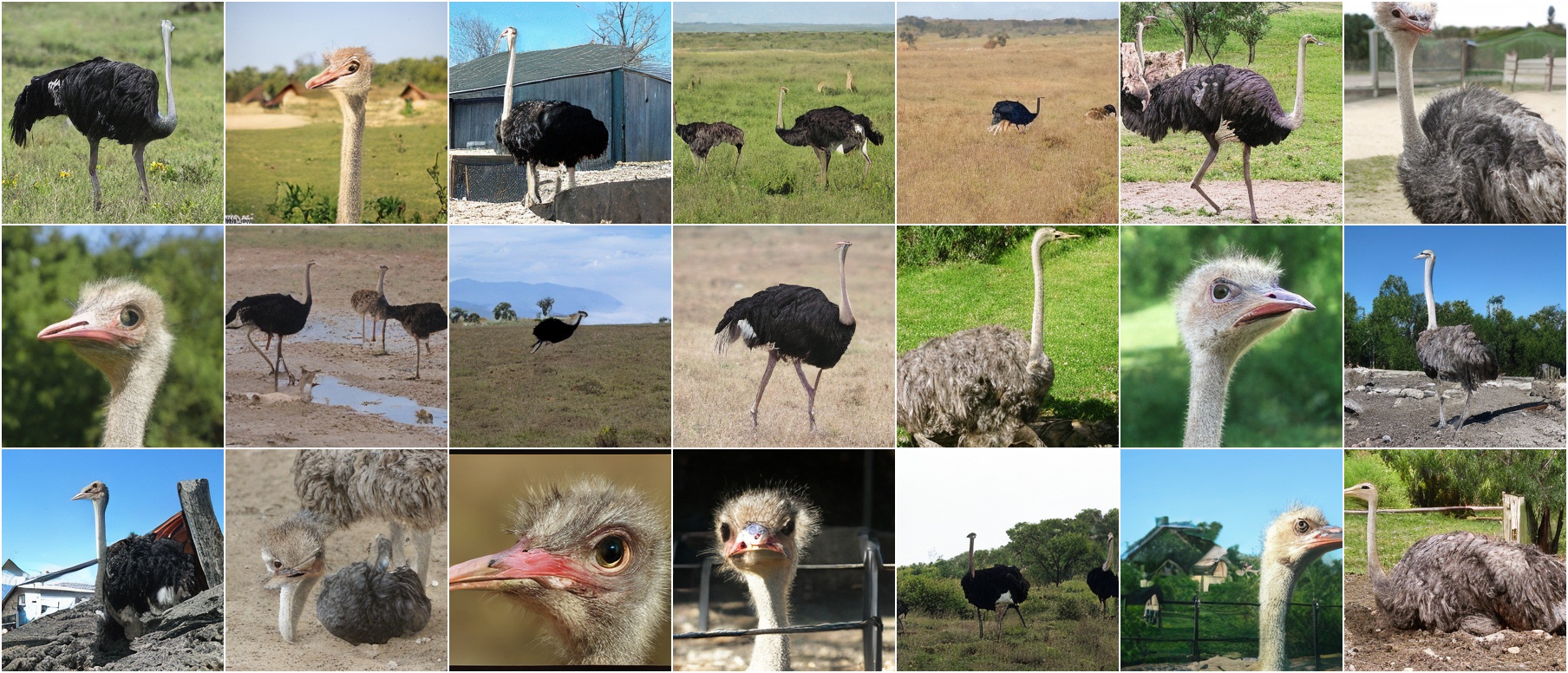}
    \vspace{-5mm}
    \caption{CAT-H/2 (Ours): class 9 (ostrich)}
\end{subfigure}
\hfill
\begin{subfigure}[t]{0.49\linewidth}
    \centering
    \includegraphics[width=\linewidth]{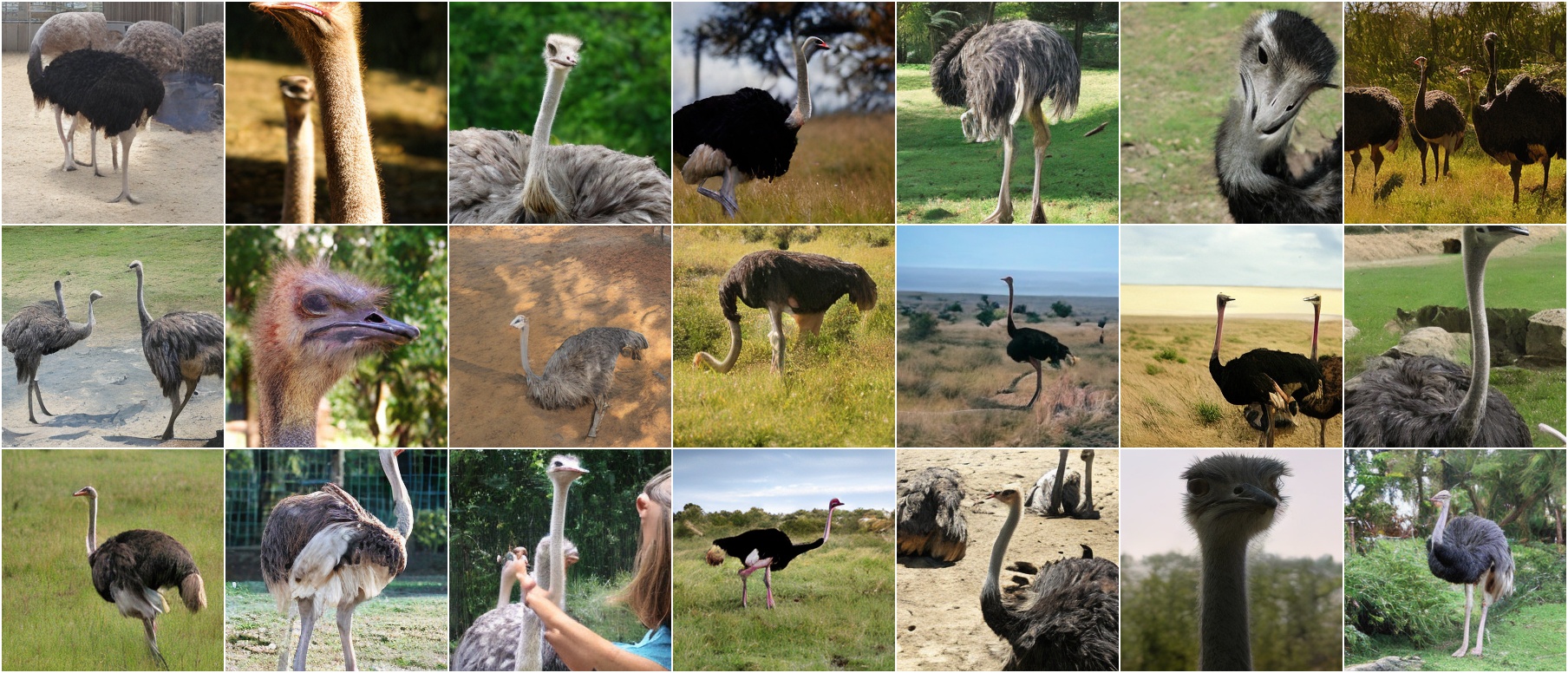}
    \vspace{-5mm}
    \caption{iMF-XL/2: class 9 (ostrich)}
\end{subfigure}


\begin{subfigure}[t]{0.49\linewidth}
    \centering
    \includegraphics[width=\linewidth]{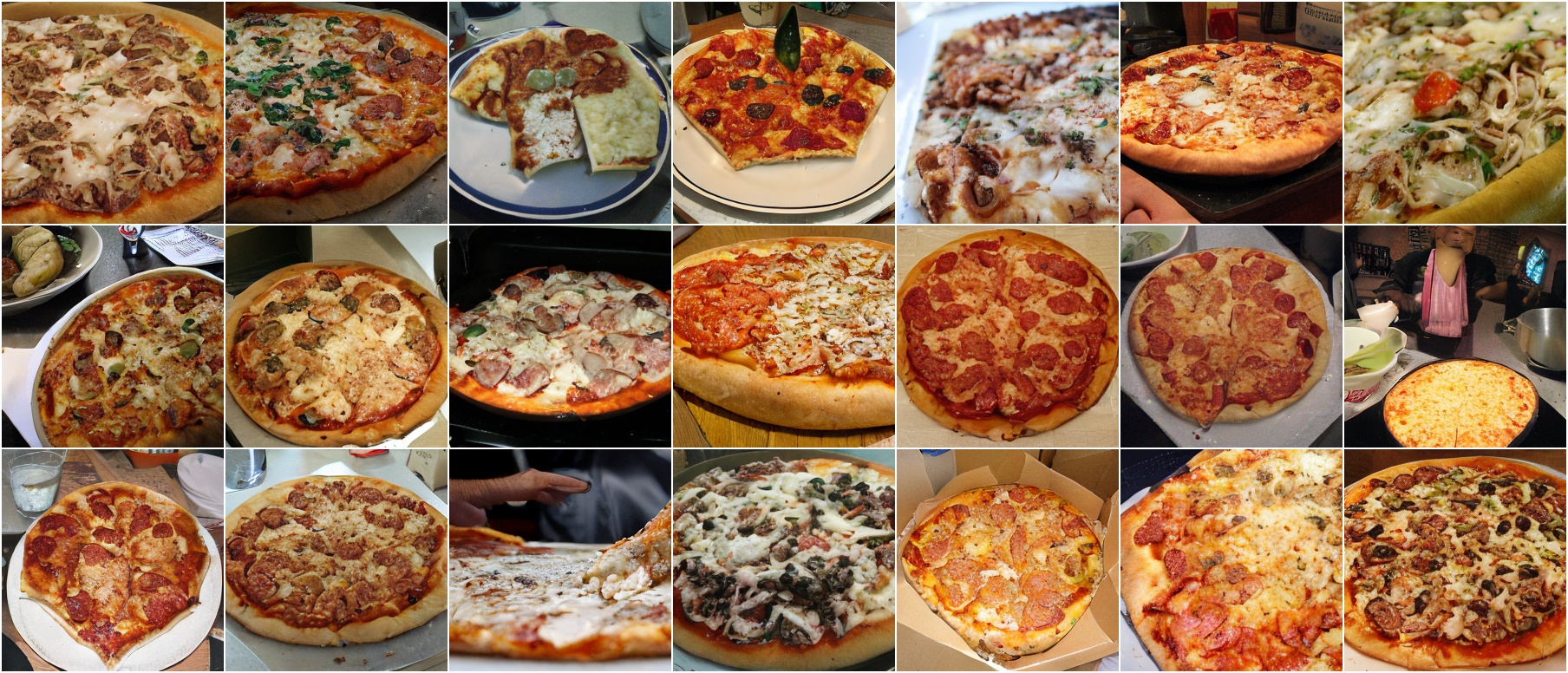}
    \vspace{-5mm}
    \caption{CAT-H/2 (Ours): class 963 (pizza)}
\end{subfigure}
\hfill
\begin{subfigure}[t]{0.49\linewidth}
    \centering
    \includegraphics[width=\linewidth]{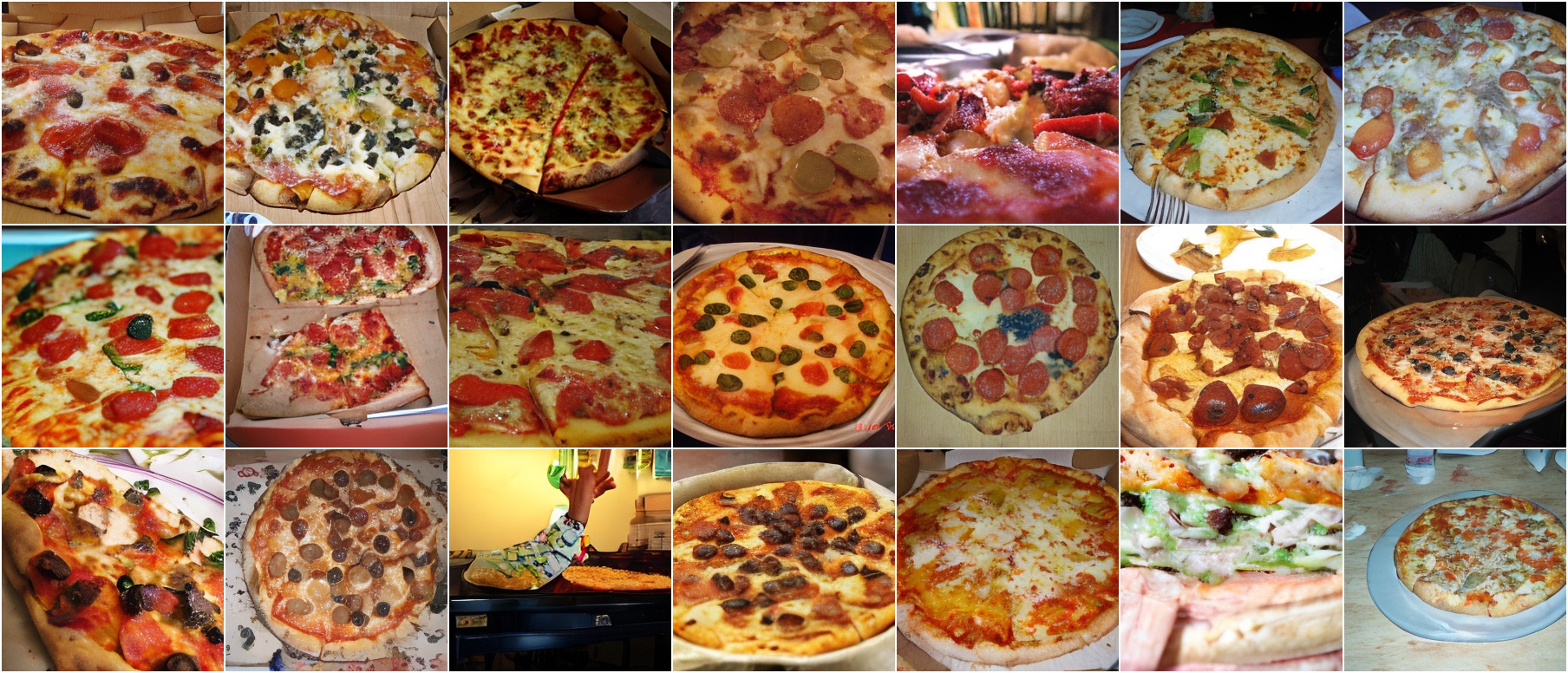}
    \vspace{-5mm}
    \caption{iMF-XL/2: class 963 (pizza)}
\end{subfigure}


\begin{subfigure}[t]{0.49\linewidth}
    \centering
    \includegraphics[width=\linewidth]{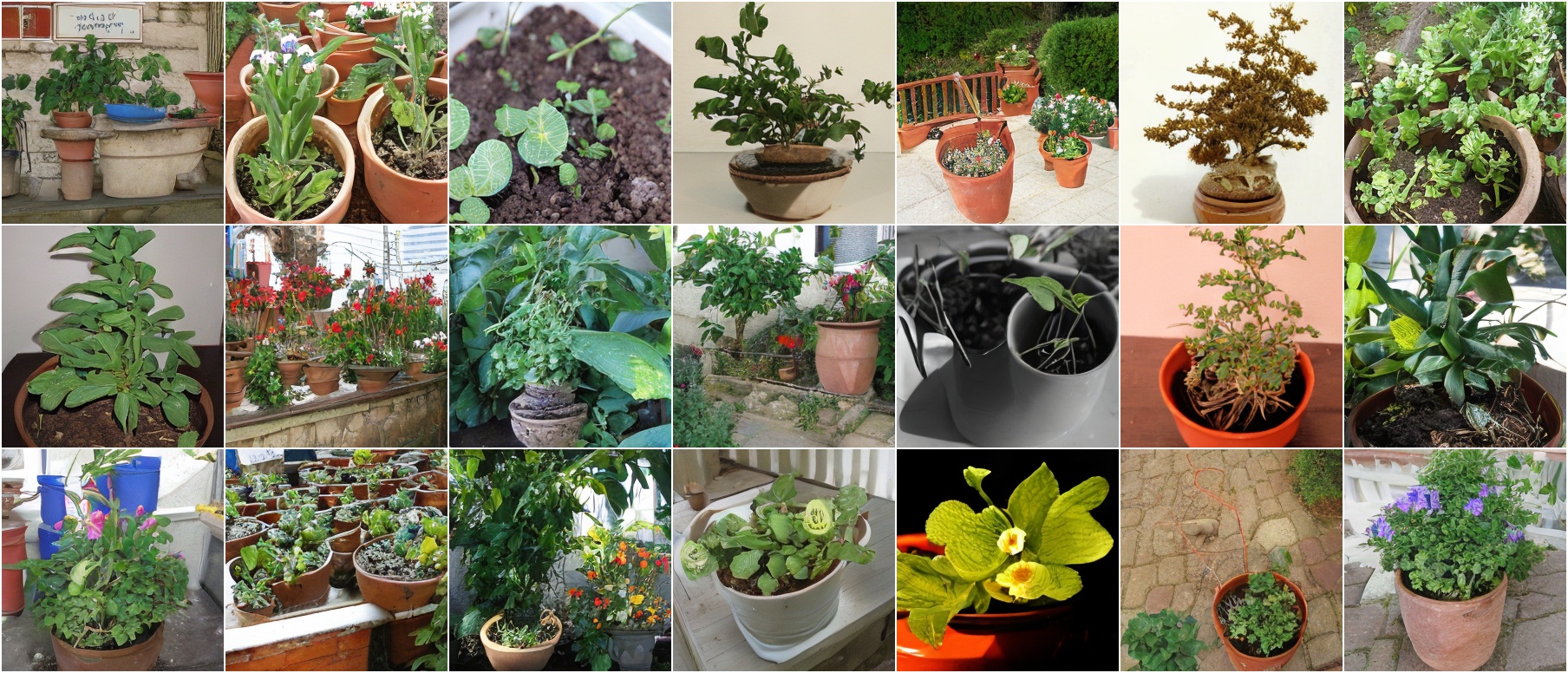}
    \vspace{-5mm}
    \caption{CAT-H/2 (Ours): class 738 (pot)}
\end{subfigure}
\hfill
\begin{subfigure}[t]{0.49\linewidth}
    \centering
    \includegraphics[width=\linewidth]{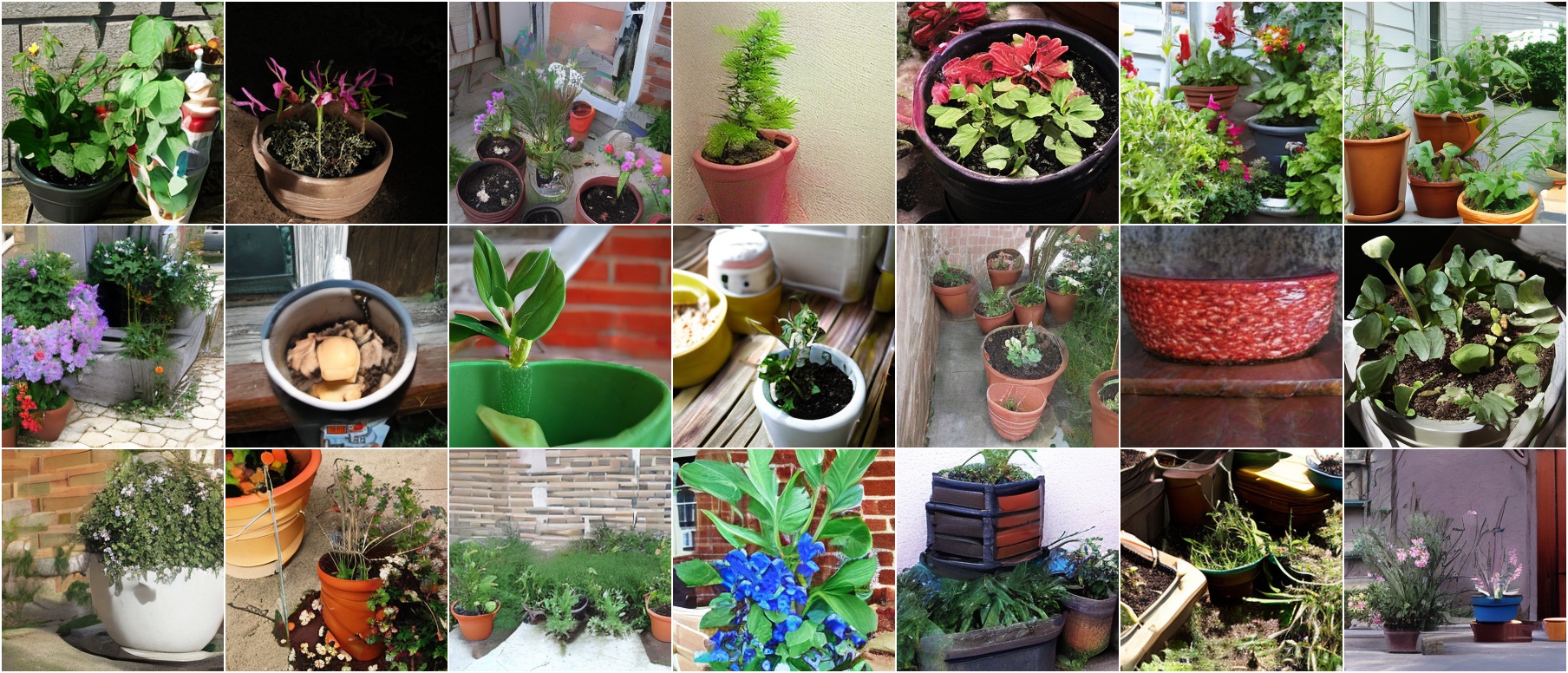}
    \vspace{-5mm}
    \caption{iMF-XL/2: class 738 (pot)}
\end{subfigure}


\begin{subfigure}[t]{0.49\linewidth}
    \centering
    \includegraphics[width=\linewidth]{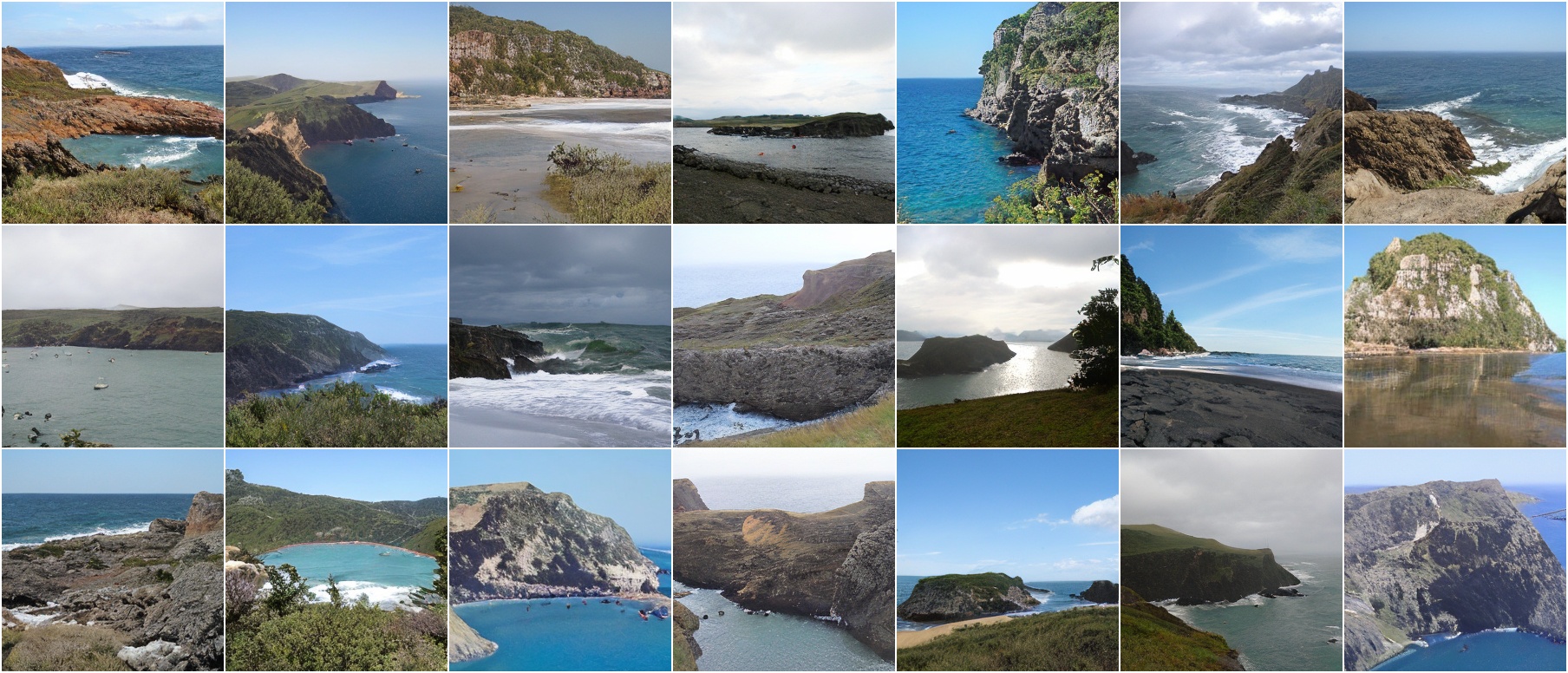}
    \vspace{-5mm}
    \caption{CAT-H/2 (Ours): class 976 (promontory)}
\end{subfigure}
\hfill
\begin{subfigure}[t]{0.49\linewidth}
    \centering
    \includegraphics[width=\linewidth]{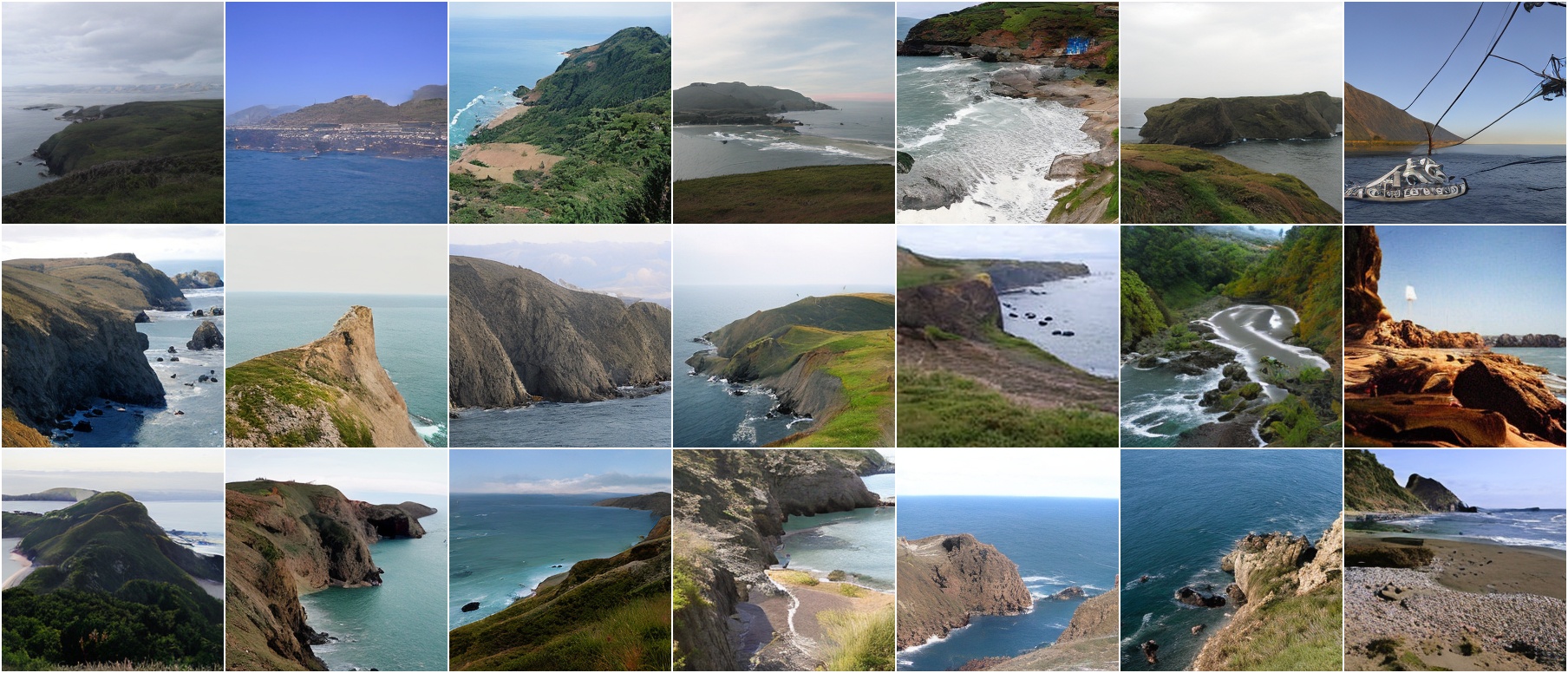}
    \vspace{-5mm}
    \caption{iMF-XL/2: class 976 (promontory)}
\end{subfigure}


\begin{subfigure}[t]{0.49\linewidth}
    \centering
    \includegraphics[width=\linewidth]{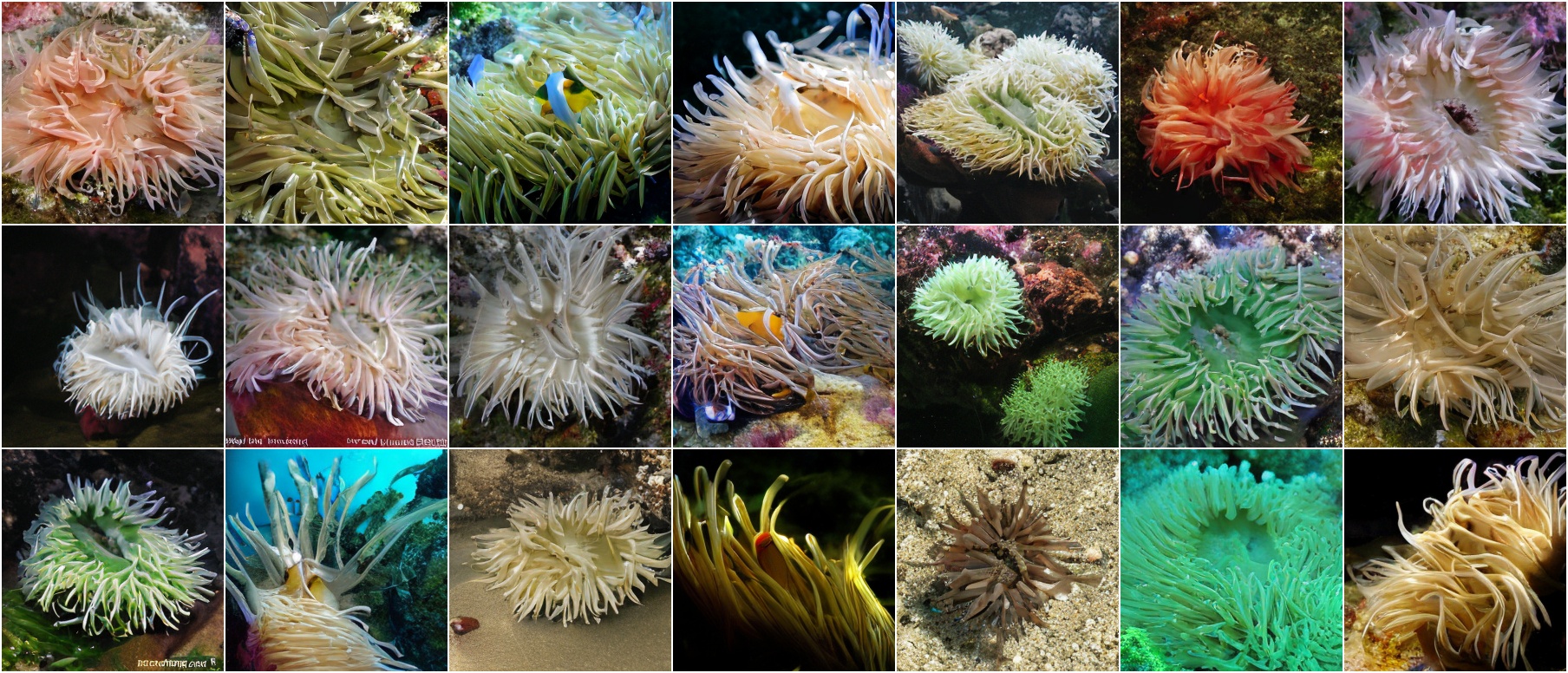}
    \vspace{-5mm}
    \caption{CAT-H/2 (Ours): class 108 (sea anemone)}
\end{subfigure}
\hfill
\begin{subfigure}[t]{0.49\linewidth}
    \centering
    \includegraphics[width=\linewidth]{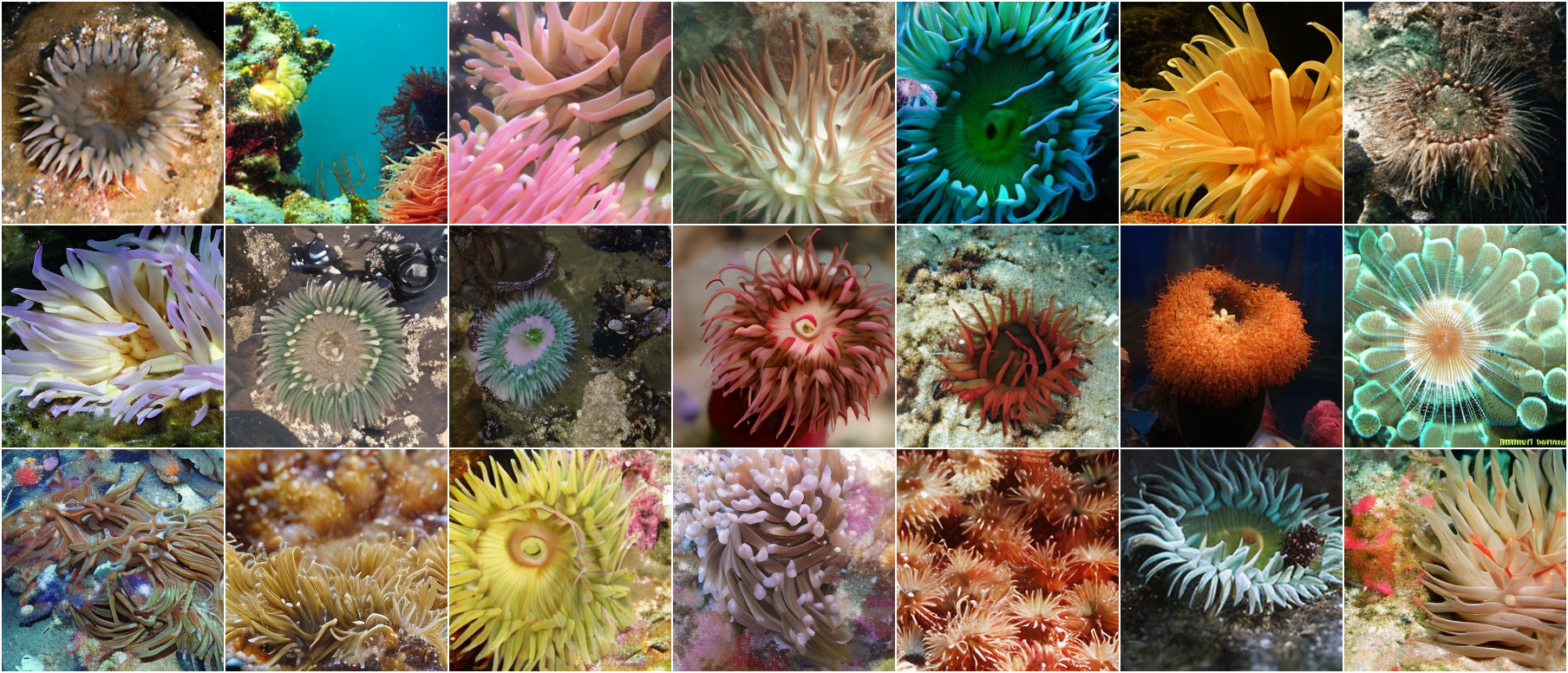}
    \vspace{-5mm}
    \caption{iMF-XL/2: class 108 (sea anemone)}
\end{subfigure}


\begin{subfigure}[t]{0.49\linewidth}
    \centering
    \includegraphics[width=\linewidth]{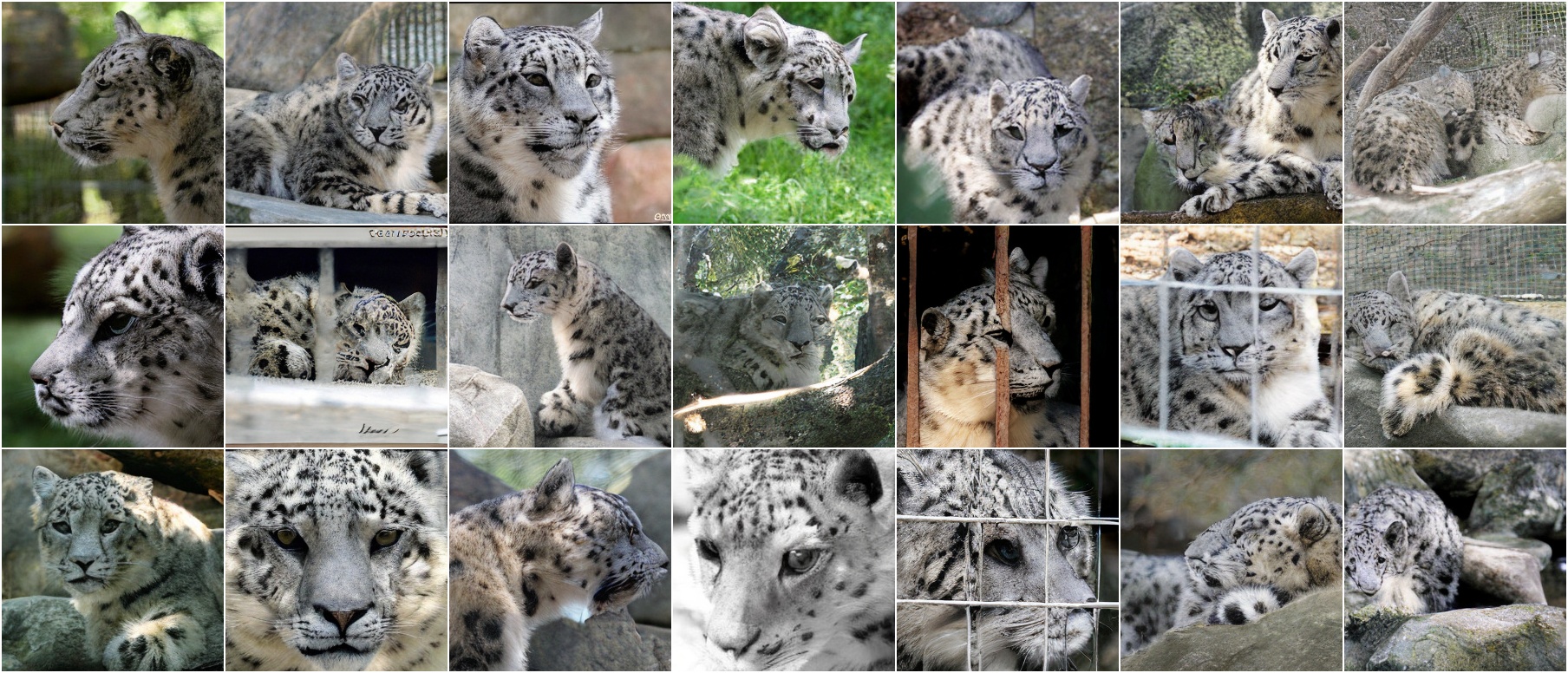}
    \vspace{-5mm}
    \caption{CAT-H/2 (Ours): class 289 (snow leopard)}
\end{subfigure}
\hfill
\begin{subfigure}[t]{0.49\linewidth}
    \centering
    \includegraphics[width=\linewidth]{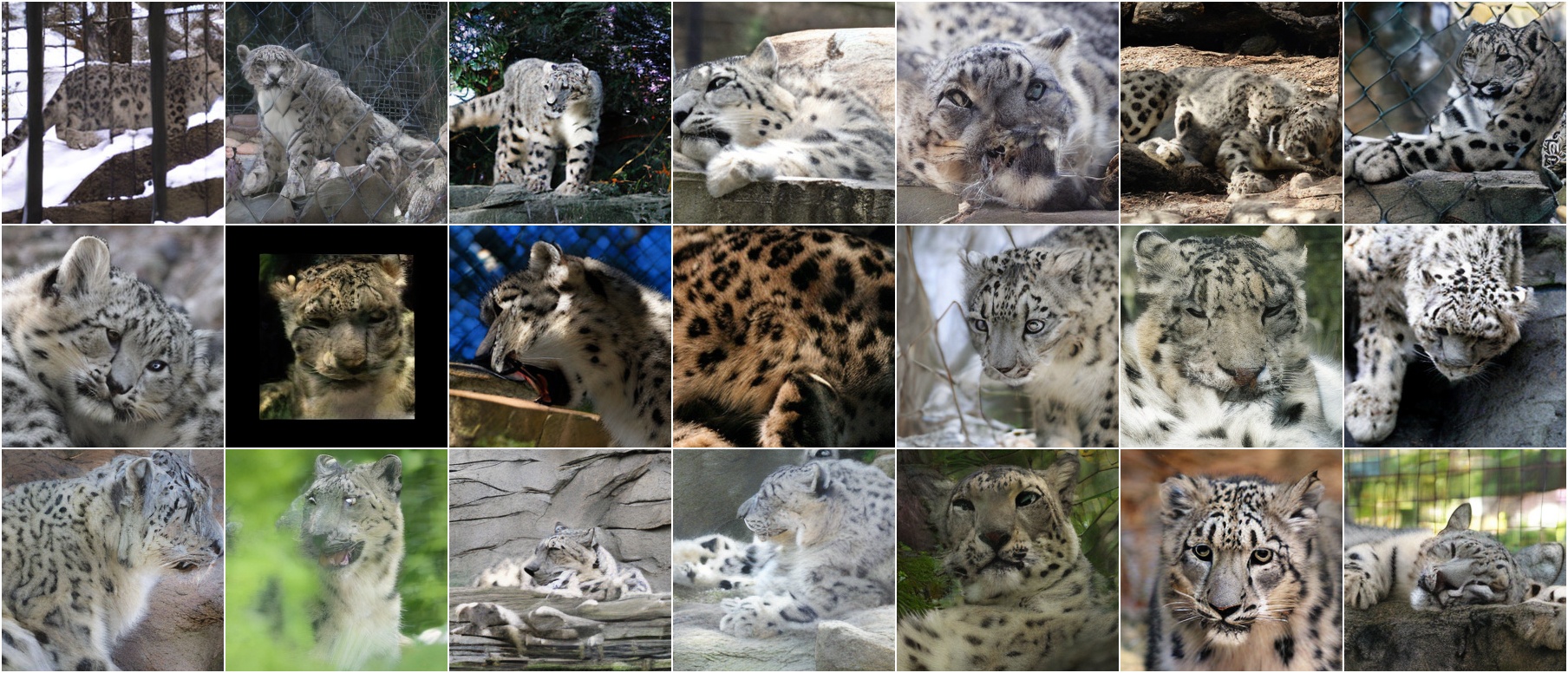}
    \vspace{-5mm}
    \caption{iMF-XL/2: class 289 (snow leopard)}
\end{subfigure}


    \caption{
    \textbf{Uncurated sample comparison with iMF.}
    We compare randomly generated ImageNet-256 samples from our model (left) and iMF-XL/2 (right), without manual curation.
    CAT-H/2 requires 60 epochs training and 166.7 inference GFLOPs, while iMF-XL/2 needs 800 epochs and 174.6 GFLOPs. 
    For sampling, our model uses truncation \(\psi=0.85\), and iMF-XL/2 follows the official configuration with interval \([0.42, 0.62]\) and \(\omega=8.0\).
    }
\label{fig:appendix qualitative_grid 3}
\vspace{-3mm}
\end{figure}



\end{document}